\documentclass[10pt,twocolumn,letterpaper]{article}

\usepackage{titling}
\usepackage{cvpr}
\usepackage{times}
\usepackage{epsfig}
\usepackage{graphicx}
\usepackage{amsmath}
\usepackage{amssymb}
\usepackage{comment}
\usepackage{subcaption}
\usepackage{capt-of}

\usepackage[breaklinks=true,bookmarks=false]{hyperref}

\cvprfinalcopy %

\def\cvprPaperID{454} %

\begin{document}

\title{Learning Detailed Face Reconstruction from a Single Image}
\author{
 Elad Richardson\textsuperscript{1} \qquad Matan Sela\textsuperscript{1}
 \qquad Roy Or-El\textsuperscript{2} \qquad Ron Kimmel\textsuperscript{1} \\
 \textsuperscript{1}Department of Computer Science, Technion - Israel Institute of Technology\\
 \textsuperscript{2}Department of Computer Science and Engineering, University of Washington\\
    {\tt\small \{eladrich,matansel,ron\}@cs.technion.ac.il} \qquad
     {\tt\small royorel@cs.washington.edu}
    }
\makeatletter
\let\@oldmaketitle\@maketitle
\renewcommand{\@maketitle}{\@oldmaketitle%
\centering
\includegraphics[width=1.0\linewidth]
    {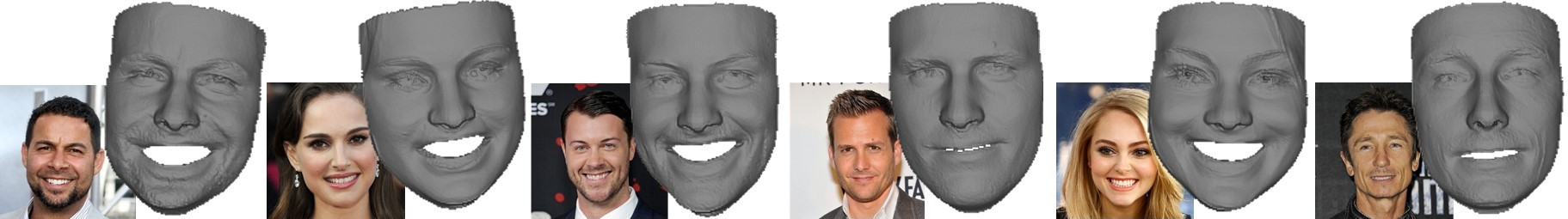}\captionof{figure}{Results of the proposed network.
    Reconstructed geometries are shown next to the corresponding input images.}\label{fig:teaser}\bigskip}
\makeatother
\maketitle
\begin{abstract}
Reconstructing the detailed geometric structure of a face from a given image is a key to many computer vision and graphics applications, such as motion capture and reenactment.
The reconstruction task is challenging as human faces vary extensively when considering expressions,
 poses, textures, and intrinsic geometries.
While many approaches tackle this complexity by using additional data
 to reconstruct the face of a single subject, extracting
 facial surface from a single image remains a difficult problem.
As a result, single-image based methods can usually provide only a rough estimate of the facial geometry.
In contrast, we propose to leverage the power of convolutional neural networks to produce
 a highly detailed face reconstruction from a single image.
For this purpose, we introduce an end-to-end CNN framework which derives the shape in a coarse-to-fine fashion.
The proposed architecture is composed of two main blocks, a network that recovers the coarse facial geometry (CoarseNet), followed by a CNN that refines the facial features of that geometry (FineNet).
The proposed networks are connected by a novel layer which renders a depth image given a mesh in 3D.
Unlike object recognition and detection problems, there are no suitable datasets for training CNNs to perform face geometry reconstruction.
Therefore, our training regime begins with a supervised phase, based on synthetic images, followed by an unsupervised phase that uses only unconstrained facial images.
The accuracy and robustness of the proposed model is demonstrated by both qualitative and quantitative evaluation tests.
\end{abstract}
\section{Introduction}
\label{sec: intro}

Faces, with all their complexities and vast number of degrees of freedom, allow
 us to communicate and express ourselves through expressions, mimics, and gestures.
Facial muscles enable us to express our emotions and feelings, while facial
 geometric features determine one's identity.
However, the flexibility of these qualities makes the recovery of facial geometry from a flat
 image a challenge.
Moreover, additional ambiguities arise as the projection of a face onto an image
 depends also on its texture and material properties,  lighting conditions,
 and viewing direction.

Various methods mitigate this uncertainty by using additional data such as a
 large photo collection of the same
 subject~\cite{roth2016adaptive, roth2015unconstrained, kemelmacher2011face,manincheddaface,piotraschke2016automated},
 continuous video frames~\cite{Weise2009,suwajanakorn2014total,cao2015real,garrido2016reconstruction}
 or a rough depth map~\cite{Weise2009,kazemi2014real}.
In many cases, however, we only have access to a single facial image.
In this setup, common schemes can be divided to 3D morphable model (3DMM) techniques
 \cite{blanz1999morphable,breuer2008automatic},
 template-based methods \cite{kemelmacher20113d,hassner2013viewing}
  and data-driven approaches  \cite{liu2016joint,tulyakov2015regressing,richardson20163d}.
Here, we propose an end-to-end neural network for reconstructing a detailed
 facial surface in 3D from a single image.
At the core of our method is the idea of breaking the reconstruction problem into
 two phases, each solved by a dedicated neural network architecture.
First, we introduce CoarseNet, a network for recovering the coarse facial geometry as
 well as the pose of the face directly from the image.
To train CoarseNet, a synthetic dataset of facial images with their matching face geometry
 and pose is synthetically generated.
The rough facial geometries are modeled using a 3DMM \cite{blanz1999morphable}, which provides a compact representation that can be recovered using the proposed network. However, this representation can only capture coarse geometry reconstruction.
Next, in order to capture fine details, we introduce FineNet, a network that operates
 on depth maps and thus is not constrained by the morphable model representation.
FineNet receives a coarse depth map, alongside the original input images and applies
 a shape-from-shading like refinement, capturing the fine facial details.
To train FineNet, we use an unlabeled set of facial images, where a dedicated loss criterion
 is introduced, to allow unsupervised training.
Finally, to connect between the CoarseNet 3DMM output and the FineNet depth map input,
 we introduce a novel layer which takes the 3DMM representation and pose parameters
 from CoarseNet, and produces a depth map that can be fed into FineNet.
This layer supports back-propagation to the 3DMM representation allowing joint training
 of the two networks, possibly refining the weights of CoarseNet.

The usage of an end-to-end network here is exciting as it connects the problem of face
 reconstruction to the rapidly expanding applications solved by CNNs,
 potentially allowing us to further improve our results following new advances in
 CNN architectures.
Moreover, it allows fast reconstructions without the need for external initialization
 or post-processing algorithms.
The potential of using a CNN for reconstructing face geometries was recently
 demonstrated in \cite{richardson20163d}.
However, their network can only produce the coarse geometry, and must be given
 an aligned template model as initialization.
These limitations force their solution to depend on external algorithms for pose
 alignment and detail refinement.

The main contributions of the proposed method include:
\begin{itemize}
\item An end-to-end network-based solution for facial surface
 reconstruction from a single image, capable of producing detailed geometric structures.
\item A novel rendering layer, allowing back-propagation from a rendered
 depth map to the 3DMM model.
 \item A network for data refinement, using a dedicated loss criterion,
 motivated by axiomatic shape-from-shading objectives.
\item A training scheme that bypasses the need for manually labeled data
 by utilizing only synthetic data and unlabeled facial images.
\end{itemize}
\section{Related Work} \label{sec: related}
Automatic face reconstruction attracts a lot of attention in the computer vision
 and computer graphics research communities.
The available solutions differ in their assumptions about the input data, the priors
 and the techniques they use.
When dealing with geometry reconstruction from a single image,
 the problem is ill-posed.
Still, there are ways for handling the intrinsic ambiguities in geometry
 reconstruction from one image.
These solutions can be roughly divided into the following categories:

\textbf{3DMM Methods.}
In \cite{blanz1999morphable}, Vetter and Blantz introduced the
 3D Morphable Model (3DMM), a principal components analysis (PCA) basis for representing faces.
One of the advantages of using the 3DMM is that the solution space is constrained to
 represent only likely solutions, thereby simplifying the problem.
While the original paper assumes manual initialization, more recent efforts propose an
 automatic reconstruction process \cite{breuer2008automatic,zhu2015high}.
Still, the automated initialization pipelines usually do not produce the same quality
 of reconstructions when only one image is used, as noted in \cite{piotraschke2016automated}.
In addition, the 3DMM solutions cannot extract fine details since they are not spanned by the principal components.

\textbf{Template-Based Methods.}
An alternative approach is to solve the problem by deforming a template to match the input image. One notable paper is that of Kemelmacher-Shlizerman and Basri~\cite{kemelmacher20113d}.
There, a reference model is aligned with the face image and a shape-from-shading (SfS) process is applied to mold the reference model to better match the image. Similarly, Hassner~\cite{hassner2013viewing} proposed to jointly maximize the appearance and depth similarities between the input image and a template face using SIFTflow~\cite{liu2008sift}. While these methods do a better job in recovering the fine facial features, their capability to capture the global face structure is limited by the provided template initialization.

\textbf{Data-Driven Methods.}
A different approach to the problem uses some form of regression to connect between the
 input image and the reconstruction representation.
Some methods apply a regression model from a set of sparse landmarks
 \cite{aldrian2010linear,dou2014robust,liu2015cascaded}, while others apply a regression
 on features derived from the image \cite{lei2008face,castelan20083d}.
\cite{liu2016joint} applies a joint optimization process that ties the sparse landmarks
 with the face geometry, recovering both.
Recently, a network was proposed to directly reconstruct the geometry from the image \cite{richardson20163d},
 without using sparse information or explicit features.
That paper demonstrated the potential of using a network for face reconstruction.
Still, it required external procedures for fine details extraction as well as
 initial guess of the face location, size, and pose.

In a sense, the proposed solution combines all of these different procedures.
Specifically, a \textbf{3DMM} is used to define the input for a \textbf{Template-Based}
 refinement step, where both parts are learned using a \textbf{Data-Driven} model.

\section{Coarse Geometry Reconstruction}\label{sec: coarse}
The first step in our framework is to extract the coarse facial
 geometry and pose from the given image.
Our solution is motivated by two recent efforts, \cite{richardson20163d} which proposed
 to train a network for face reconstruction using synthetic data, and \cite{zhu2015face}
 which solved the face alignment problem using a network.
Although the methods focus on different problems, they both use an iterative framework
 which utilizes a 3D morphable model.
The proposed method integrates both concepts into a holistic alignment and geometry
 reconstruction solution.

\subsection{Modeling The Solution Space}
In order to solve the reconstruction problem using a CNN, a representation of the solution space
 is required.
To model the facial geometries we use a 3D morphable model \cite{blanz1999morphable}, where an
 additional blendshape basis is used to model expressions, as suggested in \cite{chu20143d}.
This results in the following linear representation
  \begin{equation}
  S=\mu_{S}+A_{\small\mbox{id}}\alpha_{\small\mbox{id}}+A_{\small\mbox{exp}}\alpha_{\small\mbox{exp}}.
  \end{equation}
Where $\mu_{S}$ is the average 3D face, $A_{\small\mbox{id}}$ is the principal component basis,
 $A_{\small\mbox{exp}}$ is the blendshape basis, and $\alpha_{\small\mbox{id}}$ and
  $\alpha_{\small\mbox{exp}}$ are the corresponding coefficient vectors.
 $A_{\small\mbox{id}}$ and $A_{\small\mbox{exp}}$ are collected from the Bosphorus
 dataset~\cite{savran2008bosphorus} as in \cite{richardson20163d},
  where the identity is modeled using $200$ coefficients, and the expression using $84$.

For projecting the 3D model to the image plane, we assume a parallel weak perspective projection.
\begin{equation}
    \left[\begin{array}{c}
    p_x\\
    p_y
  \end{array}\right]=\left[\begin{array}{ccc}
  f & 0 & 0\\
  0 & f & 0
\end{array}\right]\left[R|t\right]\left[\begin{array}{c}
 P_x\\
 P_y\\
 P_z\\
 1
\end{array}\right],
\end{equation}
 where $p,P$ are the pixel location in the image plane and in the world coordinate system,
  respectively, $f$ is the focal length, and $\left[R|t\right]$ is the extrinsic matrix of the camera.  %
Hence, the face alignment is modeled using only $6$ parameters: $3$ Euler angles,
 a 2D translation vector and a scale.
The pose parameters are normalized so that a zero vector would correspond to a centralized
 front facing face.
Overall, we have a representation of $290$ parameters for both geometry and pose.
We will denote this representation as $r$.
\subsection{The CoarseNet Training Framework}
The realization that the power of single-pass systems is limited, has made the
 application of iterative networks popular.
While some methods \cite{sun2013deep,li2015convolutional} use a cascade of networks to refine
 their results, it has been shown that a single network can also be trained to iteratively correct its prediction.
This is done by adding feedback channels to the network that represent the previous output of the network as a set of feature maps.
The network is then trained to refine its prediction based on both the original input
 and the feedback channels.
This idea was first proposed by Carreira \textit{et al.} in~\cite{carreira2015human}.
\subsubsection{Feedback Representation}
\begin{figure} %
    \centering
    \begin{subfigure}[b]{0.075\textwidth}
      \includegraphics[width=\textwidth]{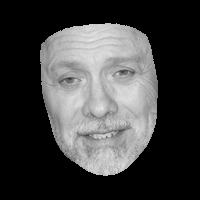}
      \caption{}
      \label{subfig:1-vis}
    \end{subfigure}
    \begin{subfigure}[b]{0.075\textwidth}
      \includegraphics[width=\textwidth]{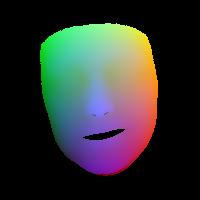}
      \caption{}
      \label{subfig:1-pncc}
    \end{subfigure}
    \begin{subfigure}[b]{0.075\textwidth}
      \includegraphics[width=\textwidth]{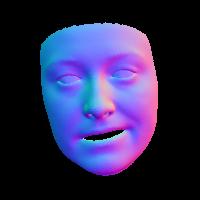}
      \caption{}
      \label{subfig:1-normals}
    \end{subfigure}
    \begin{subfigure}[b]{0.075\textwidth}
      \includegraphics[width=\textwidth]{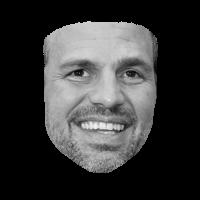}
      \caption{}
      \label{subfig:2-vis}
    \end{subfigure}
    \begin{subfigure}[b]{0.075\textwidth}
      \includegraphics[width=\textwidth]{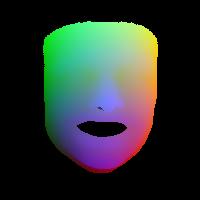}
      \caption{}
      \label{subfig:2-pncc}
    \end{subfigure}
    \begin{subfigure}[b]{0.075\textwidth}
      \includegraphics[width=\textwidth]{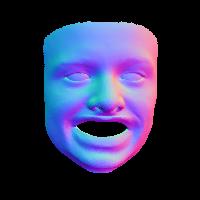}%
      \caption{}
      \label{subfig:2-normals}
    \end{subfigure}
    \caption{Feedback representation. (\subref{subfig:1-vis},\subref{subfig:2-vis}) are masked input images, (\subref{subfig:1-pncc},\subref{subfig:2-pncc}) are the corresponding PNCCs of the network's output and (\subref{subfig:1-normals},\subref{subfig:2-normals}) are the resulting normal maps.}
    \label{fig:vis_example1}
\end{figure}

Defining the feedback channels of the previous output of the network is crucial,
 as it would affect the overall performance of our iterative framework.
Roughly speaking, we would like the feedback channels to properly represent the current state of the
 coarse facial geometry.
In practice, different types of feedback channels would emphasize different features of the current state.
For instance, in~\cite{zhu2015face} the Projected Normalized Coordinate Code (PNCC) was introduced.
This feature map is computed by first normalizing the average face and painting the RGB channels
 of the current vertices with the $x$, $y$ and $z$ coordinates of the corresponding vertex on
  the average model, see Figures~\ref{subfig:1-pncc} and~\ref{subfig:2-pncc}.

Next, we propose to use the normal map as an additional channel, where each
 vertex is associated with its normal coordinates.
These normal values are then rendered as RGB values. %
The purpose of the normal map is to represent more local features of the coarse geometry,
 which are not emphasized by the PNCC.
The proposed solution uses both feedbacks, creating a richer representation of the shape.
Examples of these representations are shown in Figure~\ref{fig:vis_example1}.

\begin{figure*}
  \includegraphics[width=\linewidth]{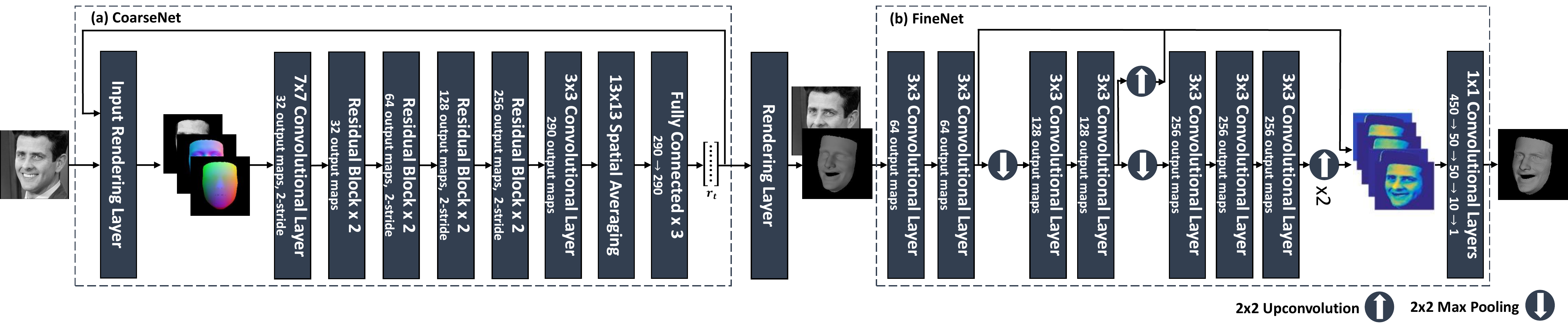}
\caption{The End-to-End network, composed of CoarseNet, FineNet and the rendering layer.}
\label{fig:end_to_end}
\end{figure*}
\subsubsection{Acquiring The Data}
\label{subsec:data_acq}
In order to train the proposed framework, a large dataset of 3D faces is required.
However, due to the complexity in acquiring accurate 3D scans for a large group of people,
 no such dataset is currently available.
Note that unlike different annotations, such as landmark positions, which can be manually
 collected for an existing set of unlabeled images, the 3D geometry has to be captured
 jointly with the photometric data.
A possible solution would be to apply existing reconstruction methods to 2D images and use these reconstructions as labels.
However, such an approach would limit the reconstruction quality to that of the
 reconstruction method we use.

Here, we choose to follow the line of thought proposed in \cite{richardson20163d} and create
 a synthetic dataset by drawing random representations of geometry and pose, $r_{gt}$, which
 are then rendered using random texture, lighting, and reflectance.
This process provides a dataset of 2D images, for which the pose and corresponding geometry
 are known by construction.
The iterative refinement process is then simulated by drawing another set of parameters, $r_t$,
 which is sampled between $r_{gt}$ and a random set of parameters, $r_{rnd}$.
\begin{equation}
  r_{t}=\beta\cdot r_{gt}+\left(1-\beta\right)\cdot r_{rnd},  \qquad 0\leq \beta \leq 1,
\end{equation}
$r_t$ represents the current estimation of the solution, and is used to generate the PNCC and normal map.
The network is then trained to predict the ground-truth, $r_{gt}$, representation from the current one, $r_t$.
Note, that unlike \cite{richardson20163d} our representation $r$ captures not only the geometry,
 but also the pose.
Hence $r_{gt}$ and $r_{rnd}$ can vary also in their position and orientation.

\subsection{The CoarseNet Architecture and Criterion}
CoarseNet is based on the ResNet architecture \cite{he2015deep}, and is detailed in Figure \ref{fig:end_to_end}. Note that the input layer includes the feedback channel and that a grayscale image is used.
The last element in the proposed architecture is the training criterion.
As our representation is composed of both geometry and pose parameters, we choose
 to apply a different training criterion for each part of the representation.
For the geometry we apply the Geometry Mean Square Error (GMSE)
 suggested in \cite{richardson20163d},
\begin{equation}
    L\left(\hat{\alpha},\alpha\right)=\left\Vert \left[A_{\small\mbox{id}}
      |A_{\small\mbox{exp}}\right]\hat{\alpha}  -\left[A_{\small\mbox{id}}|
       A_{\small\mbox{exp}}\right]\alpha\right\Vert _{2}^{2},
\end{equation}
 where $\hat{\alpha}$ is the geometry received from the network, and $\alpha$ is the known geometry.
The idea behind GMSE is to take into account how the different coefficients affect the resulting geometry.
For the pose parameters we found that a simple MSE loss over the 6 parameters is sufficient.
We weigh the two loss criteria so that we get approximately the same initial error for both.

\subsection{Using CoarseNet}
\begin{figure}[b]%
    \centering
    \begin{subfigure}[b]{0.075\textwidth}
      \includegraphics[width=\textwidth]{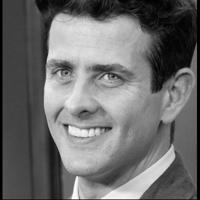}
    \end{subfigure}
    \begin{subfigure}[b]{0.075\textwidth}
        \includegraphics[width=\textwidth]{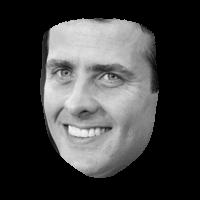}
    \end{subfigure}
    \begin{subfigure}[b]{0.075\textwidth}
        \includegraphics[width=\textwidth]{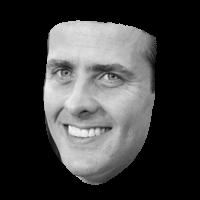}
    \end{subfigure}
    \begin{subfigure}[b]{0.075\textwidth}
      \includegraphics[width=\textwidth]{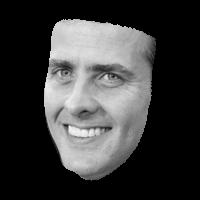}
    \end{subfigure}
    \begin{subfigure}[b]{0.075\textwidth}
        \includegraphics[width=\textwidth]{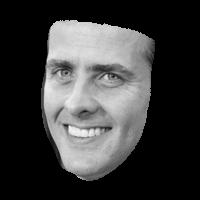}
    \end{subfigure}
    \begin{subfigure}[b]{0.075\textwidth}
        \includegraphics[width=\textwidth]{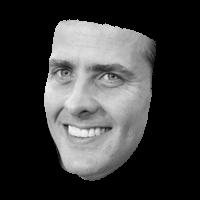}
    \end{subfigure}

    \begin{subfigure}[b]{0.075\textwidth}
      \includegraphics[width=\textwidth]{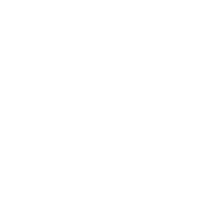}
    \end{subfigure}
    \begin{subfigure}[b]{0.075\textwidth}
        \includegraphics[width=\textwidth]{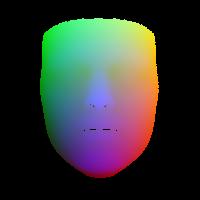}
    \end{subfigure}
    \begin{subfigure}[b]{0.075\textwidth}
        \includegraphics[width=\textwidth]{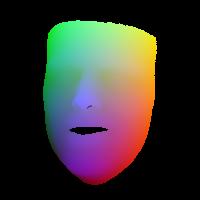}
    \end{subfigure}
    \begin{subfigure}[b]{0.075\textwidth}
      \includegraphics[width=\textwidth]{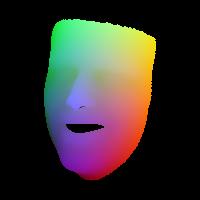}
    \end{subfigure}
    \begin{subfigure}[b]{0.075\textwidth}
        \includegraphics[width=\textwidth]{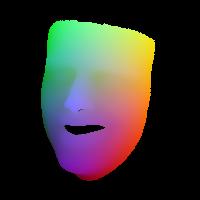}
    \end{subfigure}
    \begin{subfigure}[b]{0.075\textwidth}
        \includegraphics[width=\textwidth]{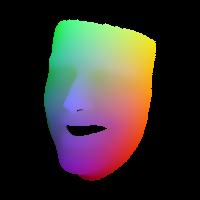}
    \end{subfigure}

    \begin{subfigure}[b]{0.075\textwidth}
      \includegraphics[width=\textwidth]{images/iterations/placeholder}
    \end{subfigure}
    \begin{subfigure}[b]{0.075\textwidth}
        \includegraphics[width=\textwidth]{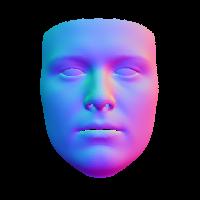}
    \end{subfigure}
    \begin{subfigure}[b]{0.075\textwidth}
        \includegraphics[width=\textwidth]{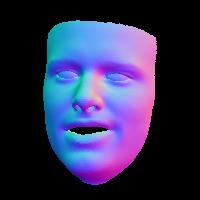}
    \end{subfigure}
    \begin{subfigure}[b]{0.075\textwidth}
      \includegraphics[width=\textwidth]{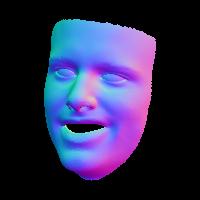}
    \end{subfigure}
    \begin{subfigure}[b]{0.075\textwidth}
        \includegraphics[width=\textwidth]{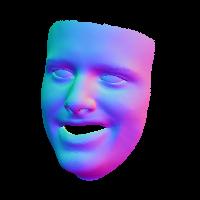}
    \end{subfigure}
    \begin{subfigure}[b]{0.075\textwidth}
        \includegraphics[width=\textwidth]{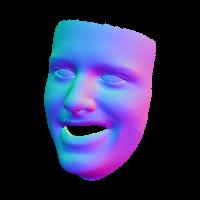}
    \end{subfigure}
    \caption{Progress through iterations. For each iteration the following are shown from top to bottom:
      The cropped input image, the PNCC and the normal map.}
    \label{fig:iter-result}
\end{figure}

We feed CoarseNet with a $200\times 200$ image of a face.
Such an image can be automatically acquired using a standard face detector,
 such as the Viola-Jones detector ~\cite{viola2001rapid}.
The initial parameters vector, $r_0$, is set to zeros, corresponding to a centered mean face $\mu_S$.
In addition, the input image is always masked in accordance with the visible vertices
 in the feedback channel.
The masking is applied in order to improve our generalization
  capability from synthetic data to real-world images, as our synthetic data is more
  accurate for the head region.
Although the mask is inaccurate in the first iteration, it is gradually refined.
The network is then applied iteratively, producing the updated geometry $r_t$, which is
 used to create the new feedback input.
This process is repeated until convergence, as shown in Figure \ref{fig:iter-result}.

\section{The Coarse to Fine Approach} \label{sec: fine}
For many tasks, such as face frontalization~\cite{zhu2015high,hassner2015effective},
 reconstructing the coarse geometry is sufficient.
However, reconstructing fine geometric structures such as wrinkles
 could be useful for other applications, see \cite{cao2015real,Sela2015}.
It is clear that while working in the morphable model domain, we cannot capture such details.
To solve that, we transfer the problem to the unconstrained image plane, representing the
 geometry as a depth map.
The role of the proposed FineNet would then be to modify the given coarse depth map,
  based on the original image, for capturing the fine details.

\subsection{The Rendering Layer}
\begin{figure}
    \centering
    \begin{subfigure}[b]{0.5\textwidth}
      \includegraphics[width=\textwidth]{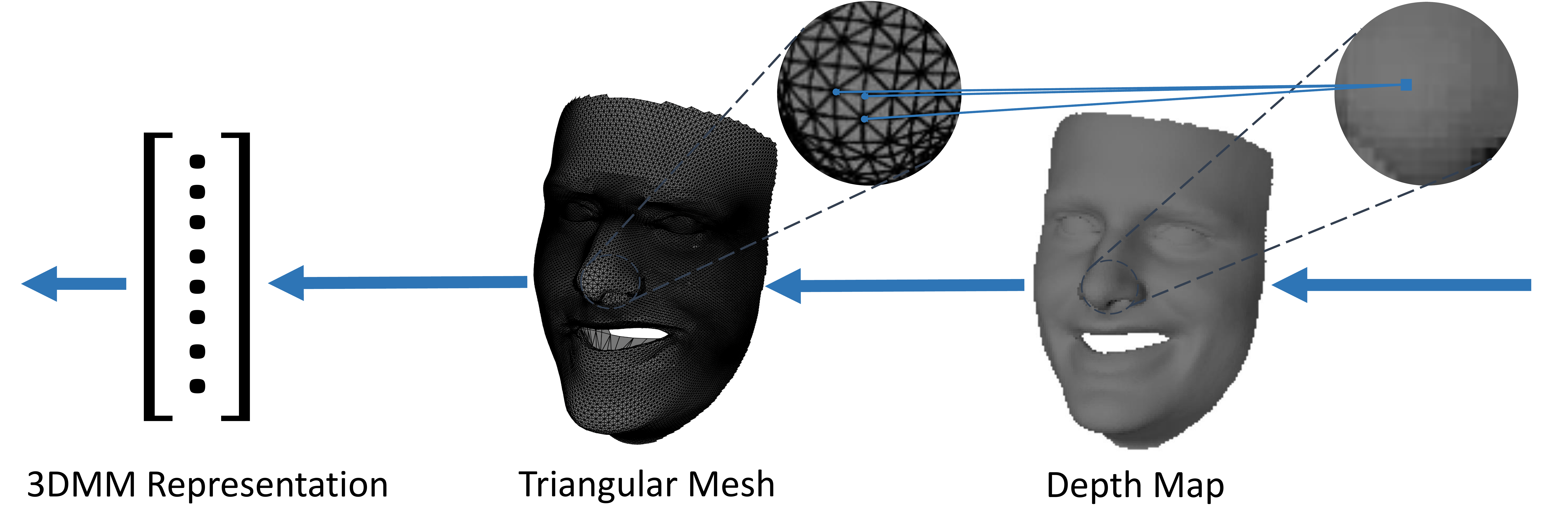}
    \end{subfigure}
    \caption{Gradient flow. Gradients from FineNet are first propagated to the depth map,
     and then propagated from each pixel to the matching vertices.
     The gradients on the triangular mesh are then propagated back to the morphable model representation.}
    \label{fig:gradient-flow}
\end{figure}

To connect CoarseNet with FineNet we propose a novel rendering layer.
The layer receives the geometry and pose representation vector as the input
 and outputs a $200\times 200$ depth map of the geometry in the corresponding pose.
This is done in two steps, first the 3D mesh is calculated from the geometry parameters
 and positioned above the image plane,
\begin{equation}
  \hspace{-2pt}\left[\begin{array}{c}
  p_x\\
  p_y\\
  p_z
  \end{array}\right]=\left[\begin{array}{ccc}
  f & 0 & 0\\
  0 & f & 0\\
  0 & 0 & 1
  \end{array}\right]R\left[A_{\mbox{id}}|A_{\mbox{exp}}\right]\hat{\alpha}+\left[\begin{array}{c}
  t_x\\
  t_y\\
  0
  \end{array}\right].
\label{eq:project}
\end{equation}
The 3D mesh is then rendered using a z-buffer renderer, where each pixel is associated
 with a single triangular face from the mesh.
In order to handle potential occlusions, when a single pixel resides in more than one triangle, the one that is closest to the image
 plane is chosen.
The value of each pixel is determined by interpolating the z-values of
 the mesh face using barycentric coordinates
\begin{equation}
  \tilde{z}=\lambda_{0}z_{0}+\lambda_{1}z_{1}+\lambda_{2}z_{2},
\end{equation}
 where $z_i$ is the z-value of the $i^{th}$ vertex  in the respective triangle
 and $\lambda_i$ is the corresponding coordinate.
During back-propagation the gradients are passed from each pixel to the matching vertex,
 weighted by the corresponding coordinate,
\begin{equation}
  \frac{dE}{dz_i}=\frac{dE}{d\tilde{z}}\frac{d\tilde{z}}{dz_i}=\frac{dE}{d\tilde{z}}\lambda_i,
\end{equation}
where $E$ is the loss criterion.
Note that we assume that the barycentric coordinates are fixed.
Alternatively, one could derive the coordinates with respect to $x_i$ and $y_i$.
Note that no gradients are propagated to hidden vertices since they do not appear in the output depth map.
A similar approach  was applied for example in~\cite{zienkiewiczreal}.
Finally, the gradients are propagated from each vertex back to the geometry basis,
 by taking the derivative of Equation~\ref{eq:project} with respect to $\hat{\alpha}$.
The gradients transfer is visualized in Figure \ref{fig:gradient-flow}.

\subsection{FineNet Framework}
 \label{subsec:finenet_arch}
Delicate facial features such as wrinkles and dimples are difficult to represent by a 3DMM low dimensional
 space, mainly due to their high diversity.
Hence, in contrast to CoarseNet, we need to use a pixel-based framework to recover the fine details.
Recently, several notable pixel-based CNN
      architectures~\cite{goodfellow2014generative,long2015fully,hariharan2015hypercolumns}
  were used for various fine grained tasks like semantic and instance
  segmentation~\cite{long2015fully,hariharan2015hypercolumns},
  optical flow~\cite{Dosovitskiy_2015_ICCV}, and human pose estimation~\cite{Wei_2016_CVPR}.
First successful attempts to reconstruct surface normals using these
 architectures~\cite{Bansal_2016_CVPR,yoon2016fine} have motivated our FineNet architecture.
The proposed framework differs from both these networks in its output (depth map vs. normal map) and
 training regime (unsupervised vs. supervised).

The FineNet is based on the hypercolumn architecture suggested in~\cite{hariharan2015hypercolumns}.
The main idea behind this architecture is to generate a per-pixel feature map which incorporates both
 structural and semantic data.
This is achieved by concatenating the output responses from several convolution layers along the path
 of the network.
Due to pooling layers, the output maps size of inner layers does not match the size of the input image,
 therefore, they are interpolated back to the original size, to create a dense per-pixel volume of features.
This volume is then processed by several $1 \times 1$ convolution layers to create the final prediction.

We choose the VGG-Face~\cite{Parkhi15} as a base for our hypercolumn network since it was fine tuned on
 a domain of faces.
For interpolating, we apply a slightly different scheme than that of \cite{hariharan2015hypercolumns}.
Instead of directly upsampling each feature map to the original size using bilinear interpolation,
 we use cascaded 2-strided $2 \times 2$ upconvolution layers to upsample the feature maps.
This is done in order to improve the quality of the features, as the interpolation is now also
 part of the learning process.
In contrast to recognition problems, refining the facial features is a relatively local problem.
Therefore, we truncate the VGG-Face network before the third pooling layer and form a
 $200 \times 200 \times 450$ hypercolumn feature volume.
This volume is then processed by a set of $1 \times 1$ convolutional layers used as a linear regressor.
Note, that this fully convolutional framework allows us to use any size of input images.
Figure~\ref{fig:end_to_end} describes the FineNet architecture.

\subsection{FineNet Unsupervised Criterion}\label{subsec:unsup_crit}
To train FineNet some form of loss function is required. One possible solution would be to simply
 use an MSE criterion between the network output and a high-quality ground-truth depth map.
This would allow the network to implicitly learn how to reconstruct detailed faces from a single image.
Unfortunately, as mentioned in Section~\ref{subsec:data_acq}, a large dataset of detailed facial
 geometries with their corresponding 2D images is currently unavailable.
Furthermore, a synthetic dataset for this task cannot be generated using morphable models as
 there is no known model that captures the diversity of fine facial details.
Instead, we propose an unsupervised learning process where the loss criterion is determined by
 an axiomatic model.
To achieve that, we need to find a measure that relates the output depth map to the 2D image.
To that end, we resort to Shape from Shading (SfS).

Recent results in SfS~\cite{kemelmacher20113d,yu2013shading,han2013high,or2015rgbd,Or-El_2016_CVPR} have
 shown that when given an initial rough surface, subtle geometry details can be accurately recovered under
 various lighting conditions and multiple surface albedos. This is achieved by optimizing some objective function which ties the geometry to the input image.
In our case, an initial surface is produced by CoarseNet and its depth map representation is fed into FineNet
 along with the input image. We then formulate an unsupervised loss criterion based on the SfS objective function, transforming the problem from an online optimization problem to a regression one.

\subsubsection{From SfS Objective to Unsupervised Loss}
Our unsupervised loss criterion was formulated in the spirit of \cite{or2015rgbd,Or-El_2016_CVPR}.
The core of our loss function is an image formation term, which describes the connection between the
 network's output depth map and its input image.
This term drives the network to learn fine detail recovery and is defined as
\begin{equation}
	E_{sh} = \left\Vert \rho\left\langle \vec{l},\vec{Y}(\hat{z})\right\rangle - I\right\Vert_2^2.
\end{equation}
Here, $\hat{z}$ is the reconstructed depth map, $I$ is the input intensity image, $\rho$ is the albedo
 image, and $\vec{l}$ are the first-order spherical harmonics coefficients.
$Y(\hat{z})$ represents the matching spherical harmonics basis,
 \begin{equation}
   Y(\hat{z})=\left(1,n_{x}(\hat{z}),n_{y}(\hat{z}),n_{z}(\hat{z})\right),
 \end{equation}
 where $(n_{x}(\hat{z}),n_{y}(\hat{z}),n_{z}(\hat{z}))$ is the normal expressed as a function of the depth.
  Notice that while $I$ is an input to FineNet, the scene lighting $\vec{l}$ and albedo map $\rho$ are unknowns. Generally, the need to recover both lighting and albedo is part of the ambiguity in SfS problems. However, here we can utilize the fact we do not solve a general SfS problem, but one constrained to human faces.
This is done by limiting the space of possible albedos to a low dimensional 3DMM texture subspace.
 \begin{equation}
 \rho \approx T = \mu_T + A_T\alpha_T.\,
 \end{equation}
 where $\mu_T$ is the average face texture, $A_T$ is a principal component basis and $\alpha_T$
 is the corresponding coefficients vector.
In our implementation, $10$ coefficients were used.

\begin{figure} %
    \centering
	\includegraphics[width=0.075\textwidth]{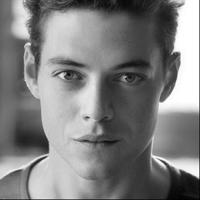}
    \includegraphics[width=0.075\textwidth]{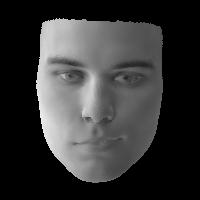}
    \includegraphics[width=0.075\textwidth]{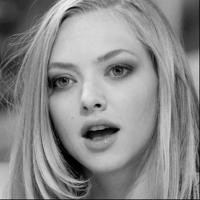}
    \includegraphics[width=0.075\textwidth]{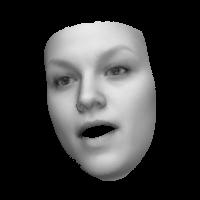}
    \includegraphics[width=0.075\textwidth]{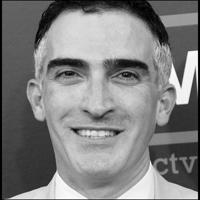}
    \includegraphics[width=0.075\textwidth]{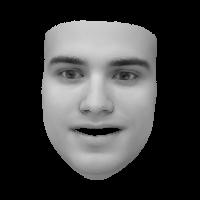}
    \caption{Light and albedo recovery. Images are presented next to the recovered albedo, rendered with
             the recovered lighting.}
      \label{fig:light_albedo}
\end{figure}

Now, as shown in \cite{kemelmacher20113d}, the global lighting can be correctly recovered by assuming
 the average facial albedo, $\hat{\rho}=\mu_T$, using the coarse depth map, $z_0$, as follows
  \begin{equation}
{\vec{l}}^{*} = \underset{\vec{l}}{\operatorname{argmin}}\ \left\Vert\hat{\rho}\left\langle \vec{l},\vec{Y}(z_{0})\right\rangle -I\right\Vert_2^{2}.
  \label{eq:lighting_regression}
  \end{equation}
Note that this is an overdetermined linear problem that can be easily solved using least squares. Given the lighting coefficients, the albedo can also be easily recovered as
\begin{equation}
\alpha_{T}^{*} = \underset{\alpha_{T}}{\operatorname{argmin}}\ \left\Vert\left(\mu_{T}+A_{T}\alpha_{T}\right)\left\langle \vec{l}^{*},\vec{Y}(z_{0})\right\rangle-I\right\Vert_2^{2}.
\label{eq:albedo_regression}
\end{equation}
As in Equation \ref{eq:lighting_regression}, this is an overdetermined linear problem that can be solved directly. Based on the resulting albedo and lighting coefficients we can calculate $E_{sh}$ and its gradient with
 respect to $\hat{z}$.
A few recovery samples are presented in Figure \ref{fig:light_albedo}.
To regularize the solution, fidelity and smoothness terms are added to the criterion of FineNet.
\begin{eqnarray}
 E_f &=& \|\hat{z} - z_0\|_2^2, \cr
 E_{sm} &=& \|\Delta\hat{z}\|_1,
\end{eqnarray}
 where $\Delta$ is the discrete Laplacian operator.
These terms guarantee that the solution would be smooth and would not stray from the prediction of CoarseNet. The final per-pixel loss function is then defined as
\begin{equation}
L(\hat{z},z_{0},I)=\lambda_{sh}E_{sh}(\hat{z},I)+\lambda_{f}E_{f}(\hat{z},z_{0})+\lambda_{sm}E_{sm}(\hat{z}).
\end{equation}
Where the $\lambda$s determine the balance between the terms and were set
 as $\lambda_{sh}=1$, $\lambda_{f}=5e^{-3}$, $\lambda_{sm}=1$.
The gradient of $L$ with respect to $\hat{z}$ is then calculated and used for backpropagation.

 \subsubsection{Unsupervised Loss - a Discussion}
The usage of unsupervised criterion has some desired traits.
First, it eliminates the need for an annotated dataset.
Second, it ensures that the network is not limited by the performance of any algorithm or the
 quality of the dataset.
This results from the fact that the loss function is entirely dependent on the input, in contrast to
 supervised learning SfS schemes such as~\cite{yoon2016fine} and~\cite{Bansal_2016_CVPR}, where the
  data is generated by either photometric stereo or raw Kinect scans, respectively.
 In addition, unlike traditional SfS algorithms, the fact that the albedo and lighting coefficients are
 calculated only as part of the loss function means that at test time the network can produce accurate
   results directly from the intensity and depth inputs, without explicitly calculating the albedo and
    lighting information.
Although the CoarseNet can be trained to generate the lighting and albedo parameters,
 we chose not to include them in the pipeline for two reasons.
First, the lighting and albedo are only needed for the training stage and have no use during testing.
Second, both~\eqref{eq:lighting_regression} and ~\eqref{eq:albedo_regression} are over-determined systems
 which can be solved efficiently with least squares, thus, using a CNN for this task would be redundant.

\subsection{End-to-End Network Training}
Finally, in order to train FineNet, we connect it to CoarseNet using the proposed rendering layer which is added
 between the two networks.
Thus, a single end-to-end network is created.
We then use images from the VGG face dataset \cite{Parkhi15}, and propagate them through the framework.
The forward pass can be divided into three main steps.
First, each such image is propagated through CoarseNet for four iterations, creating the coarse
 geometry representation.
Then, the rendering layer transforms the 3DMM representation to a depth map.
Finally, the depth map, alongside the original input image, is propagated through FineNet resulting
 in the dense updated depth map.
The criterion presented in \ref{subsec:unsup_crit} is then used to calculate the loss gradient.
The gradient is backpropagated through the network allowing us to train FineNet and fine-tune CoarseNet.

\begin{figure}
    \centering
    \begin{subfigure}[b]{0.48\textwidth}
      \includegraphics[width=\textwidth]{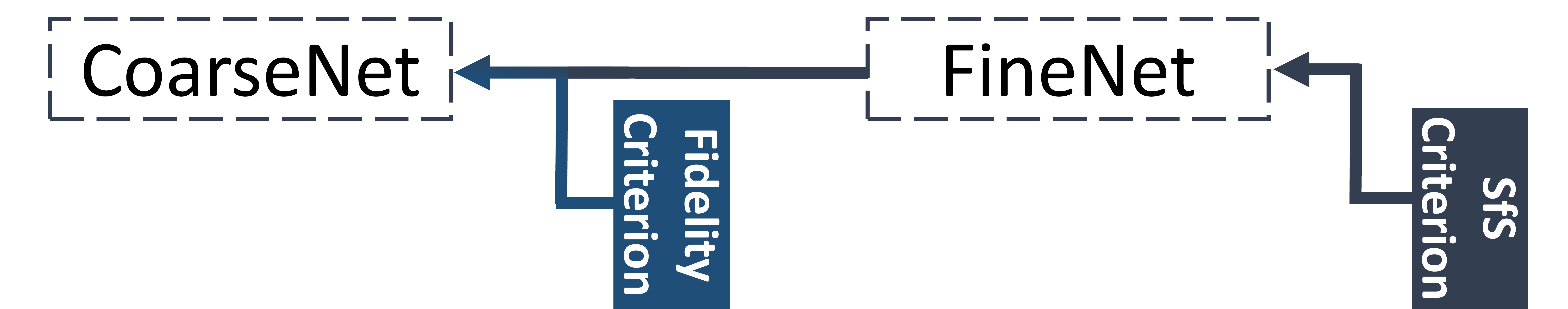}
    \end{subfigure}
    \caption{Criterion flow.
         Gradients from both loss criteria are propagated back to CoarseNet.}
    \label{fig:criterion-flow}
    \vspace{-0.5cm}
\end{figure}

Note that the fact that CoarseNet was already trained is crucial for a successful training.
This stems from the fact that the unsupervised loss function depends on the coarse
 initialization, which cannot be achieved without the synthetic data.
In order to prevent CoarseNet from deviating too much from the original coarse solution, a fidelity criterion
 is added to CoarseNet's output.
This criterion is the MSE between the current CoarseNet solution and the original one.
Gradients from both FineNet and the fidelity loss are then weighted and passed through CoarseNet,
 fine-tuning it, as presented in Figure \ref{fig:criterion-flow}.

\section{Experiments} \label{sec: experiments}
In order to evaluate the proposed framework we performed several experiments to test its accuracy on both 3D facial datasets and \textit{in the wild} inputs. Both qualitative and quantitative evaluations are used to demonstrate the strength of the proposed solution.
Our method is compared to the template based method of~\cite{kemelmacher20113d}, to the 3DMM based method introduced as part of~\cite{zhu2015high} and to the data driven method of~\cite{richardson20163d}. Note that unlike our method, all of the above require alignment information. We use the state-of-the-art alignment method of \cite{Kazemi_2014_CVPR} to provide input for these algorithms.

\begin{figure}[b] %
\vspace{-0.7cm}
    \centering
    \begin{subfigure}[b]{.091 \textwidth}
  \includegraphics[width=\textwidth]{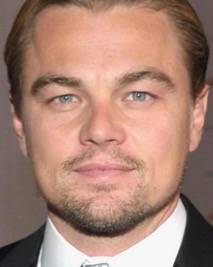}
  \caption{}
  \label{fig:im1}
  \end{subfigure}
  \begin{subfigure}[b]{.092 \textwidth}
  \includegraphics[width=\textwidth]{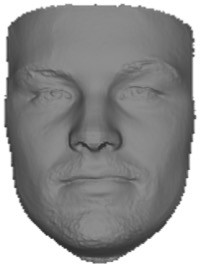}
  \caption{}
  \label{fig:rec1}
  \end{subfigure}
  \begin{subfigure}[b]{.083 \textwidth}
  \includegraphics[width=\textwidth]{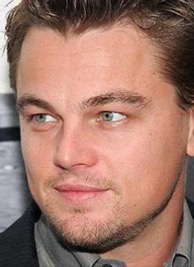}
  \caption{}
  \label{fig:im2}
  \end{subfigure}
  \begin{subfigure}[b]{.092 \textwidth}
  \includegraphics[width=\textwidth]{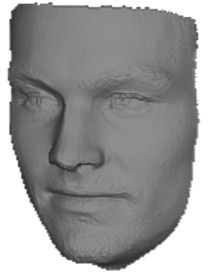}
  \caption{}
  \label{fig:rec2}
  \end{subfigure}
  \begin{subfigure}[b]{.096 \textwidth}
  \includegraphics[width=\textwidth]{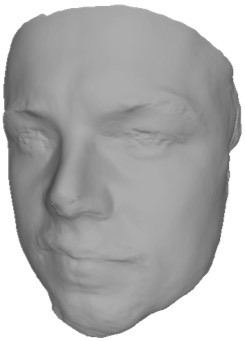}
  \caption{}
  \label{fig:unconst}
  \end{subfigure}
\vspace{-0.3cm}
  \caption{(\subref{fig:im1}) and (\subref{fig:im2}) are two input images, (\subref{fig:rec1})
    and (\subref{fig:rec2}) are their 3D reconstruction via the proposed method.
   (\subref{fig:unconst}) is a reconstruction of the same subject, based on $100$
    different images recovered with the method proposed in \cite{roth2016adaptive}. }
  \label{fig:ours_vs_multiple}
\vspace{-0.2cm}
\end{figure}

\begin{figure}[b] %
    \centering
	\includegraphics[height=0.09\textwidth]{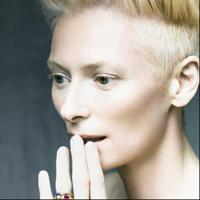}
    \includegraphics[height=0.09\textwidth]{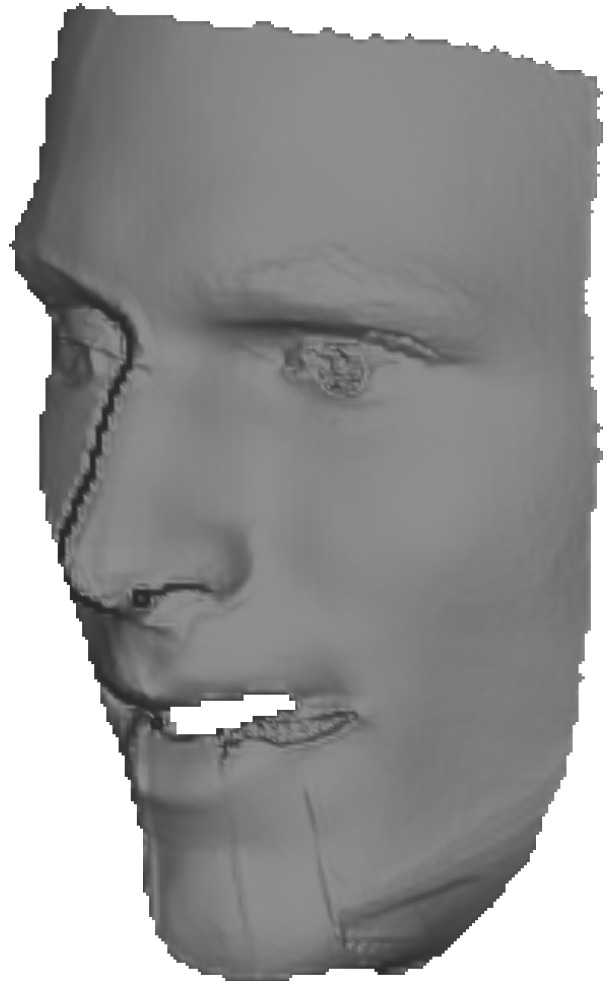}
    \includegraphics[height=0.09\textwidth]{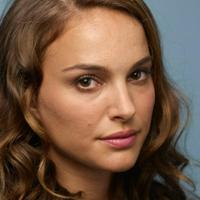}
    \includegraphics[height=0.09\textwidth]{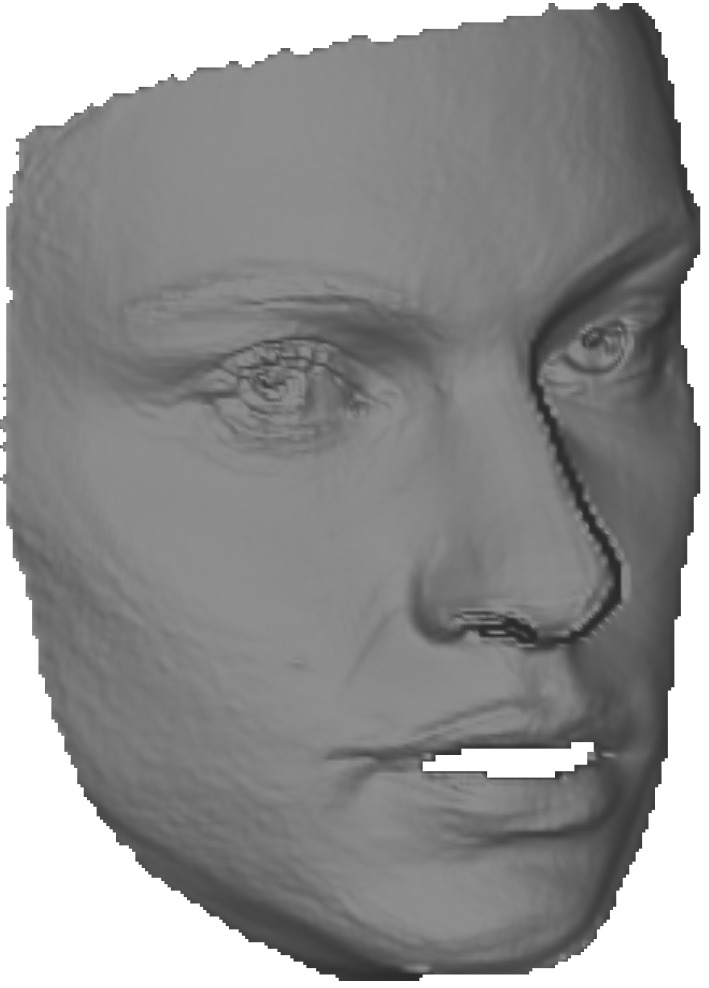}
    \includegraphics[height=0.09\textwidth]{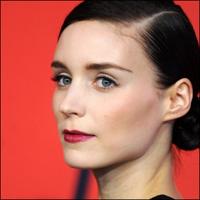}
    \includegraphics[height=0.09\textwidth]{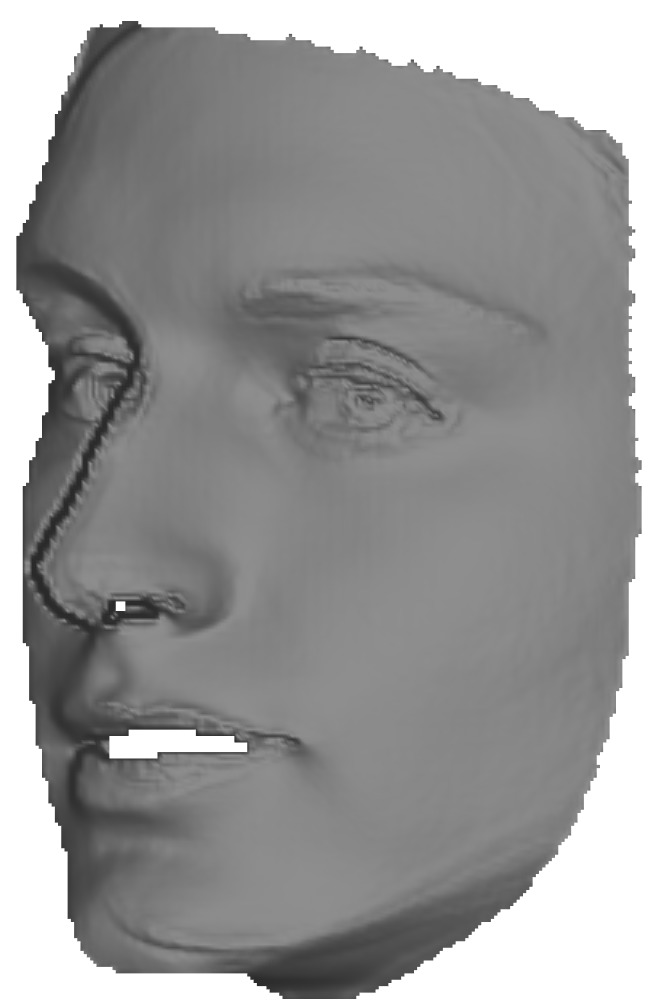}
\vspace{-0.2cm}
    \caption{Method robustness. Our method shows some robustness to extreme orientations, even in nearly $90^{\circ}$ angles.}
    \label{fig:robustness}
\end{figure}

\begin{figure*} %
    \centering
    \begin{tabular}{ccccccccc}
 \includegraphics[height=0.12\textwidth]{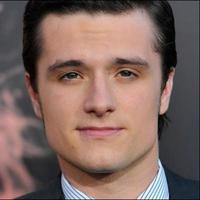}&
 \includegraphics[height=0.12\textwidth]{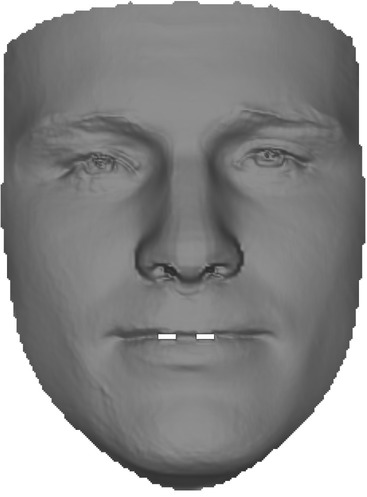}&
 \includegraphics[height=0.12\textwidth]{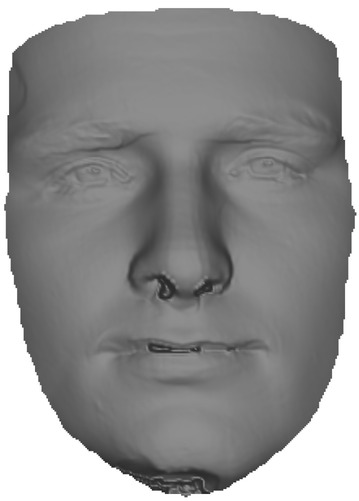}&
 \includegraphics[height=0.12\textwidth]{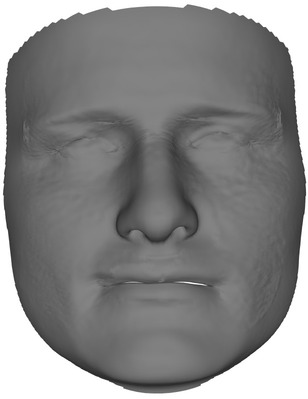}&
 \includegraphics[height=0.12\textwidth]{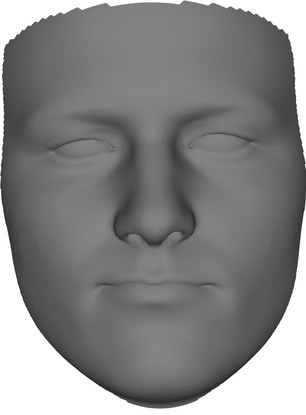}&
 \includegraphics[height=0.12\textwidth]{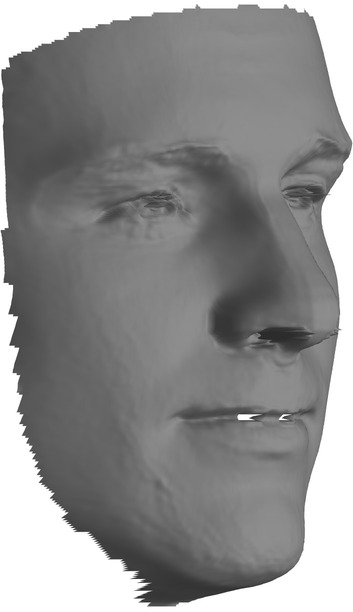}&
 \includegraphics[height=0.12\textwidth]{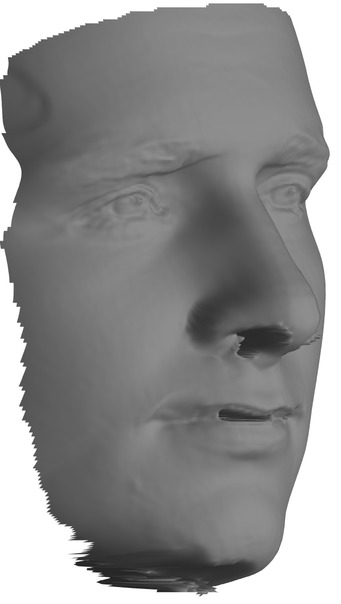}&
 \includegraphics[height=0.12\textwidth]{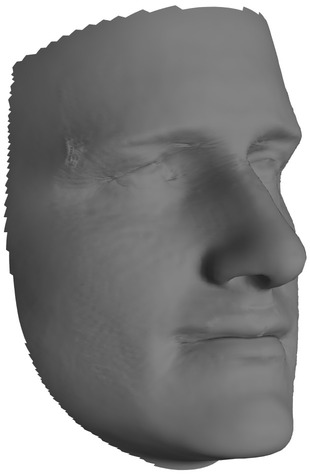}&
 \includegraphics[height=0.12\textwidth]{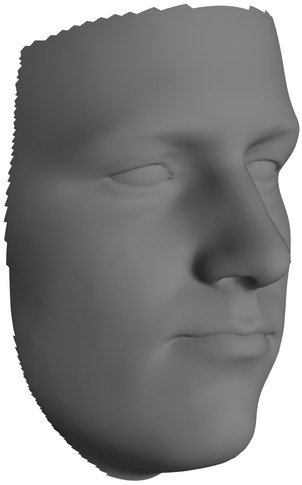}\tabularnewline
 \includegraphics[height=0.12\textwidth]{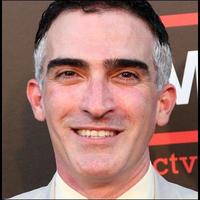}&
 \includegraphics[height=0.12\textwidth]{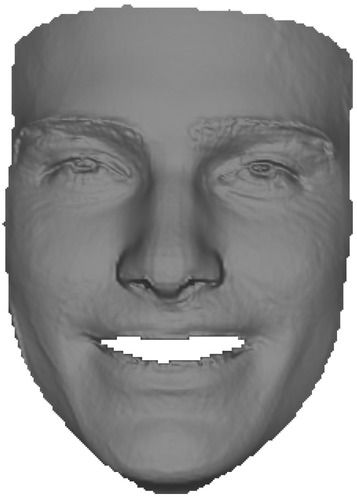}&
 \includegraphics[height=0.12\textwidth]{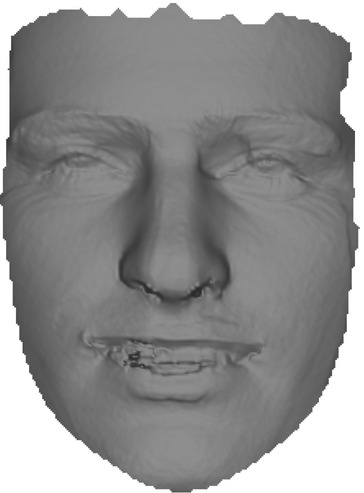}&
 \includegraphics[height=0.12\textwidth]{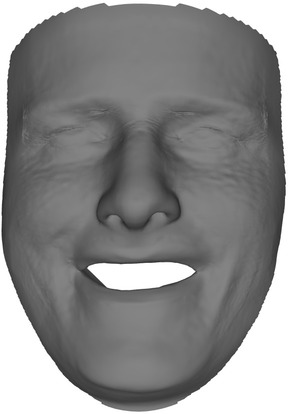}&
 \includegraphics[height=0.12\textwidth]{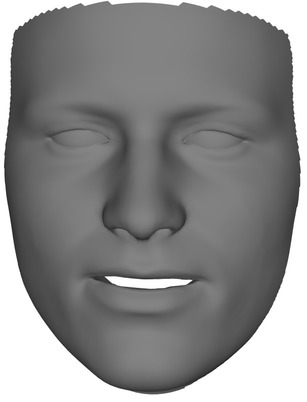}&
 \includegraphics[height=0.12\textwidth]{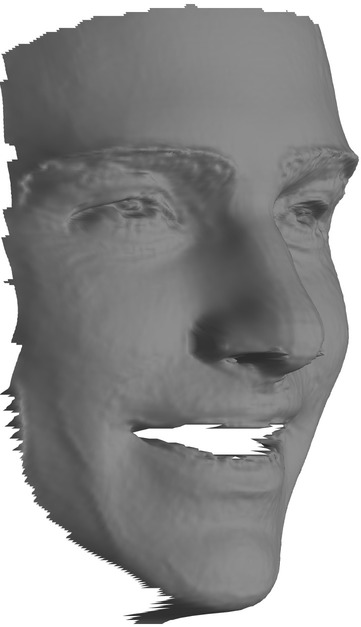}&
 \includegraphics[height=0.12\textwidth]{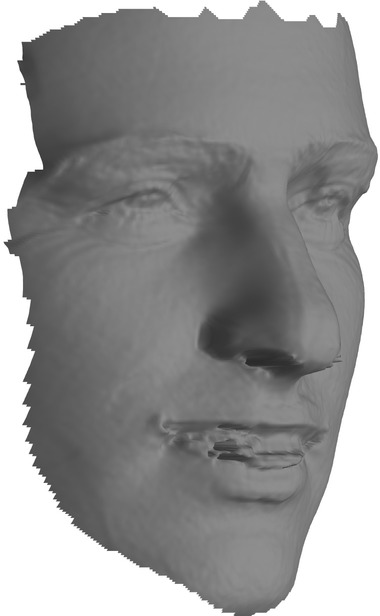}&
 \includegraphics[height=0.12\textwidth]{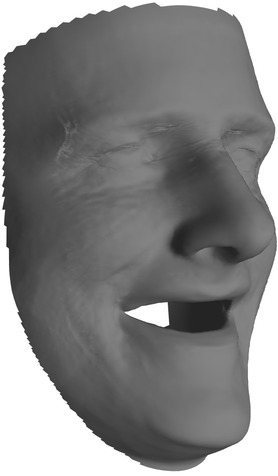}&
 \includegraphics[height=0.12\textwidth]{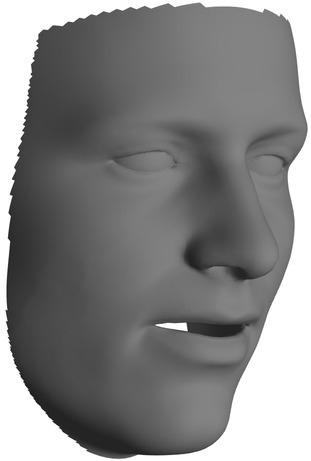}\tabularnewline
 \includegraphics[height=0.12\textwidth]{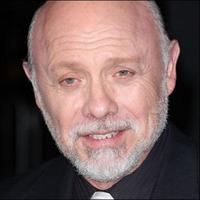}&
 \includegraphics[height=0.12\textwidth]{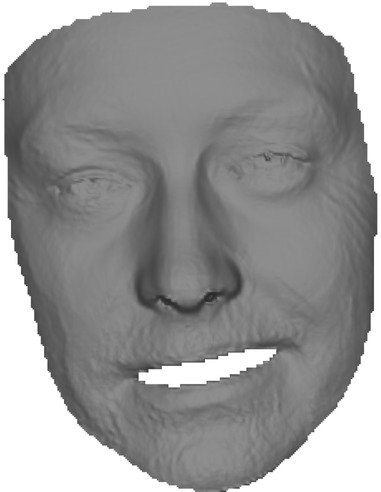}&
 \includegraphics[height=0.12\textwidth]{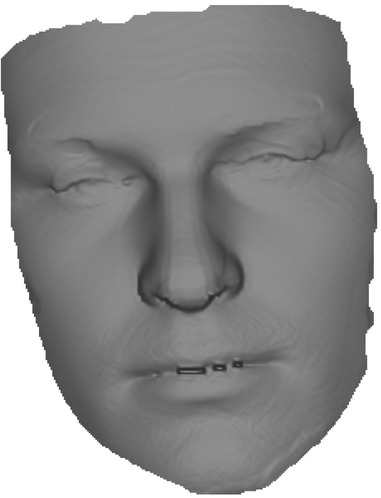}&
 \includegraphics[height=0.12\textwidth]{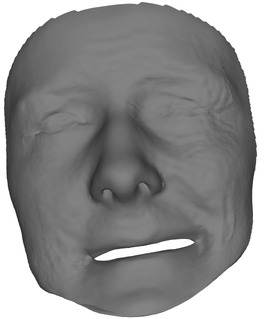}&
 \includegraphics[height=0.12\textwidth]{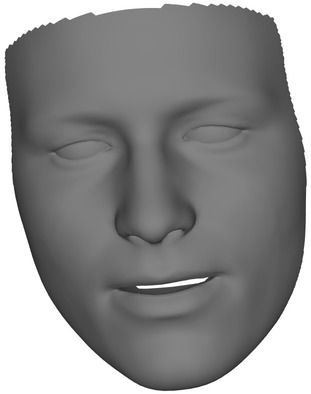}&
 \includegraphics[height=0.12\textwidth]{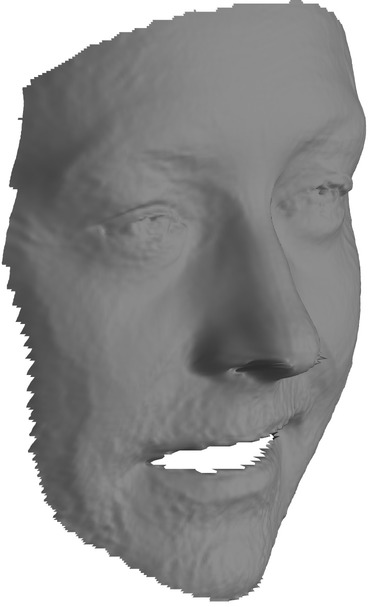}&
 \includegraphics[height=0.12\textwidth]{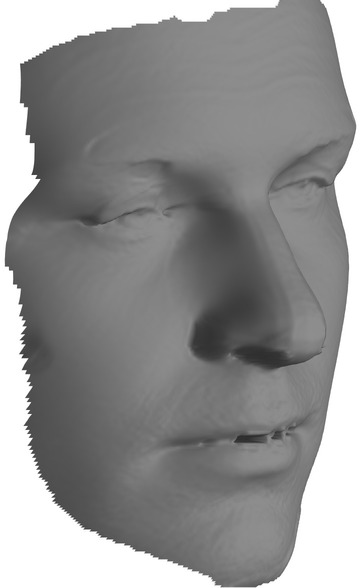}&
 \includegraphics[height=0.12\textwidth]{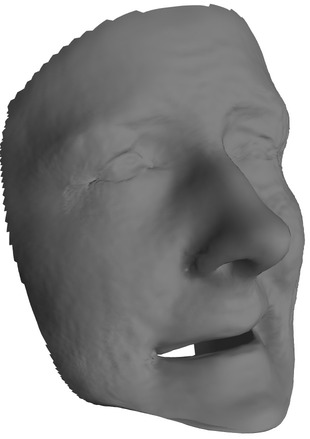}&
 \includegraphics[height=0.12\textwidth]{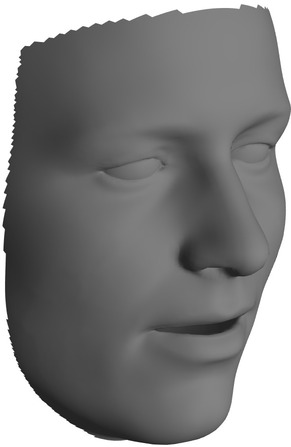}\tabularnewline
 \includegraphics[height=0.12\textwidth]{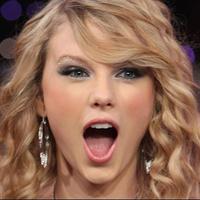}&
 \includegraphics[height=0.12\textwidth]{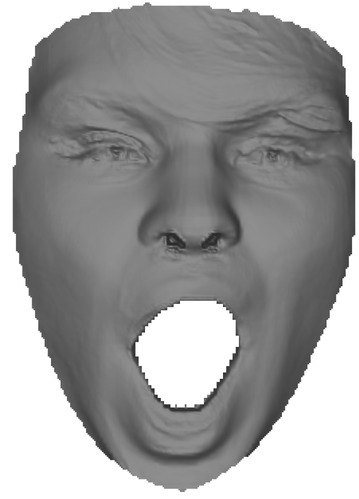}&
 \includegraphics[height=0.12\textwidth]{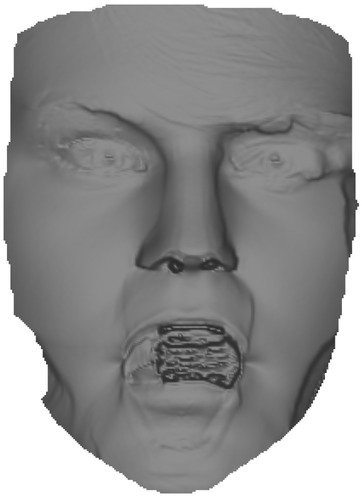}&
 \includegraphics[height=0.12\textwidth]{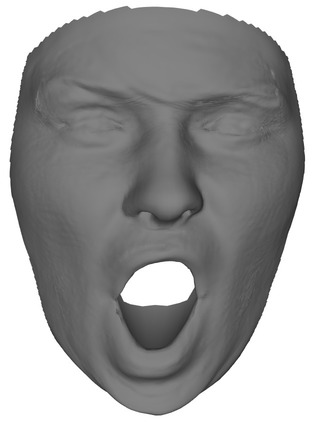}&
 \includegraphics[height=0.12\textwidth]{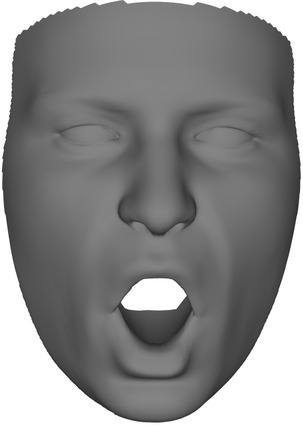}&
 \includegraphics[height=0.12\textwidth]{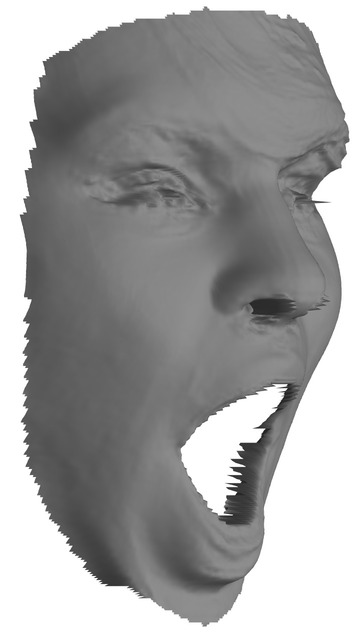}&
 \includegraphics[height=0.12\textwidth]{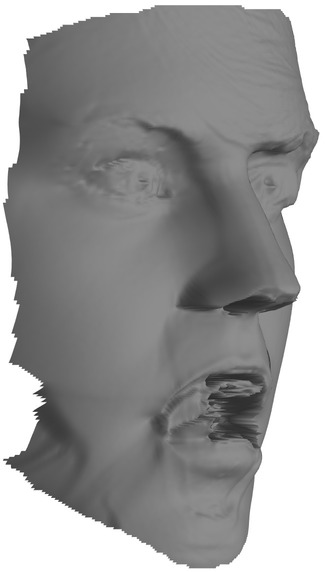}&
 \includegraphics[height=0.12\textwidth]{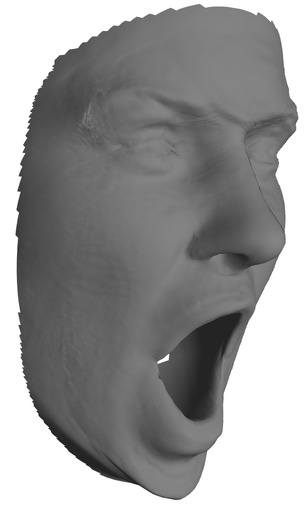}&
 \includegraphics[height=0.12\textwidth]{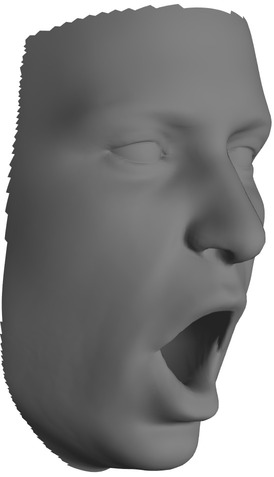}\tabularnewline
 Input & Ours & \cite{kemelmacher20113d} & \cite{richardson20163d} & \cite{zhu2015high} & Ours & \cite{kemelmacher20113d} & \cite{richardson20163d} & \cite{zhu2015high}
\end{tabular}
    \caption{Qualitative results. Input images are presented alongside the reconstruction results of different methods from two different viewpoints. Note that unlike the other methods, the proposed approach is robust to pose and expression variations, while still capturing subtle facial details. }
    \label{fig:vis_example2}
\end{figure*}

For a qualitative analysis we show our results on $400\times400$ \textit{in-the-wild} images of faces.
As can be seen in Figure~\ref{fig:vis_example2}, our method exposes the fine facial details as
 opposed to~\cite{zhu2015high,richardson20163d} and is more robust to expressions and different
 poses than~\cite{kemelmacher20113d}.
In addition, we compare our reconstructions with a state of the art method for reconstruction from multiple
 images~\cite{roth2016adaptive}.
The results are shown in Figure~\ref{fig:ours_vs_multiple}, one can see that our method is able to
 produce a comparable high quality geometry from only a single image.
Finally, Figure~\ref{fig:robustness} shows our method robustness to different poses,
 while Figure~\ref{fig:teaser} shows some more reconstruction results.

For a quantitative analysis of our results we used the Face Recognition Grand Challenge dataset V2 \cite{phillips2005overview}.
This dataset consists of roughly two thousand color facial images aligned with ground truth depth of each pixel.
Each method provided an estimated depth image and a binary mask representing the valid pixels.
For the purpose of fair judgment, we evaluated the accuracy of each method on pixels which were denoted as valid by all the methods.
As shown in Table~\ref{tbl:real_data}, our method produce the lowest depth error among the tested methods.

\begin{figure}
\centering
\begin{subfigure}{0.11\textwidth}
  \includegraphics[width=\textwidth]{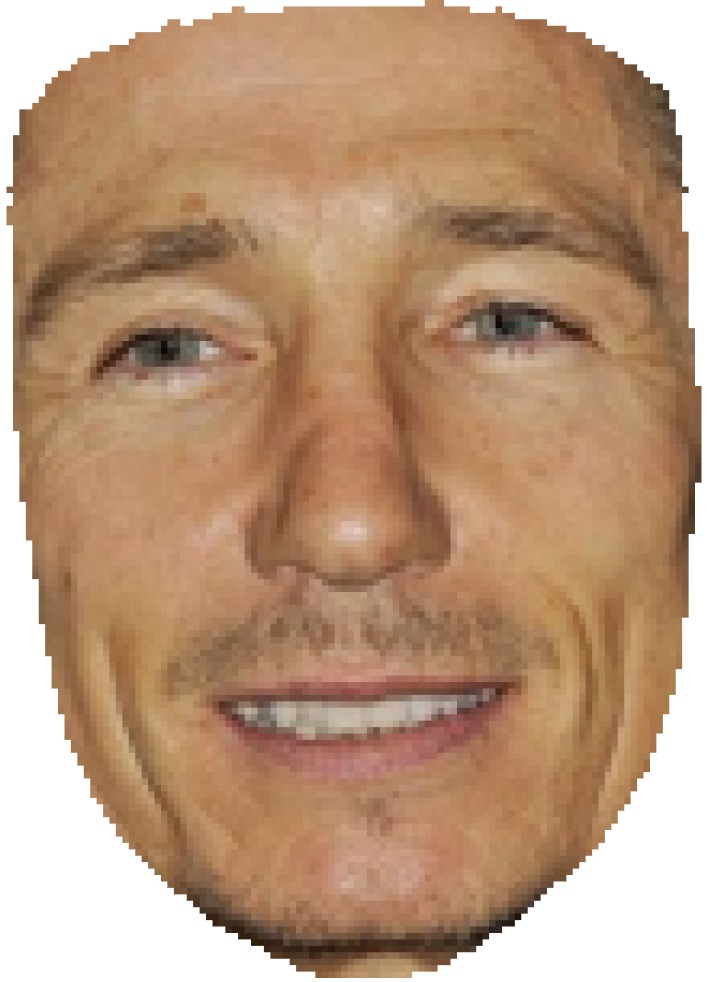}
  \caption{}
  \label{subfig:scale-color}
\end{subfigure}
\begin{subfigure}{0.11\textwidth}
  \includegraphics[width=\textwidth]{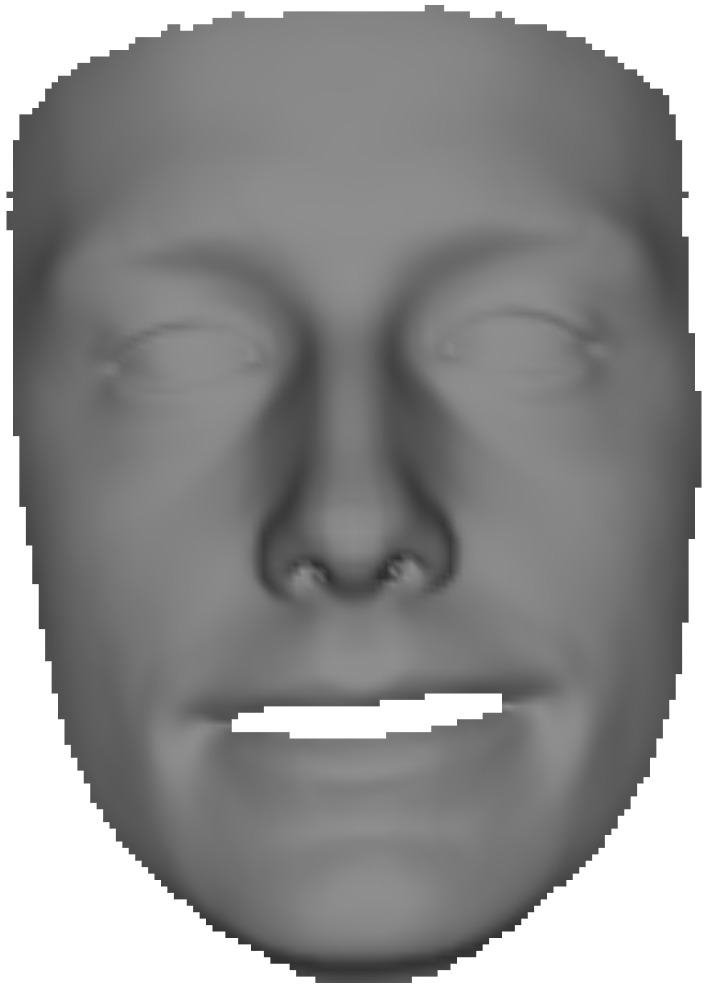}
  \caption{}
   \label{subfig:scale-coarse}
\end{subfigure}
\begin{subfigure}{0.11\textwidth}
  \includegraphics[width=\textwidth]{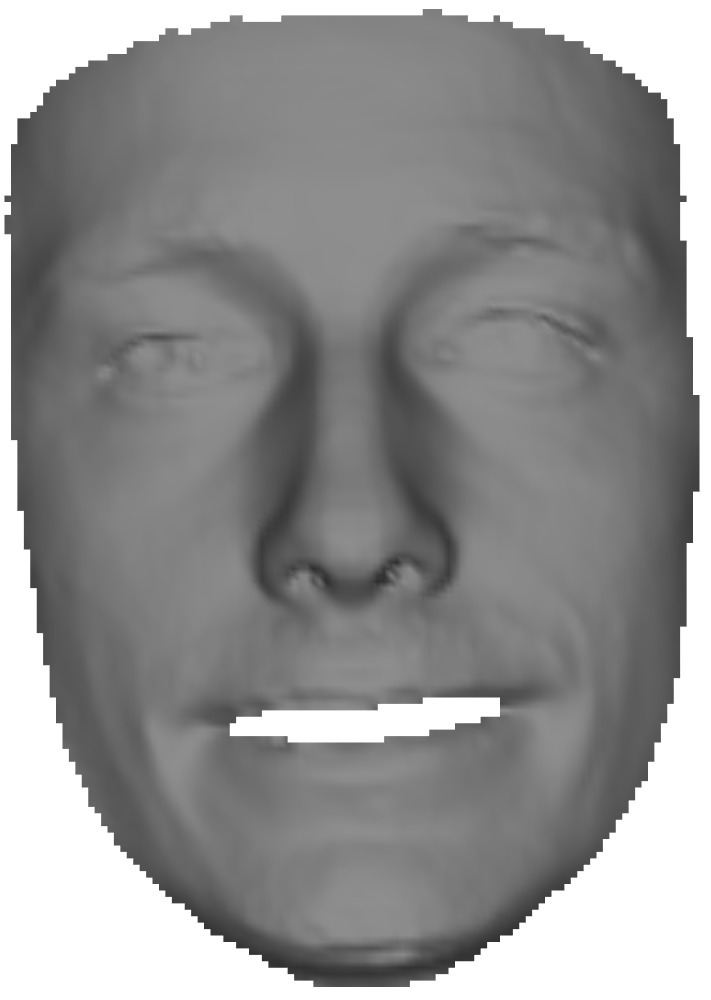}
  \caption{}
  \label{subfig:scale-low}
\end{subfigure}
\begin{subfigure}{0.11\textwidth}
  \includegraphics[width=\textwidth]{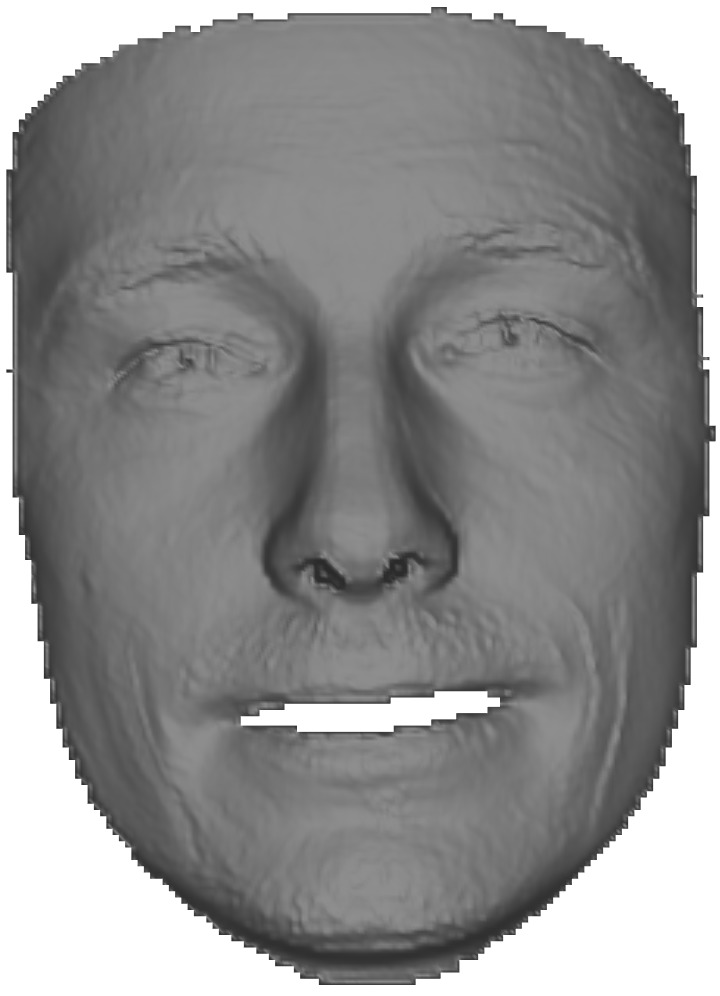}
  \caption{}
  \label{subfig:scale-high}
\end{subfigure}
\caption{Input scaling. (\subref{subfig:scale-color}) is the input image and (\subref{subfig:scale-coarse}) is the coarse depth map from CoarseNet. In (\subref{subfig:scale-low}) the output of FineNet for a $200\times200$ input is presented, while in (\subref{subfig:scale-high}) a $400\times400$ input is used.}
\label{fig:input_size_ablation}
\vspace{-0.55cm}
\end{figure}

Finally, as noted in Section~\ref{subsec:finenet_arch} the fully convolutional FineNet can receive inputs with varying sizes. This size invariance is a vital property for our detail extraction network, as it allows the network to extract more details when a high quality input image is available.  Figure~\ref{fig:input_size_ablation} shows that although our network was trained only on $200\times200$ images it gracefully scales up for $400\times400$ inputs.

\section{Discussion}\label{sec: discussion}
The proposed framework separated the training process into two phases, starting with the training of CoarseNet using synthetic data. While using artificial data allows us to gather the large amounts of data required for training, it does present some limitations in terms of generalization. For example, we found that our network might fail when tested upon unique facial features that were not part of the training data, such as beards, makeup, and glasses, as can be seen in the supplementary material. The second phase of the training is the unsupervised end-to-end training scheme. While we found that this step successfully trains FineNet, it only slightly tunes CoarseNet. We believe that is because the loss function of FineNet is more sensitive to high frequencies, while the 3DMM model captures mainly coarse facial geometries.
Still, it would be interesting to see whether one can push the idea of end-to-end training further, to significantly affect CoarseNet and possibly even to remove its dependency on synthetic data.

\begin{table}
\renewcommand{\arraystretch}{1.1}
\centering
\begin{tabular}{| c || c | c | c  |}
\hline
Method  & Ave. Depth Err. [mm] & 90\% Depth Err. [mm] \\
\hline
\hline
Ours & \textbf{3.22}  & \textbf{6.69}  \\
\hline
\cite{kemelmacher20113d} & 3.33  & 7.02  \\
\hline
\cite{richardson20163d} & 4.11  & 8.70 \\
\hline
\cite{zhu2015high} & 3.46  & 7.36 \\
\hline
\end{tabular}
\vspace{-0.2cm}
\caption {Quantitative comparison. Depth estimation errors of the different methods are presented.}
\label{tbl:real_data}
\vspace{-0.5cm}
\end{table}

\section{Conclusion} \label{sec: conclusion}
We proposed an end-to-end approach for detailed face reconstruction from a single image.
The method is comprised of two main blocks, a network for recovering a rough estimation of the face geometry
 followed by a fine details reconstruction network.
While the former is trained with synthetic images, the latter is trained with real facial images in an end-to-end
 unsupervised training scheme. To connect the two networks a differentiable rendering layer is introduced.
As demonstrated by our comparisons, the proposed framework outperforms recent state-of-the-art approaches.

\vspace{-0.3cm}
\subsubsection*{Acknowledgments}
\vspace{-0.1cm}

Research leading to these results was supported by European Community's
FP7- ERC program, grant agreement no. 267414.

{\small
\bibliographystyle{ieee}
\bibliography{egbib}
}

\newpage
\onecolumn
\appendix
\title{Supplementary Material}
\date{}
\author{}
\makeatletter
\renewcommand{\@maketitle}{\@oldmaketitle}
\maketitle

\section{Supplementary Qualitative Results}
\begin{figure*}[h] %
    \centering
    \begin{tabular}{ccccccccc}
 \includegraphics[height=0.12\textwidth]{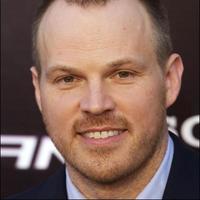}&
 \includegraphics[height=0.12\textwidth]{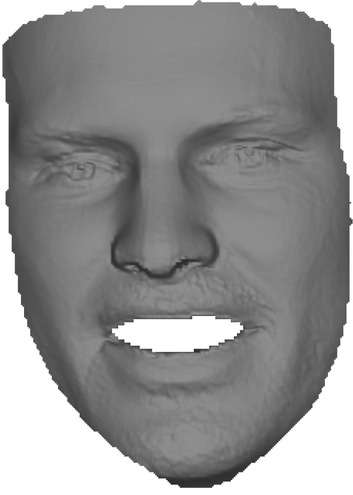}&
 \includegraphics[height=0.12\textwidth]{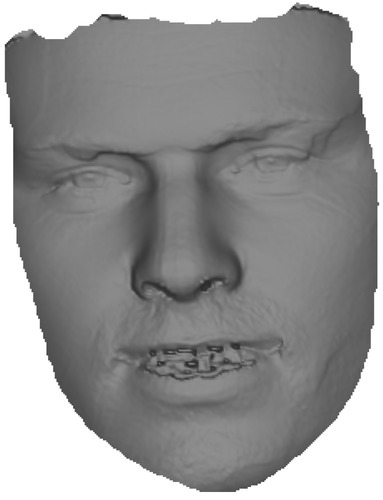}&
 \includegraphics[height=0.12\textwidth]{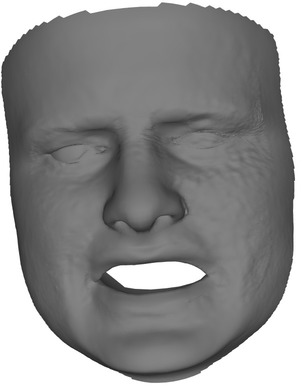}&
 \includegraphics[height=0.12\textwidth]{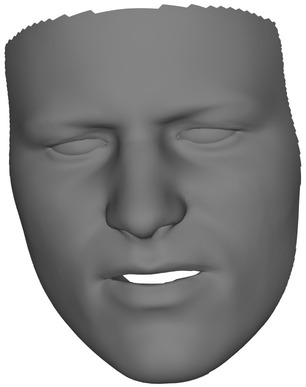}&
 \includegraphics[height=0.12\textwidth]{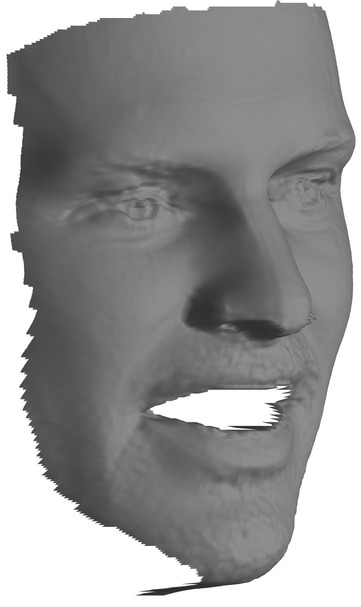}&
 \includegraphics[height=0.12\textwidth]{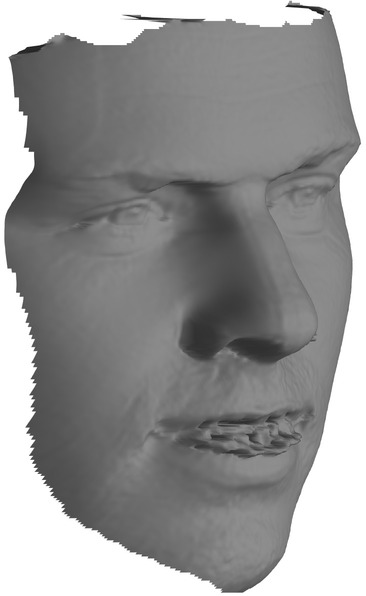}&
 \includegraphics[height=0.12\textwidth]{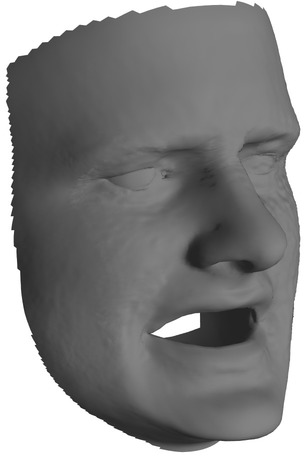}&
 \includegraphics[height=0.12\textwidth]{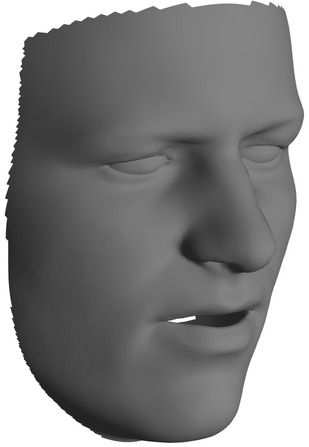}\tabularnewline
 \includegraphics[height=0.12\textwidth]{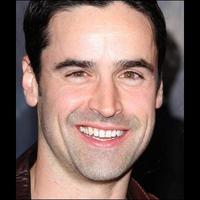}&
 \includegraphics[height=0.12\textwidth]{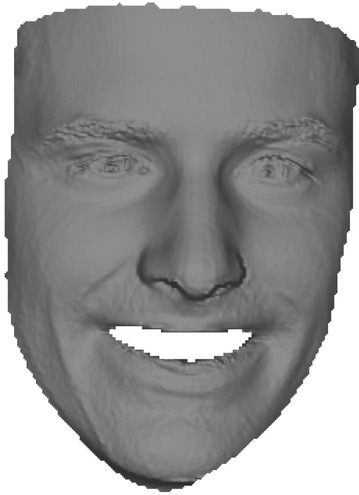}&
 \includegraphics[height=0.12\textwidth]{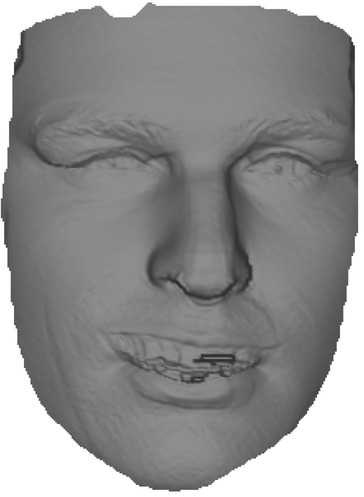}&
 \includegraphics[height=0.12\textwidth]{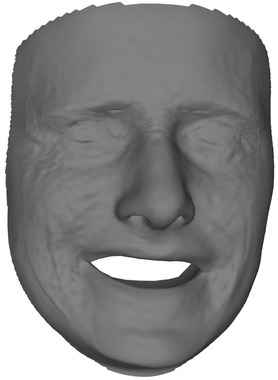}&
 \includegraphics[height=0.12\textwidth]{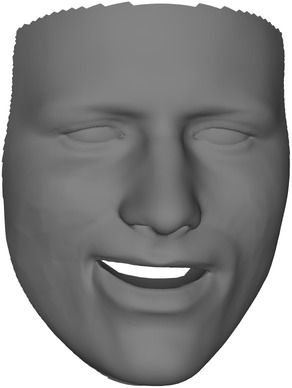}&
 \includegraphics[height=0.12\textwidth]{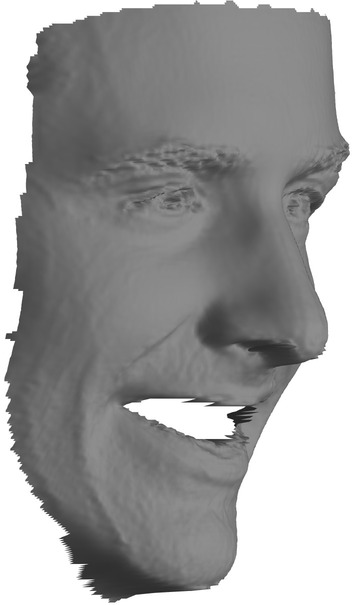}&
 \includegraphics[height=0.12\textwidth]{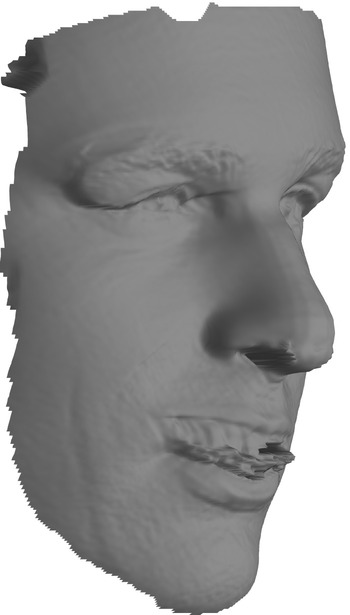}&
 \includegraphics[height=0.12\textwidth]{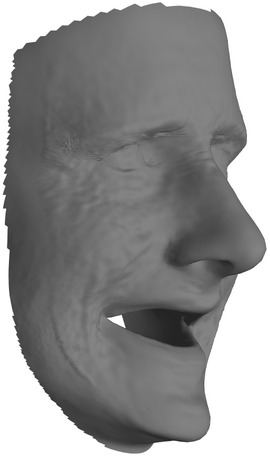}&
 \includegraphics[height=0.12\textwidth]{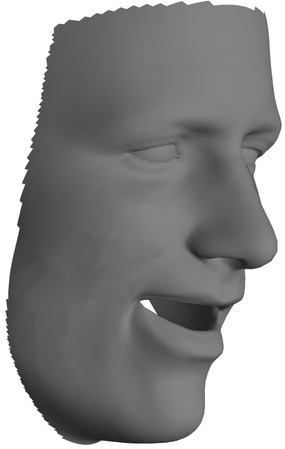}\tabularnewline
 \includegraphics[height=0.12\textwidth]{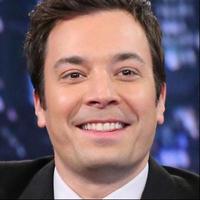}&
 \includegraphics[height=0.12\textwidth]{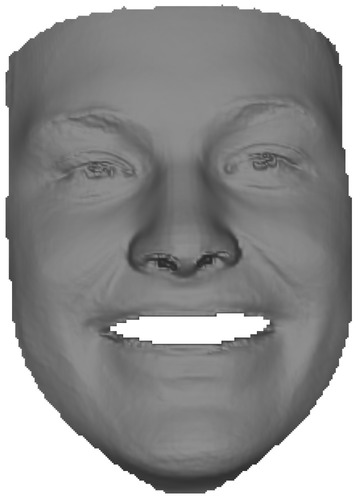}&
 \includegraphics[height=0.12\textwidth]{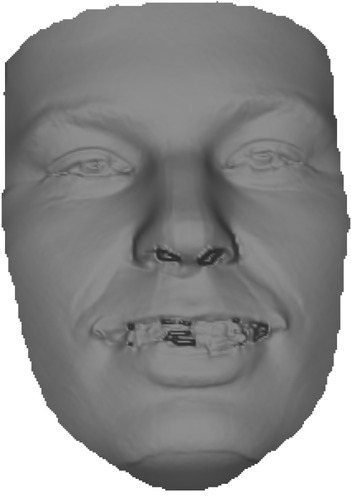}&
 \includegraphics[height=0.12\textwidth]{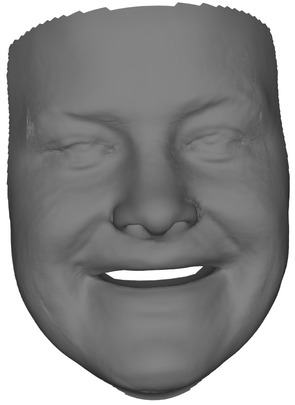}&
 \includegraphics[height=0.12\textwidth]{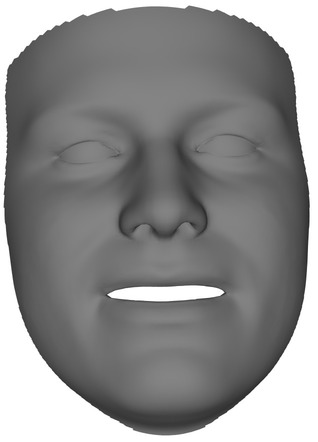}&
 \includegraphics[height=0.12\textwidth]{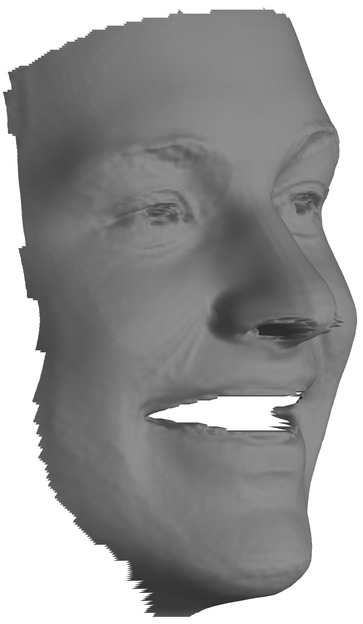}&
 \includegraphics[height=0.12\textwidth]{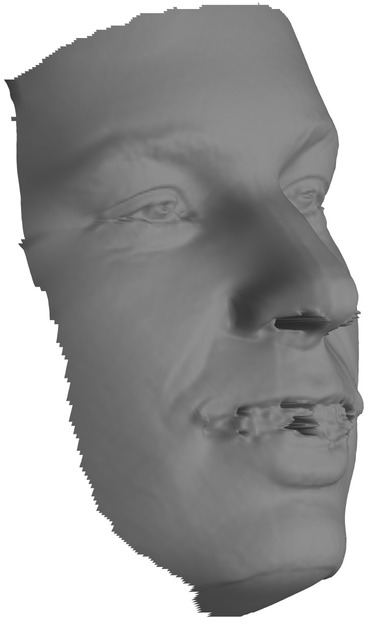}&
 \includegraphics[height=0.12\textwidth]{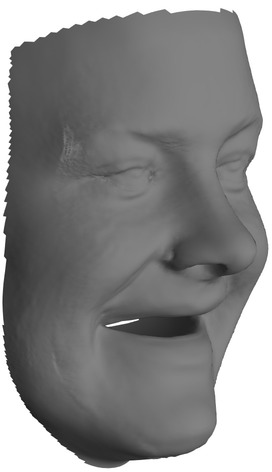}&
 \includegraphics[height=0.12\textwidth]{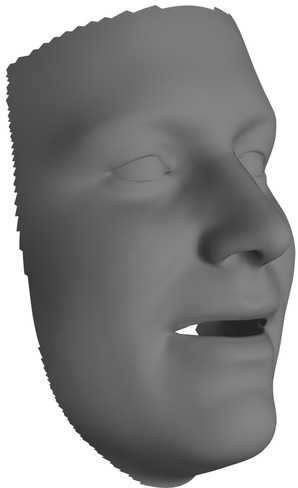}\tabularnewline
 \includegraphics[height=0.12\textwidth]{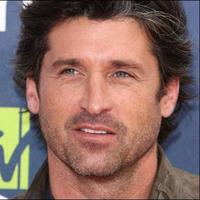}&
 \includegraphics[height=0.12\textwidth]{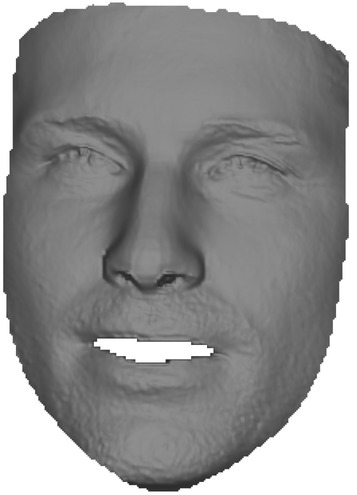}&
 \includegraphics[height=0.12\textwidth]{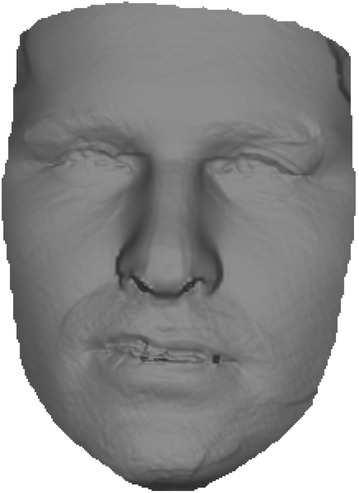}&
 \includegraphics[height=0.12\textwidth]{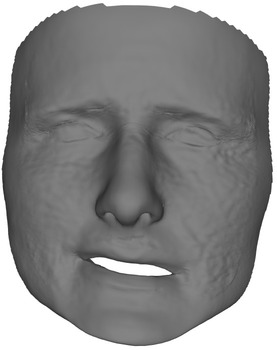}&
 \includegraphics[height=0.12\textwidth]{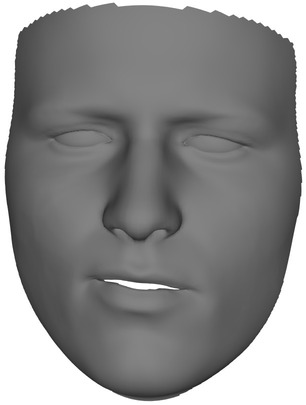}&
 \includegraphics[height=0.12\textwidth]{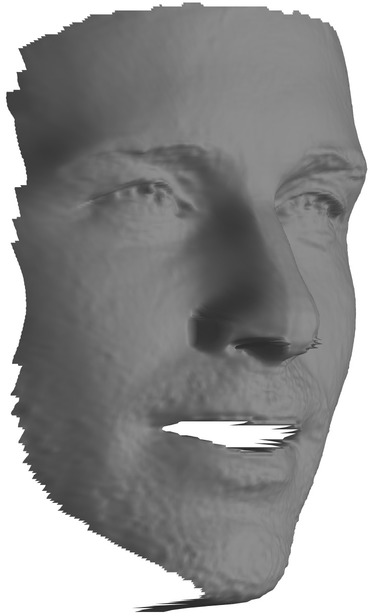}&
 \includegraphics[height=0.12\textwidth]{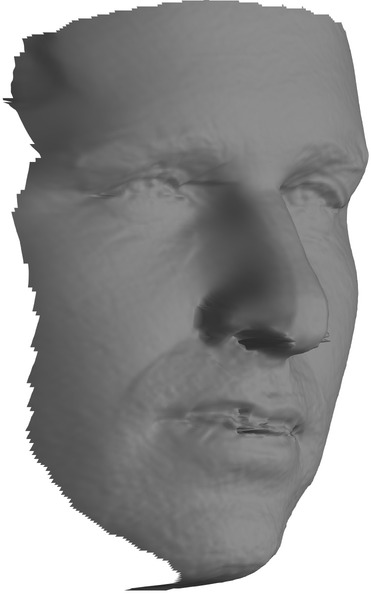}&
 \includegraphics[height=0.12\textwidth]{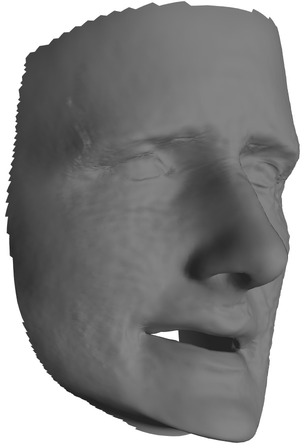}&
 \includegraphics[height=0.12\textwidth]{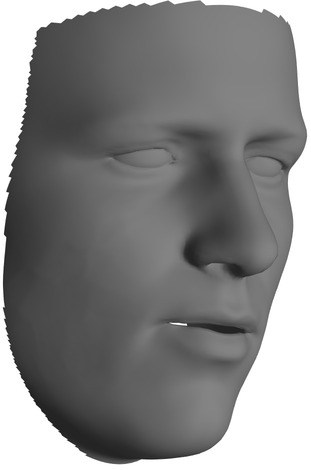}\tabularnewline
 \includegraphics[height=0.12\textwidth]{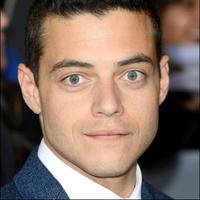}&
 \includegraphics[height=0.12\textwidth]{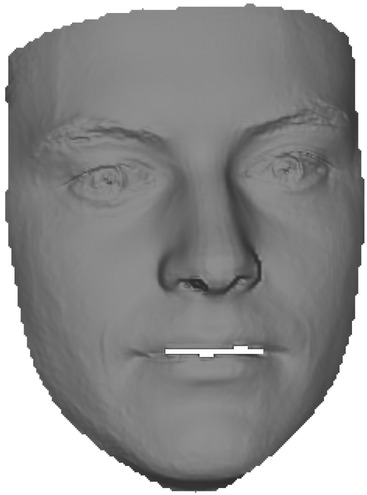}&
 \includegraphics[height=0.12\textwidth]{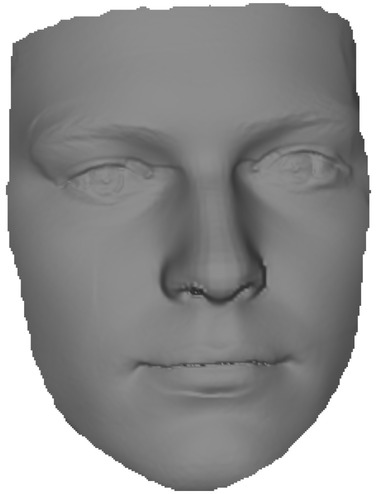}&
 \includegraphics[height=0.12\textwidth]{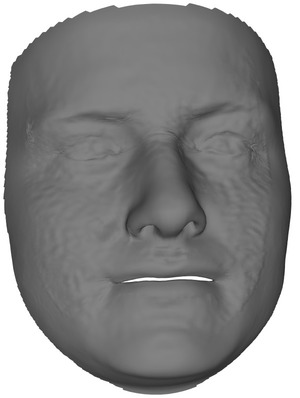}&
 \includegraphics[height=0.12\textwidth]{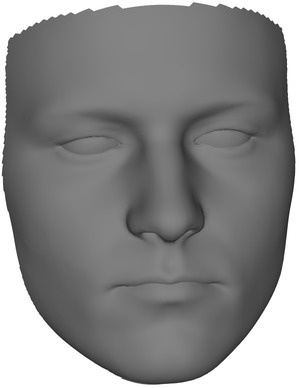}&
 \includegraphics[height=0.12\textwidth]{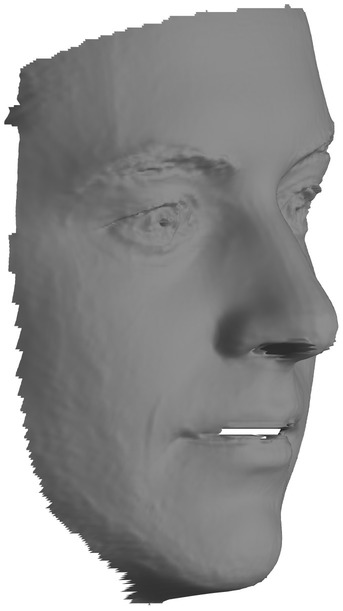}&
 \includegraphics[height=0.12\textwidth]{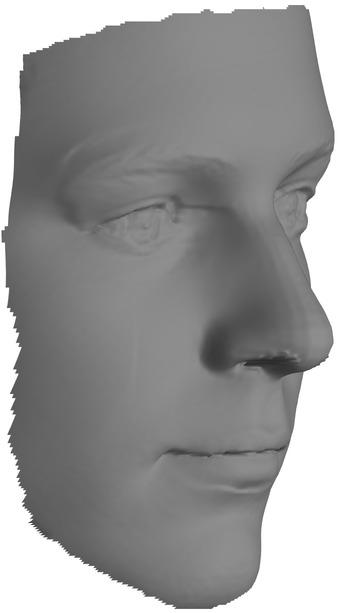}&
 \includegraphics[height=0.12\textwidth]{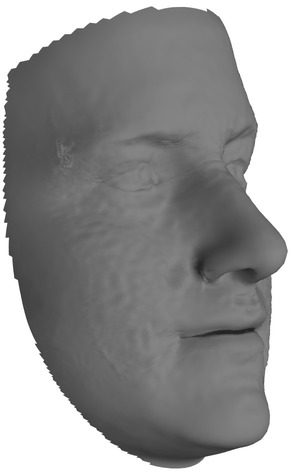}&
 \includegraphics[height=0.12\textwidth]{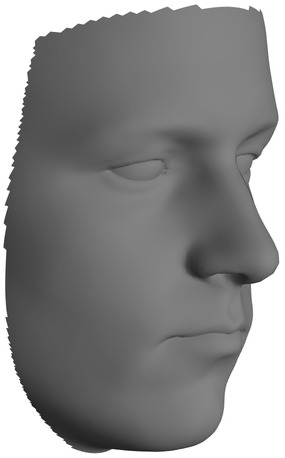}\tabularnewline
 \includegraphics[height=0.12\textwidth]{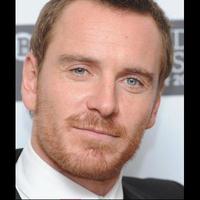}&
 \includegraphics[height=0.12\textwidth]{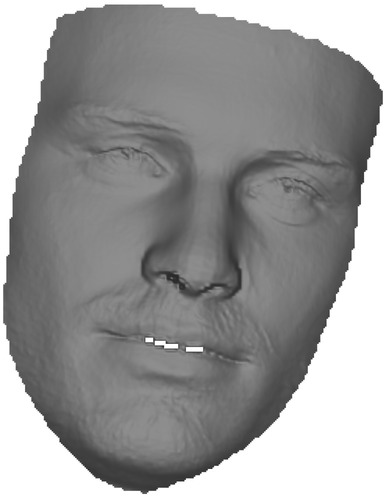}&
 \includegraphics[height=0.12\textwidth]{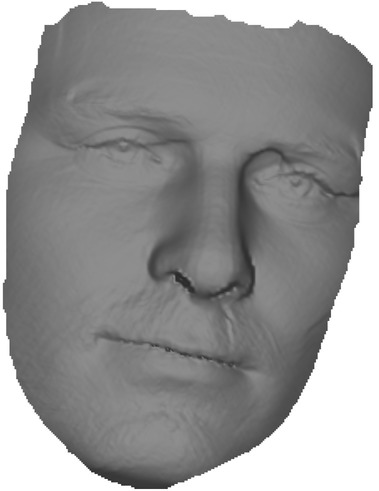}&
 \includegraphics[height=0.12\textwidth]{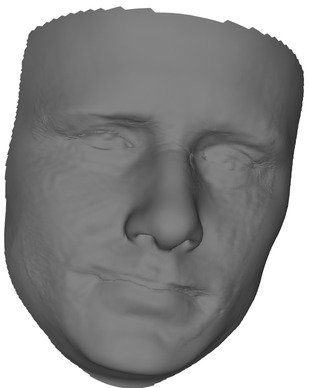}&
 \includegraphics[height=0.12\textwidth]{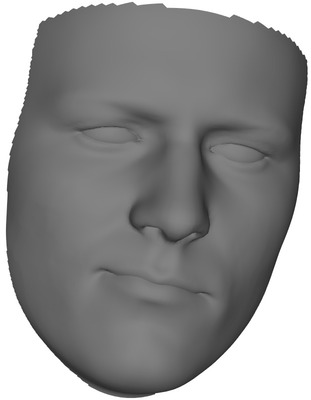}&
 \includegraphics[height=0.12\textwidth]{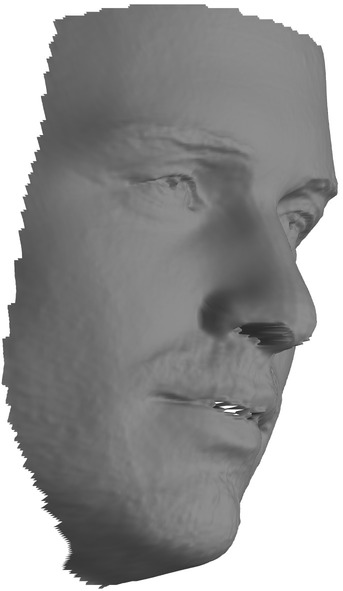}&
 \includegraphics[height=0.12\textwidth]{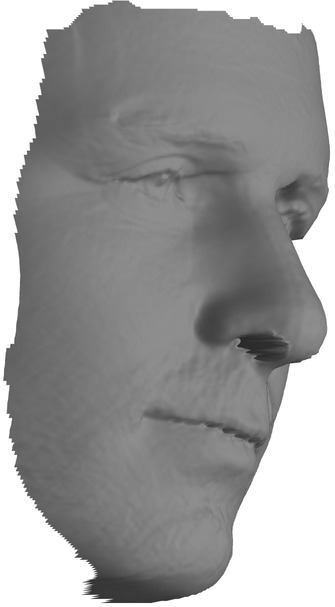}&
 \includegraphics[height=0.12\textwidth]{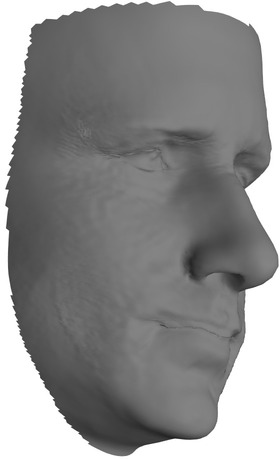}&
 \includegraphics[height=0.12\textwidth]{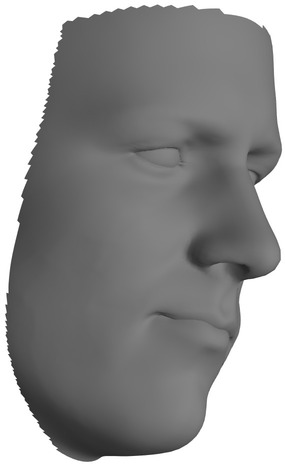}\tabularnewline
 Input & Ours & \cite{kemelmacher20113d} & \cite{richardson20163d} & \cite{zhu2015high} & Ours & \cite{kemelmacher20113d} & \cite{richardson20163d} & \cite{zhu2015high}
\end{tabular}
    \caption{Additional qualitative results.}
    \label{fig:vis_example}
\end{figure*}

\begin{figure*} %
        \centering
        \begin{tabular}{cccc}
        \centering
     \includegraphics[height=0.25\textwidth]{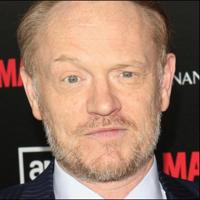}&
     \includegraphics[height=0.25\textwidth]{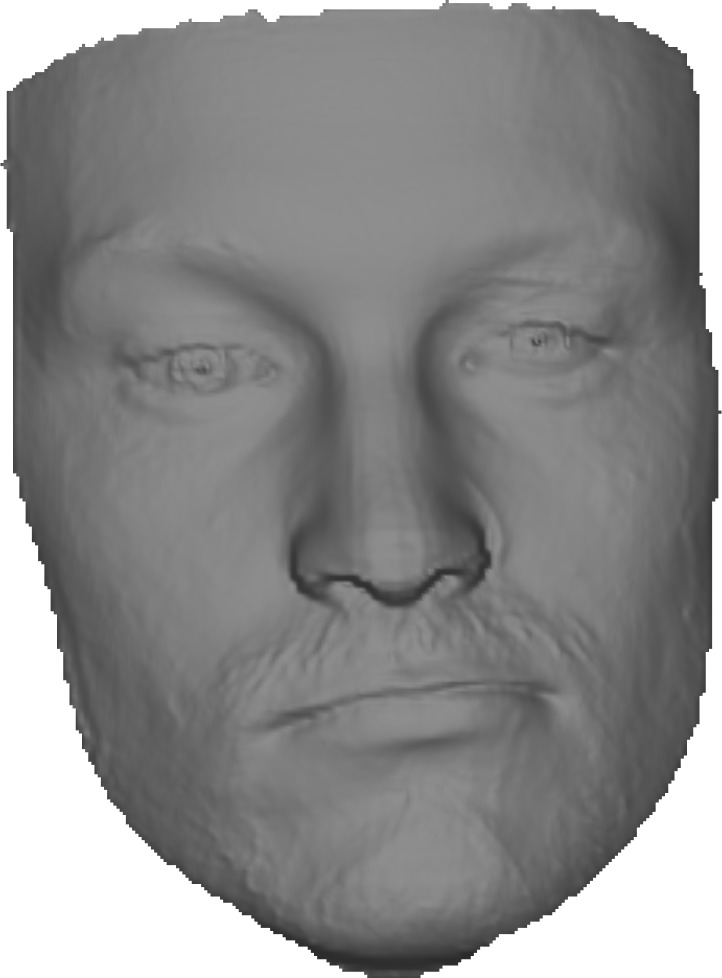}&
     \includegraphics[height=0.25\textwidth]{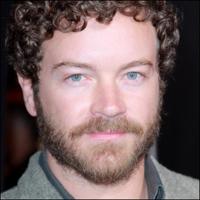}&
     \includegraphics[height=0.25\textwidth]{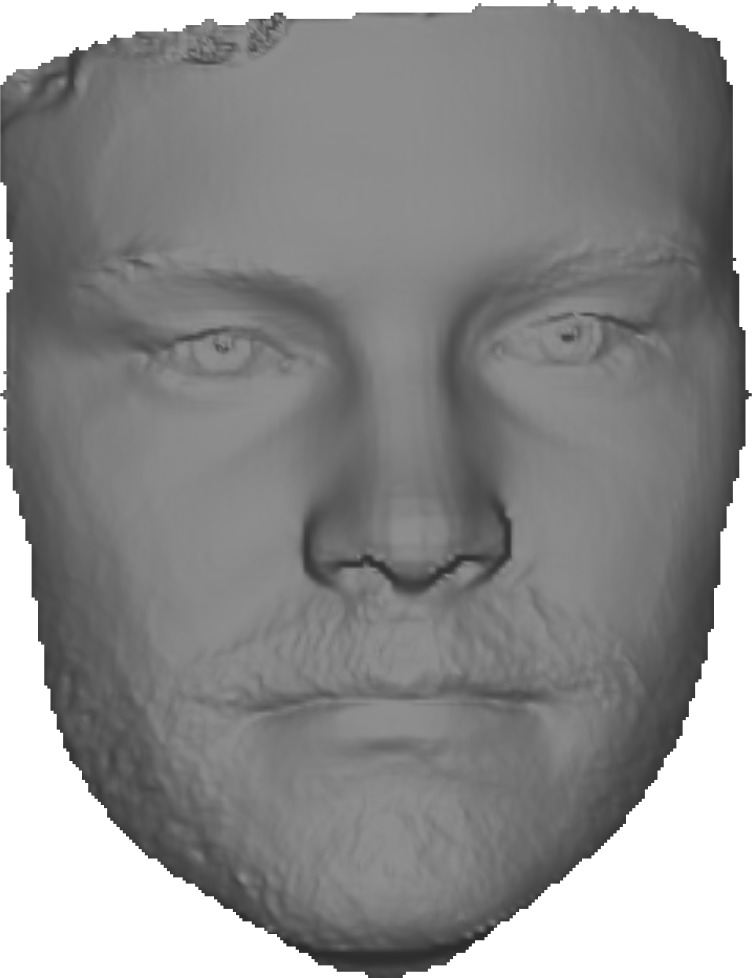}\tabularnewline
     \includegraphics[height=0.25\textwidth]{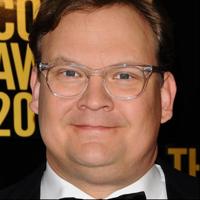}&
     \includegraphics[height=0.25\textwidth]{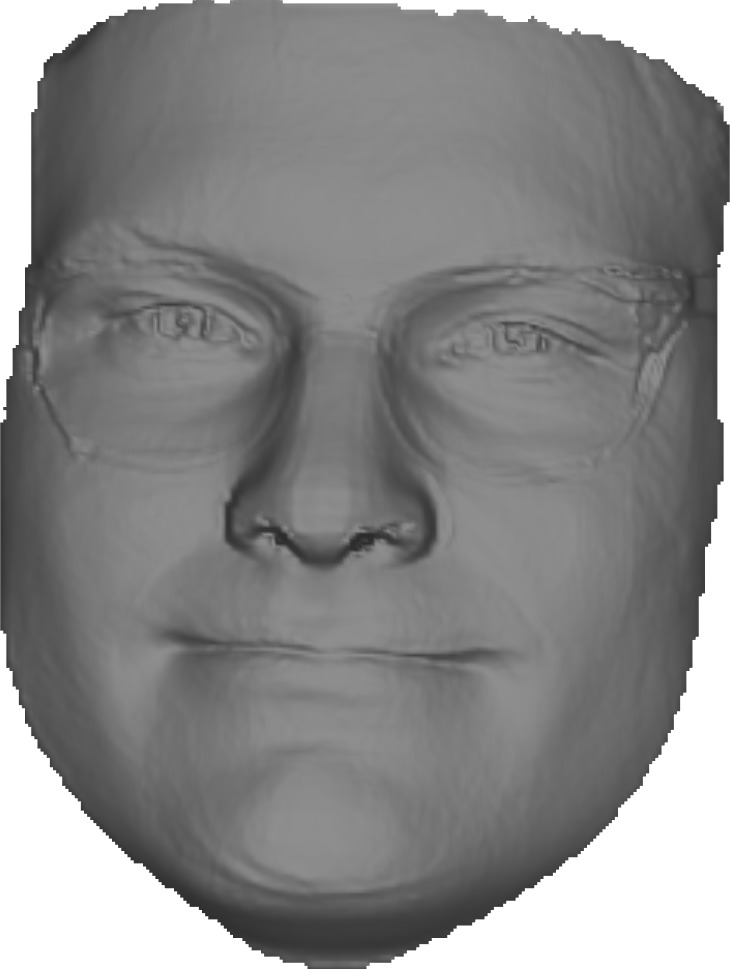}&
     \includegraphics[height=0.25\textwidth]{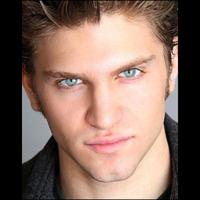}&
     \includegraphics[height=0.25\textwidth]{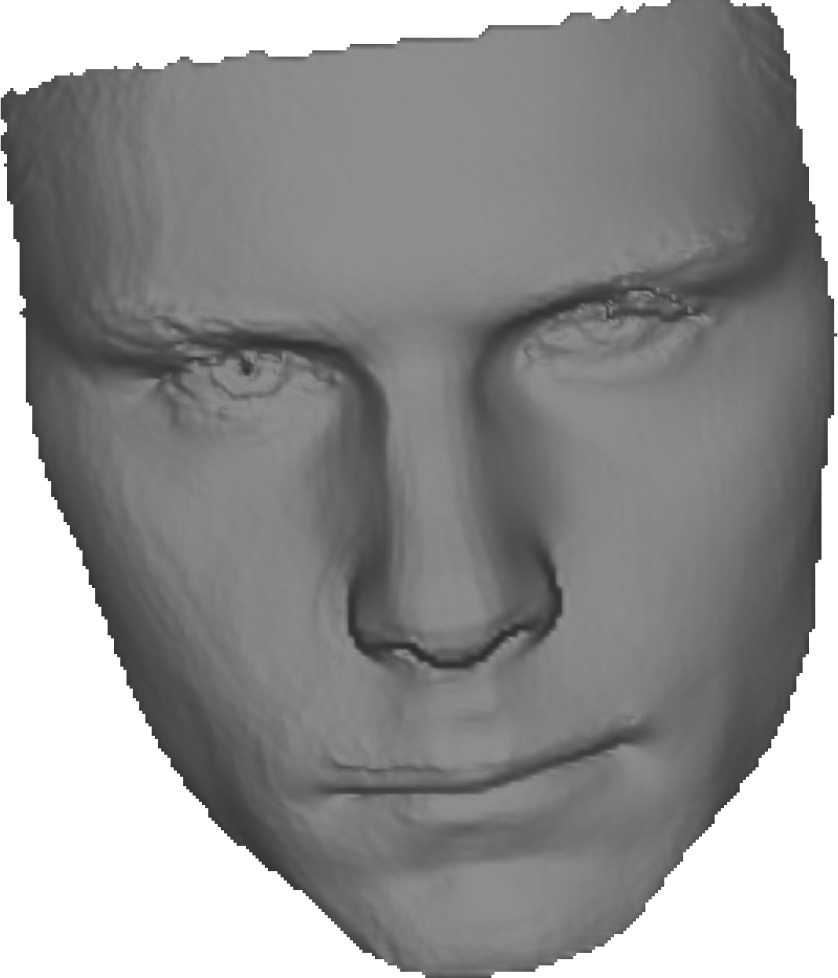}\tabularnewline
     \includegraphics[height=0.25\textwidth]{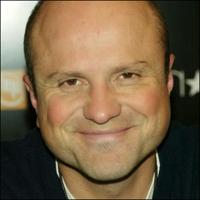}&
     \includegraphics[height=0.25\textwidth]{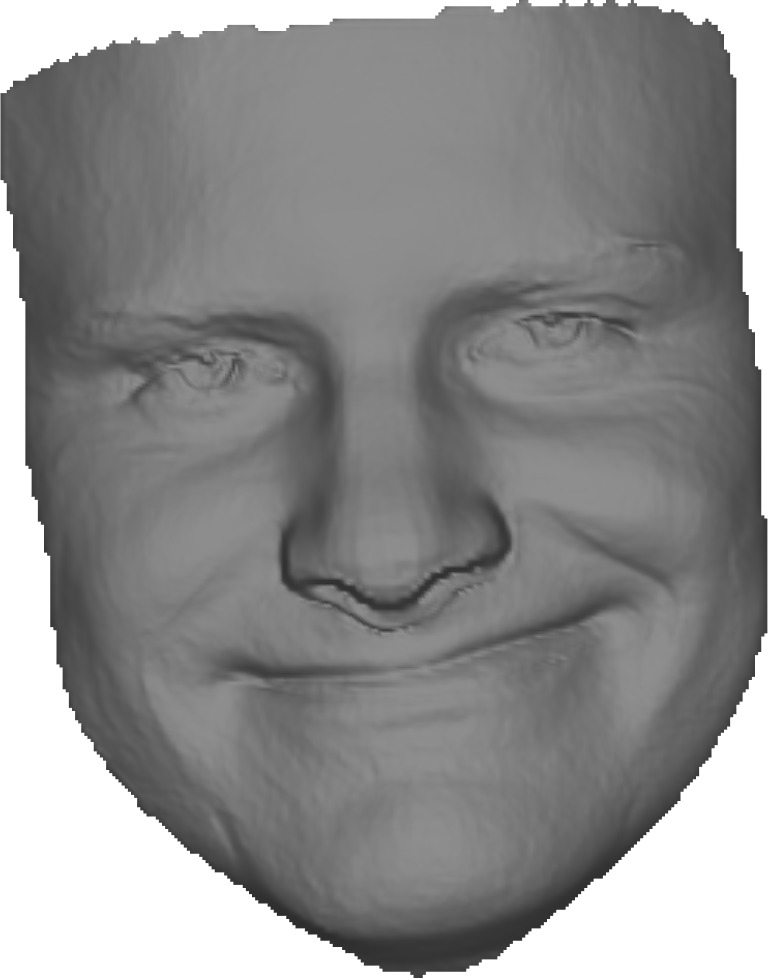}&
     \includegraphics[height=0.25\textwidth]{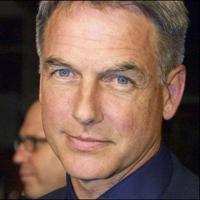}&
     \includegraphics[height=0.25\textwidth]{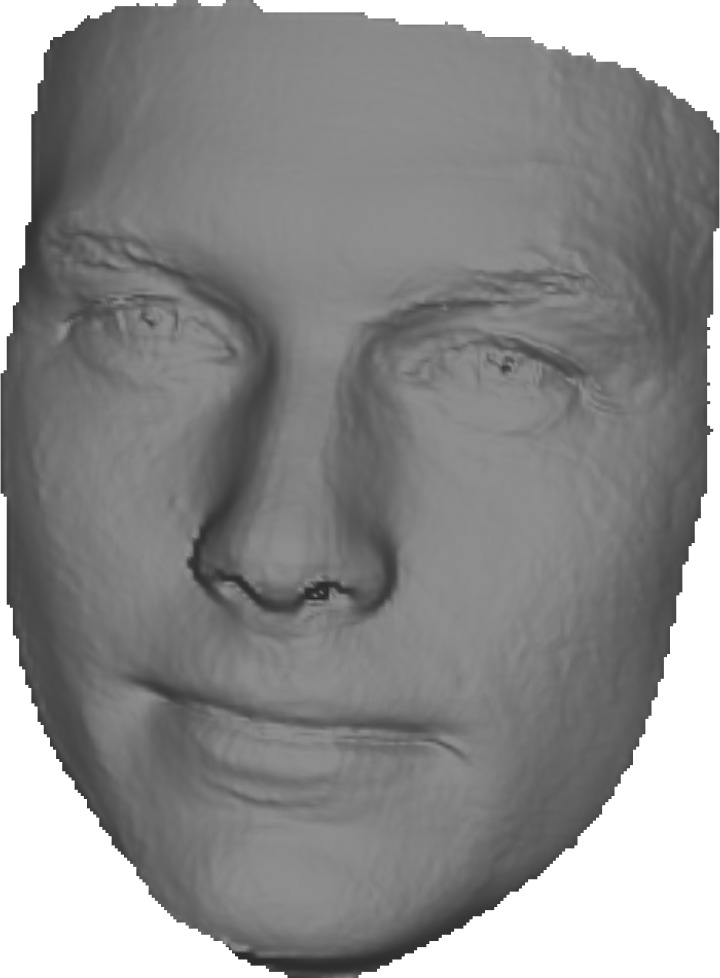}\tabularnewline
     \includegraphics[height=0.25\textwidth]{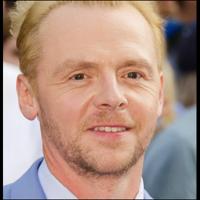}&
     \includegraphics[height=0.25\textwidth]{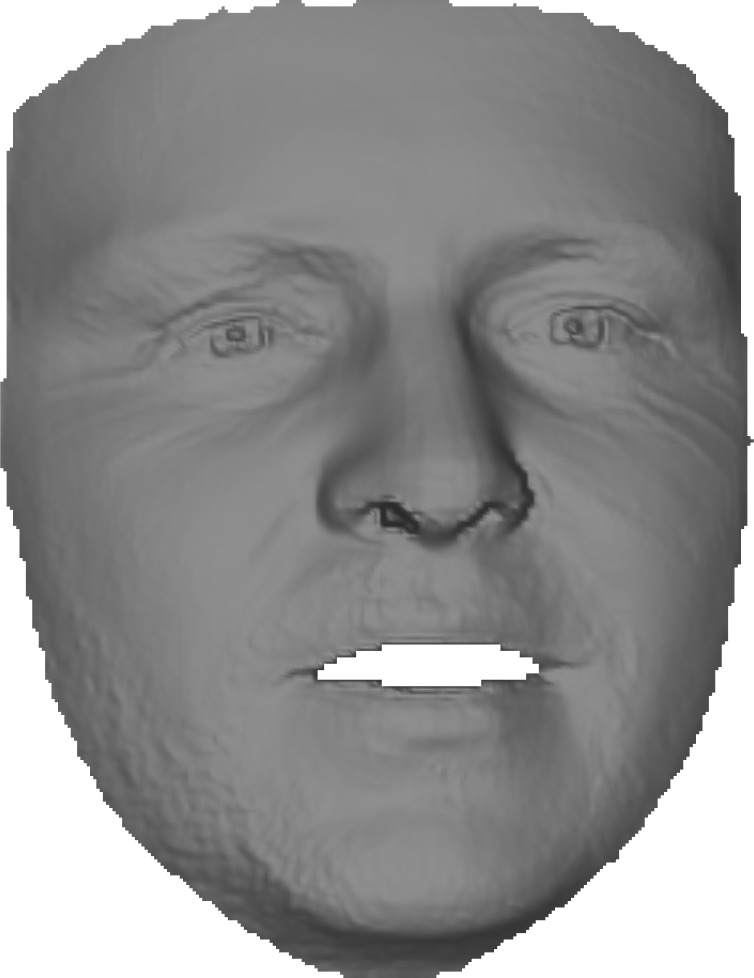}&
     \includegraphics[height=0.25\textwidth]{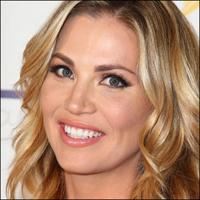}&
     \includegraphics[height=0.25\textwidth]{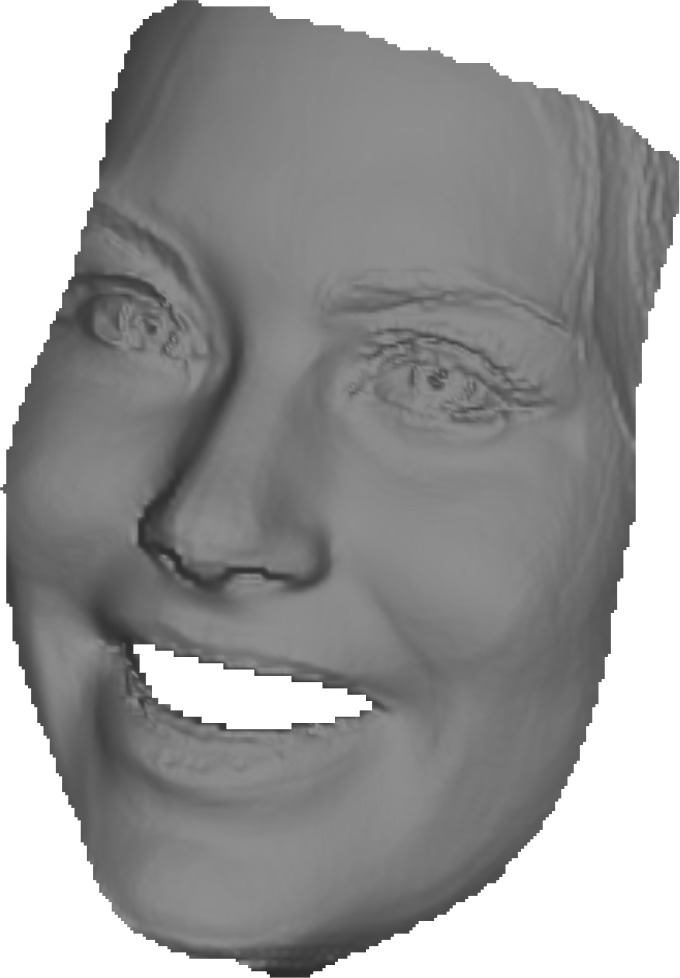}\tabularnewline
    \end{tabular}
        \caption[]{Additional qualitative results.}
        \label{fig:examples_large}
    \end{figure*}

In Figure~\ref{fig:vis_example} we present additional qualitative comparisons. First, note how our network correctly infers the face alignment without any external information, producing similar alignment to the state-of-the-art alignment from \cite{Kazemi_2014_CVPR}. The proposed method is able to produce fine facial details, as opposed to \cite{richardson20163d,zhu2015high}, while being more robust to different expressions compared to the template-based method of  \cite{kemelmacher20113d}. Figure~\ref{fig:examples_large} shows additional reconstructions.

\clearpage

\section{Synthetic Data}
\begin{figure}[h]
    \centering
      \includegraphics[width=\textwidth]{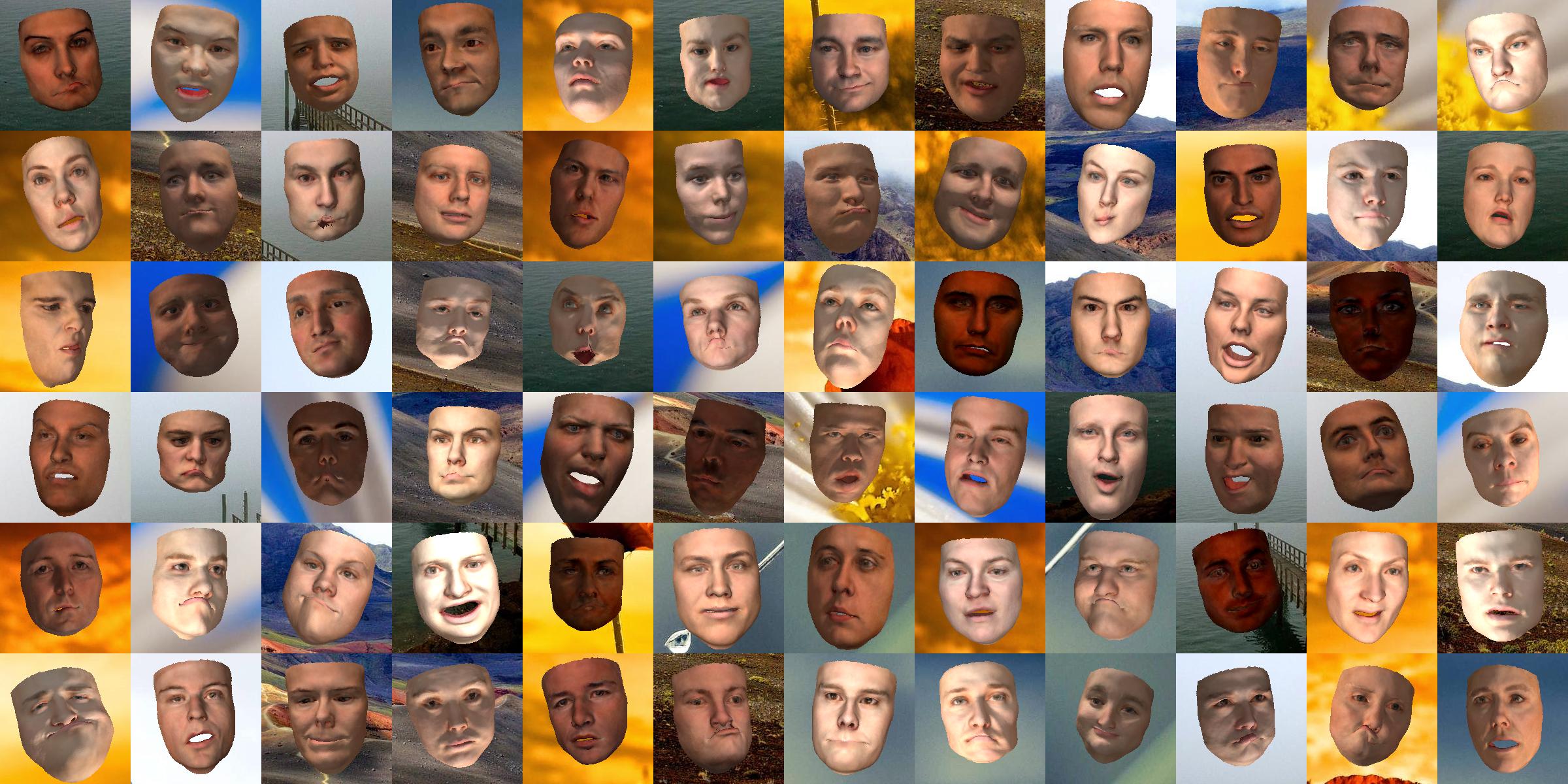}
    \caption{Synthetic data samples.}
    \label{fig:synthetic}
\end{figure}

In Figure~\ref{fig:synthetic} sampled synthetic examples are visualized, where random backgrounds are used. The rendered faces differ extensively in their geometry, texture, illumination, and reflectance properties.

\subsection{Generalizing from Synthetic Data}
\begin{figure*}[h] %
    \centering
    \begin{tabular}{ccccccccc}
  \includegraphics[height=0.10\textwidth]{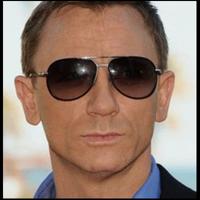}&
 \includegraphics[height=0.10\textwidth]{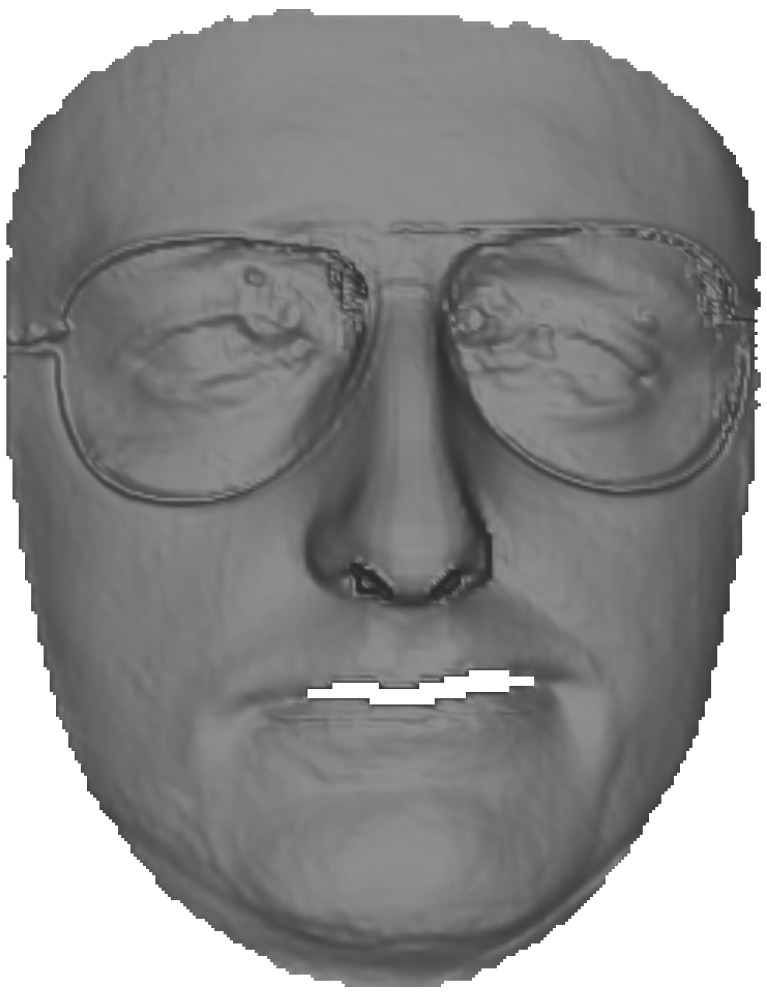}&
 \includegraphics[height=0.10\textwidth]{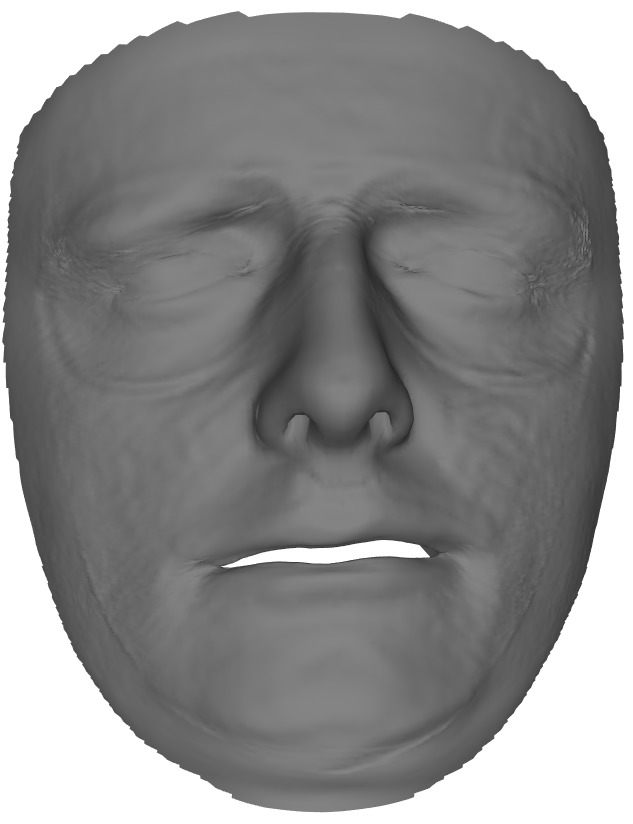}&
 \includegraphics[height=0.10\textwidth]{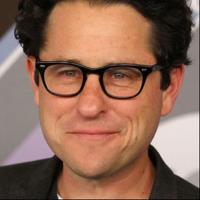}&
 \includegraphics[height=0.10\textwidth]{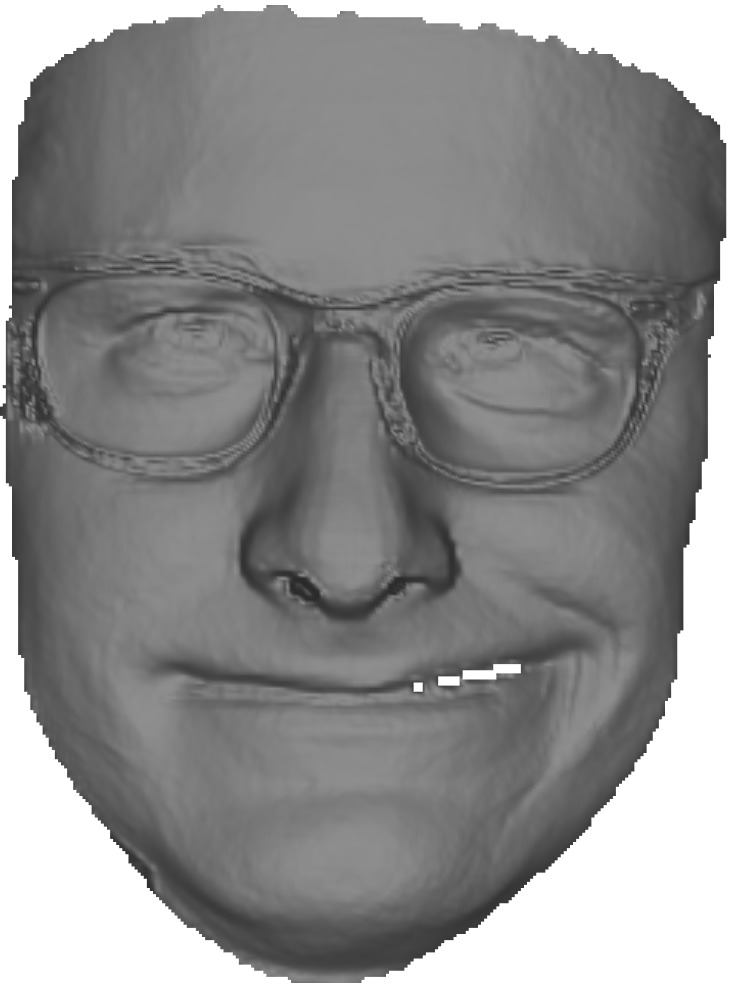}&
 \includegraphics[height=0.10\textwidth]{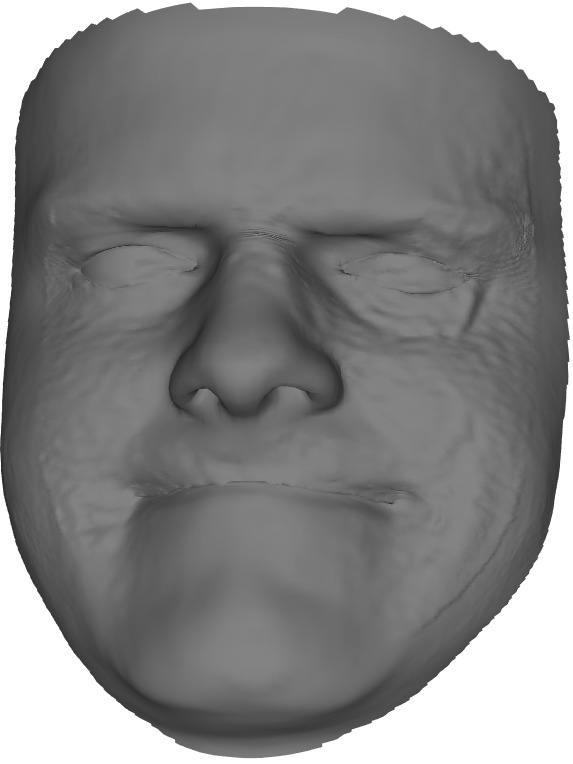}&
 \includegraphics[height=0.10\textwidth]{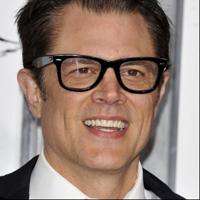}&
 \includegraphics[height=0.10\textwidth]{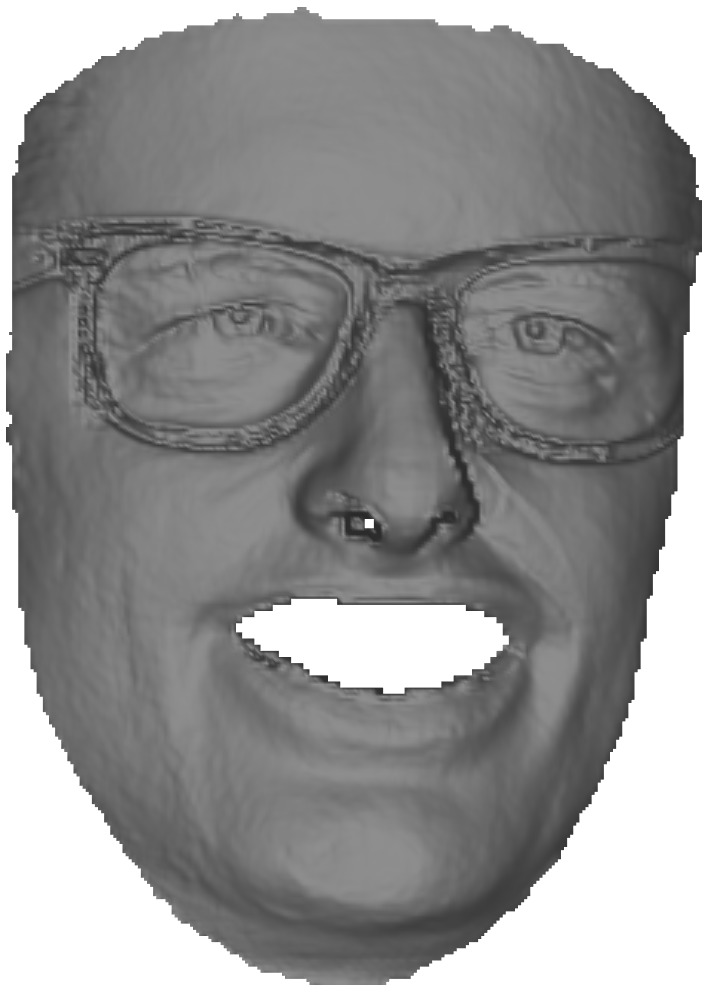}&
 \includegraphics[height=0.10\textwidth]{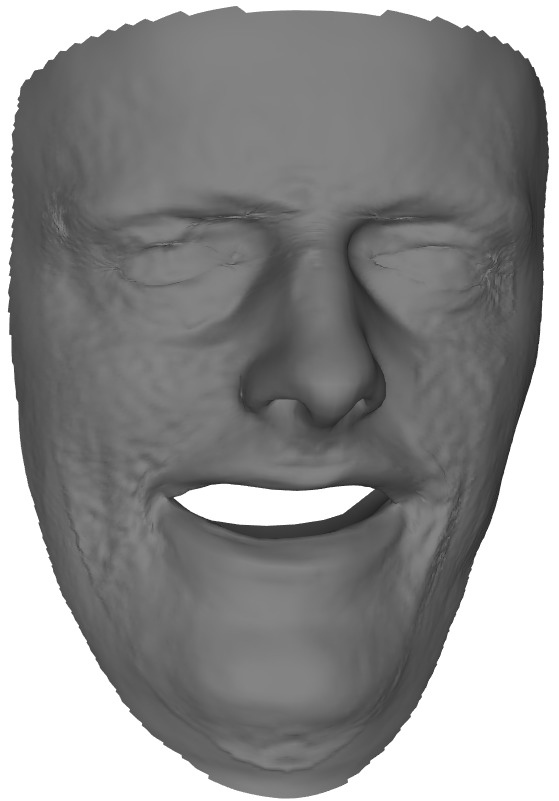}\tabularnewline
  \includegraphics[height=0.10\textwidth]{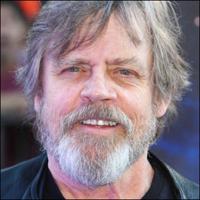}&
 \includegraphics[height=0.10\textwidth]{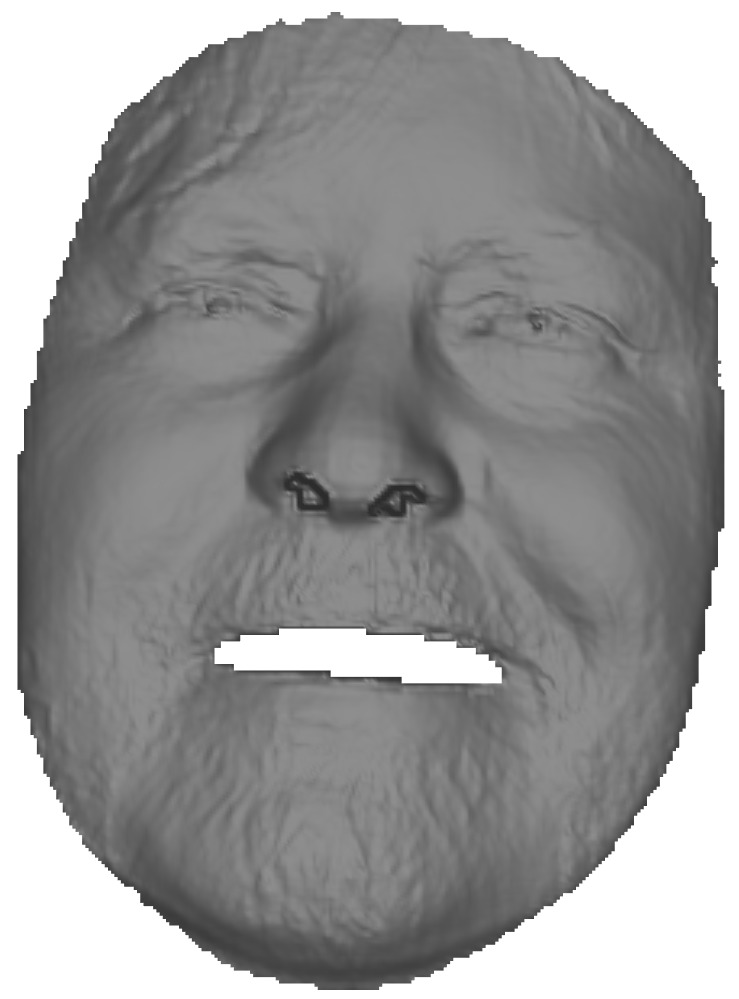}&
 \includegraphics[height=0.10\textwidth]{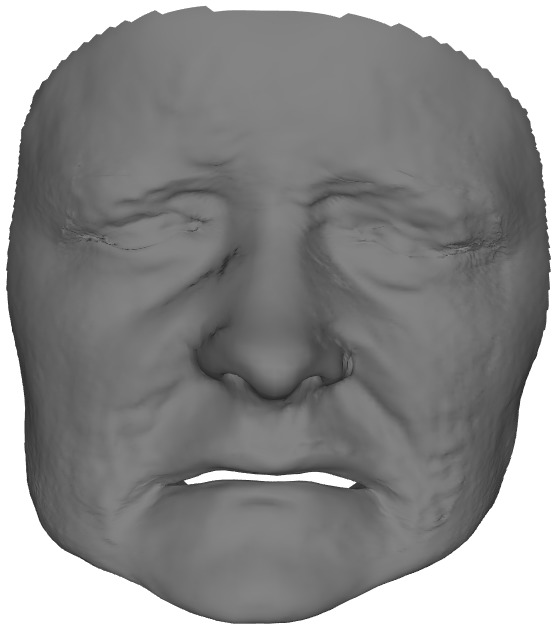}&
 \includegraphics[height=0.10\textwidth]{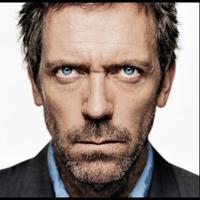}&
 \includegraphics[height=0.10\textwidth]{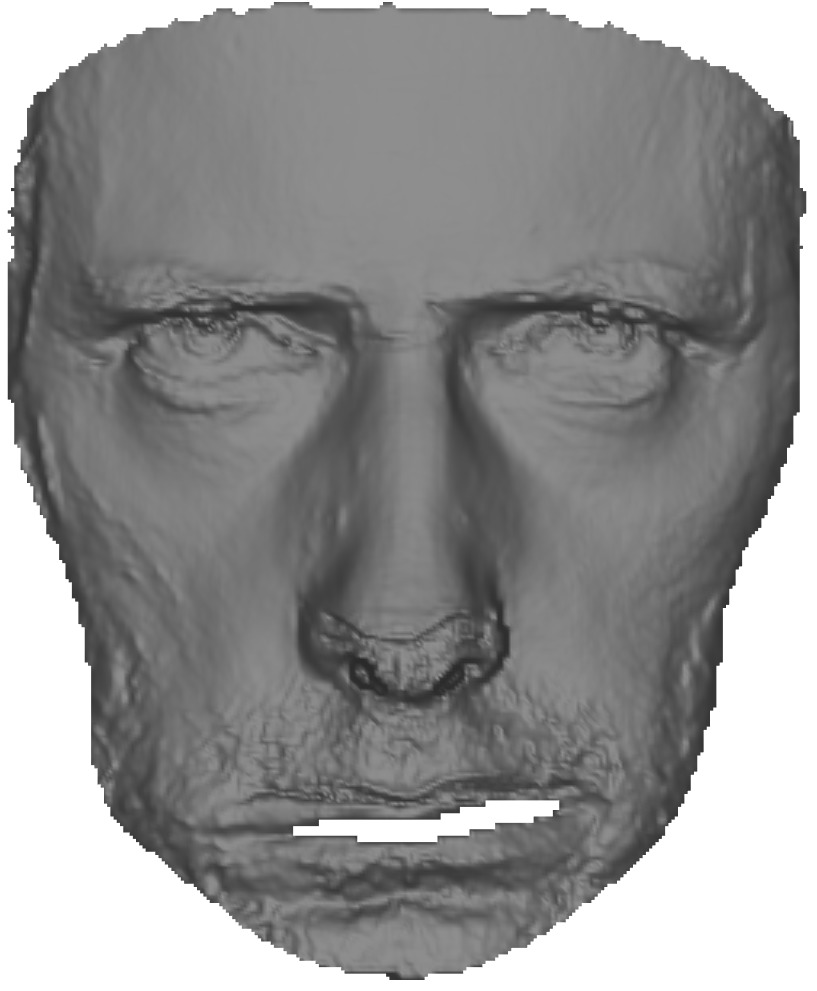}&
 \includegraphics[height=0.10\textwidth]{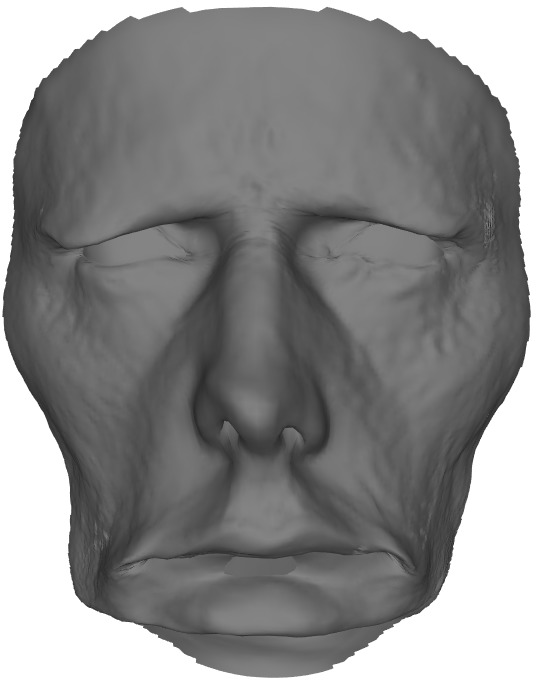}&
 \includegraphics[height=0.10\textwidth]{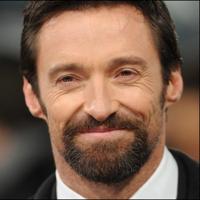}&
 \includegraphics[height=0.10\textwidth]{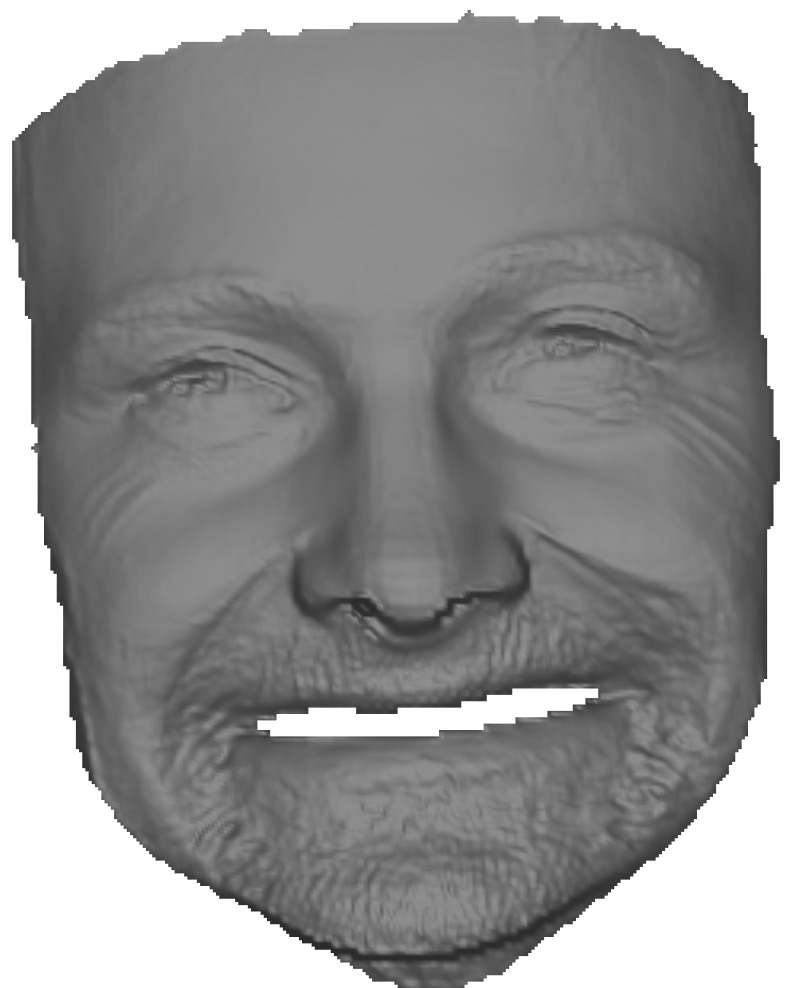}&
 \includegraphics[height=0.10\textwidth]{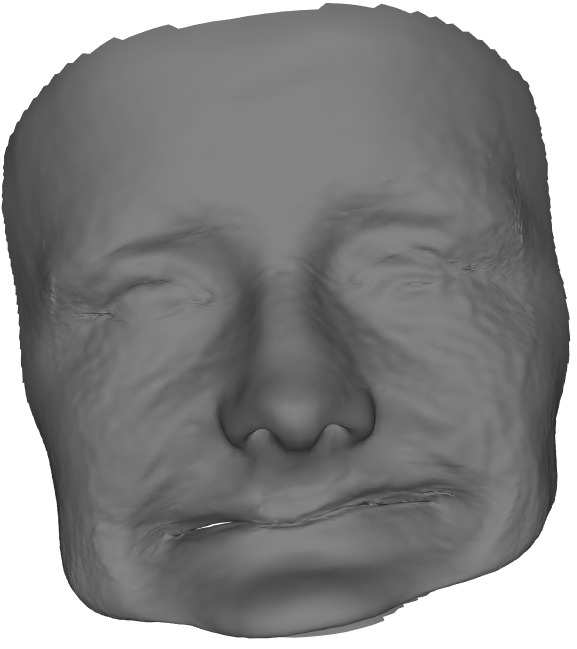}\tabularnewline

 Input & Ours & \cite{richardson20163d} & Input & Ours & \cite{richardson20163d} & Input & Ours & \cite{richardson20163d}
\end{tabular}
    \caption{Generalization from synthetic data.}
    \label{fig:challenge_example}
\end{figure*}

While data generation grants us flexibility and allows generation of large-scale datasets, there are still some limitations for synthetic data.
As noted in \cite{richardson20163d}, while training on synthetic faces generally produces plausible results on \textit{in-the-wild} images, the network might fail when the input contains details that are not seen in the synthetic dataset, such as glasses or facial hair. In Figure \ref{fig:challenge_example} we show how our method handles such examples compared to \cite{richardson20163d}. From the results we can see that both methods show some robustness to eyeglasses, even when the eyes themselves are occluded. Regarding facial hair, one can see that a dominant beard might confuse both methods and make them misalign the chin or the mouth. Still, our method is able to produce more viable results than those of \cite{richardson20163d}.
\newpage
\section{Further Analysis}
Next, we present a few additional experiments conducted on the different elements of the proposed network.
\subsection{CoarseNet}
\begin{figure}[h] %
    \centering
	\includegraphics[width=0.44\textwidth]{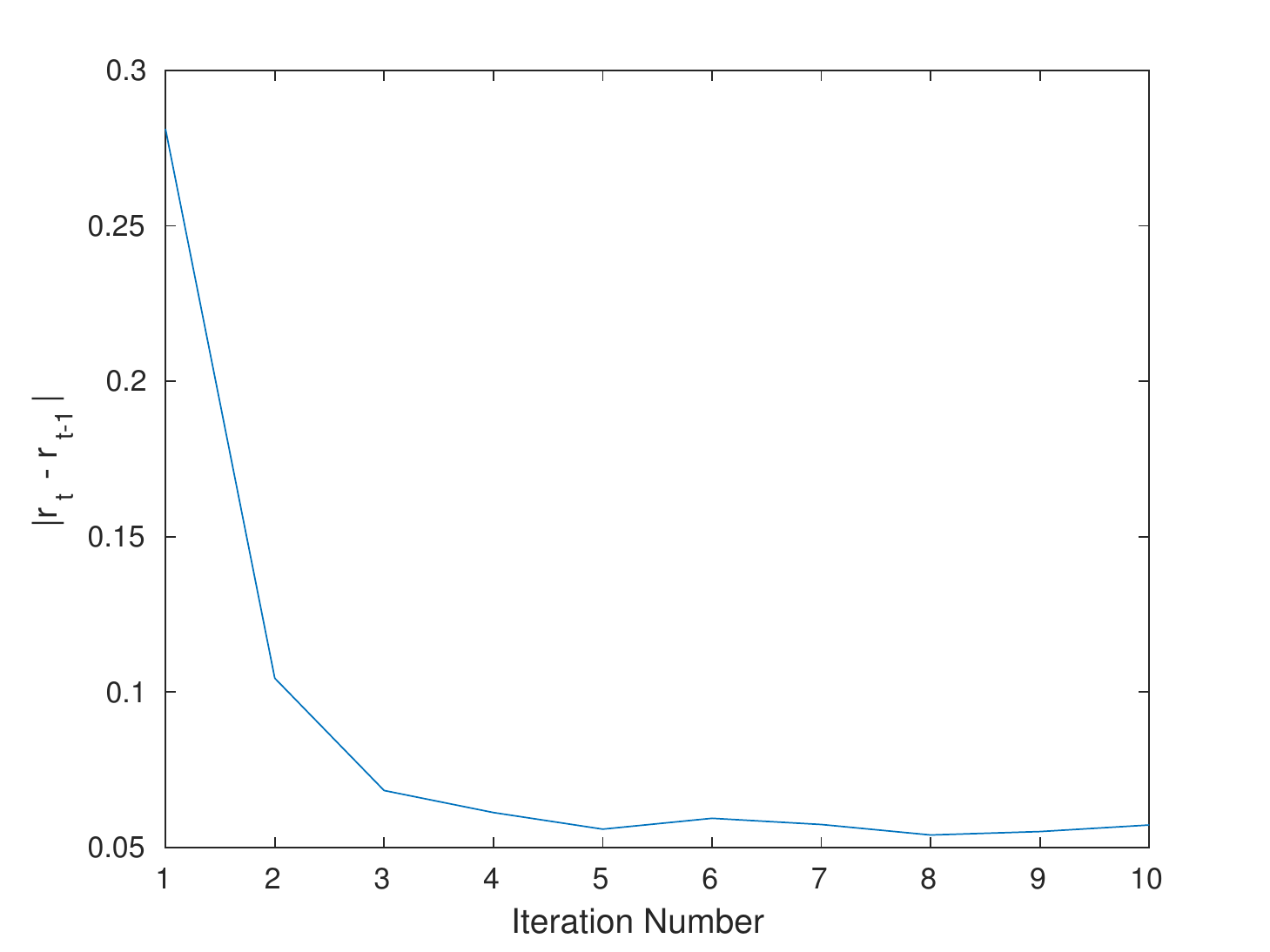}
    \caption{Convergence graph. The average MSE as a function of the iteration number.}
    \label{fig:iter-error}
\end{figure}

A key property of iterative networks is convergence. To validate that CoarseNet meets this requirement, we calculated the average change in the output of CoarseNet between different iterations. As can be seen in Figure~\ref{fig:iter-error}, the network indeed converges after a few iterations.

\subsection{FineNet}
\begin{figure}[h]
\centering
\begin{subfigure}{0.16\textwidth}
  \includegraphics[width=\textwidth]{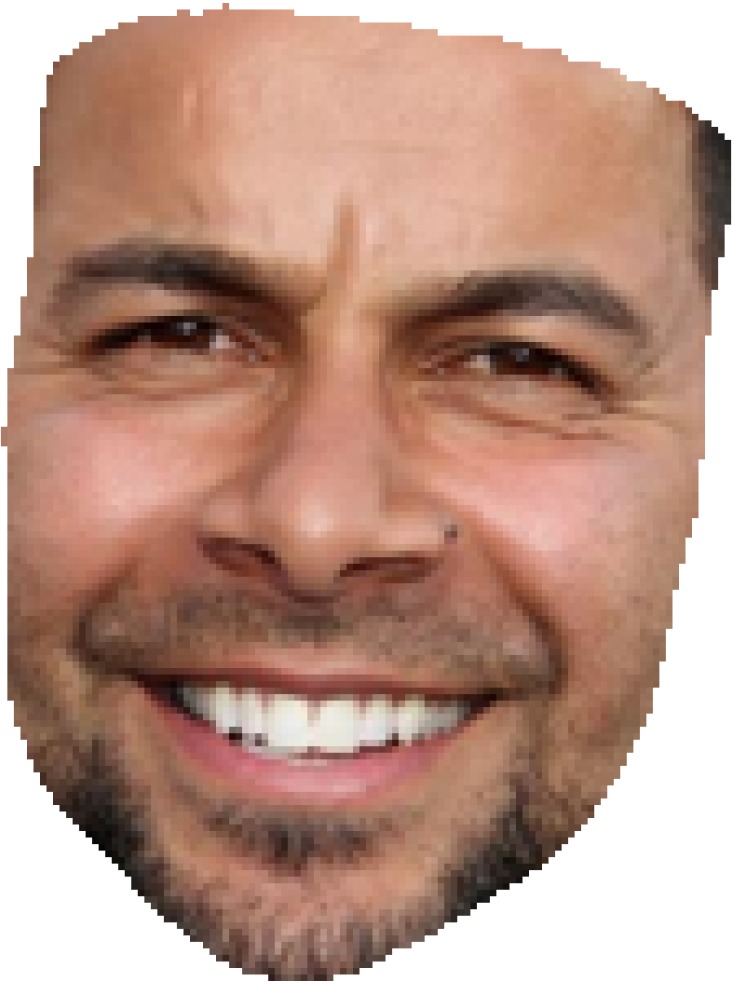}
\end{subfigure}
\begin{subfigure}{0.16\textwidth}
  \includegraphics[width=\textwidth]{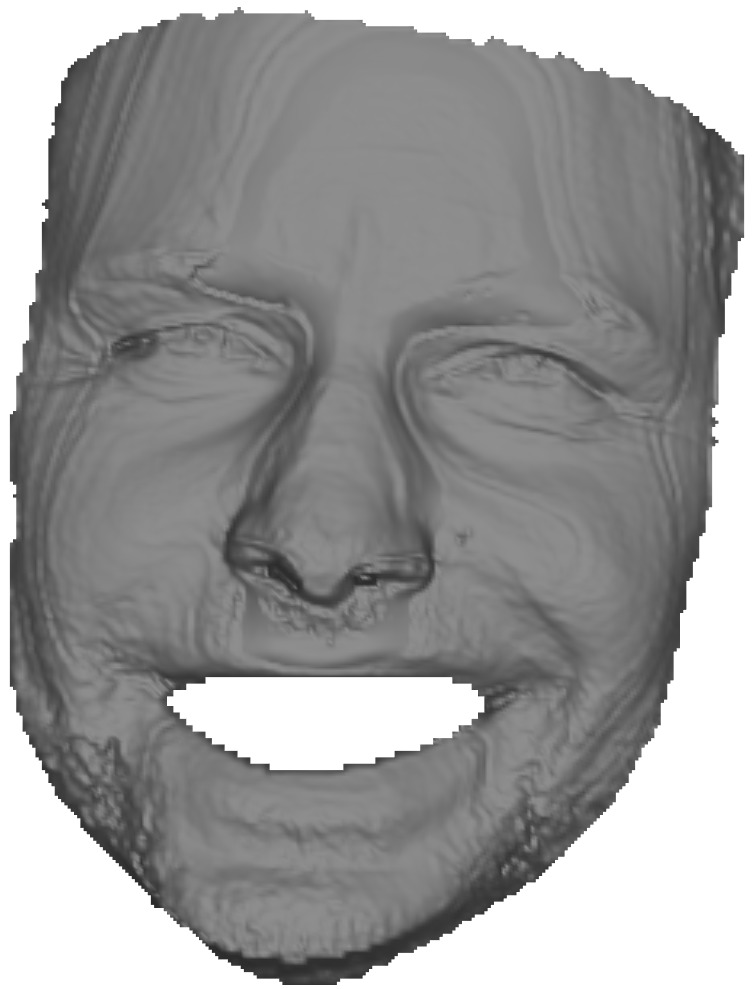}
\end{subfigure}
\begin{subfigure}{0.16\textwidth}
  \includegraphics[width=\textwidth]{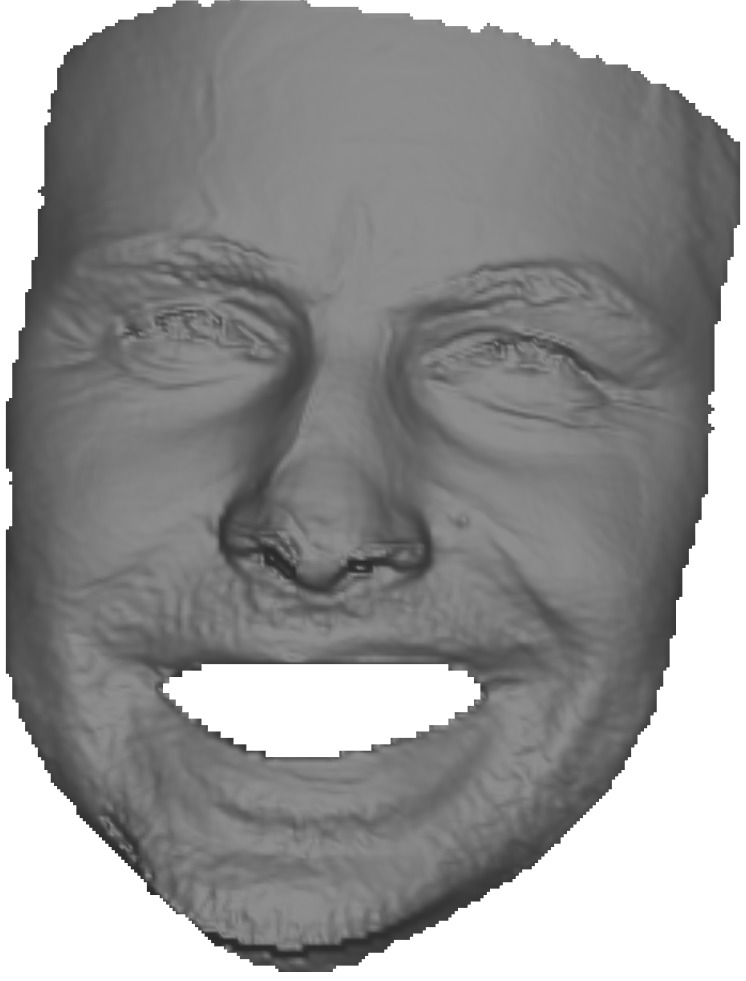}
\end{subfigure}
\begin{subfigure}{0.16\textwidth}
  \includegraphics[width=\textwidth]{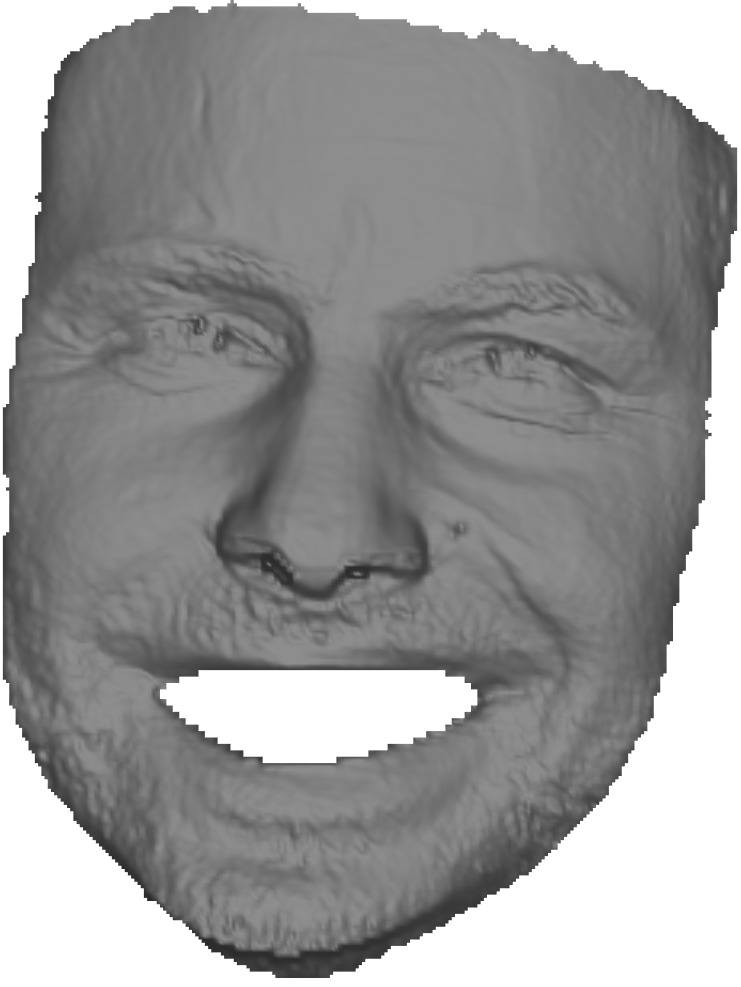}
\end{subfigure}
\begin{subfigure}{0.16\textwidth}
  \includegraphics[width=\textwidth]{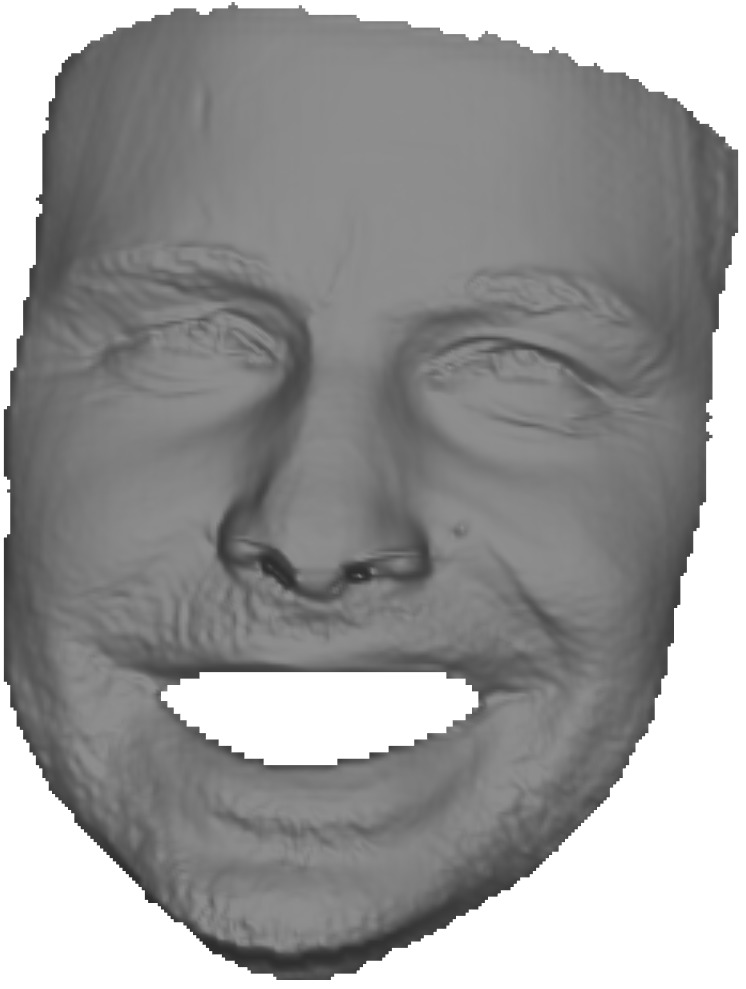}
\end{subfigure}
\caption{Network depth. From left to right, the input image, and FineNet results using 1 to 4 blocks from VGG Face Net. }
\label{fig:finenet_depth_ablation}
\end{figure}

As detailed in the paper, FineNet starts with a set of convolutional blocks from the VGG Face Net~\cite{Parkhi15}, each followed by a pooling layer. The output of these blocks is then connected together to form a set of dense feature maps. While using more VGG blocks could possibly provide more data for the final prediction, it would also result in a larger network, increasing the overall training and runtime complexity. As Shape-from-Shading mainly relies on local features we choose to truncate the network after the third pooling operation. As shown in Figure~\ref{fig:finenet_depth_ablation}, while using only a single block results in discontinuances and artifacts, using two or more blocks produces reasonable results.

Another interesting property of FineNet presented in the paper is its robustness to different input sizes, allowing it to extract more details when a high-resolution input is given. Note that the same does not hold for CoarseNet which uses a fixed averaging operator. However, as CoarseNet recovers only the coarse geometry it does not require a high-resolution input and would not benefit from it. In practice, we always scale the input given to CoarseNet to $200\times200$, while feeding FineNet with inputs in the desired scale.

\newpage
\section{Supplementary Quantitative Analysis}
\begin{figure}[h] %
    \centering
    \begin{tabular}{ccccc}
 \includegraphics[height=0.18\textwidth,trim = 130 265 150 210,clip=true]{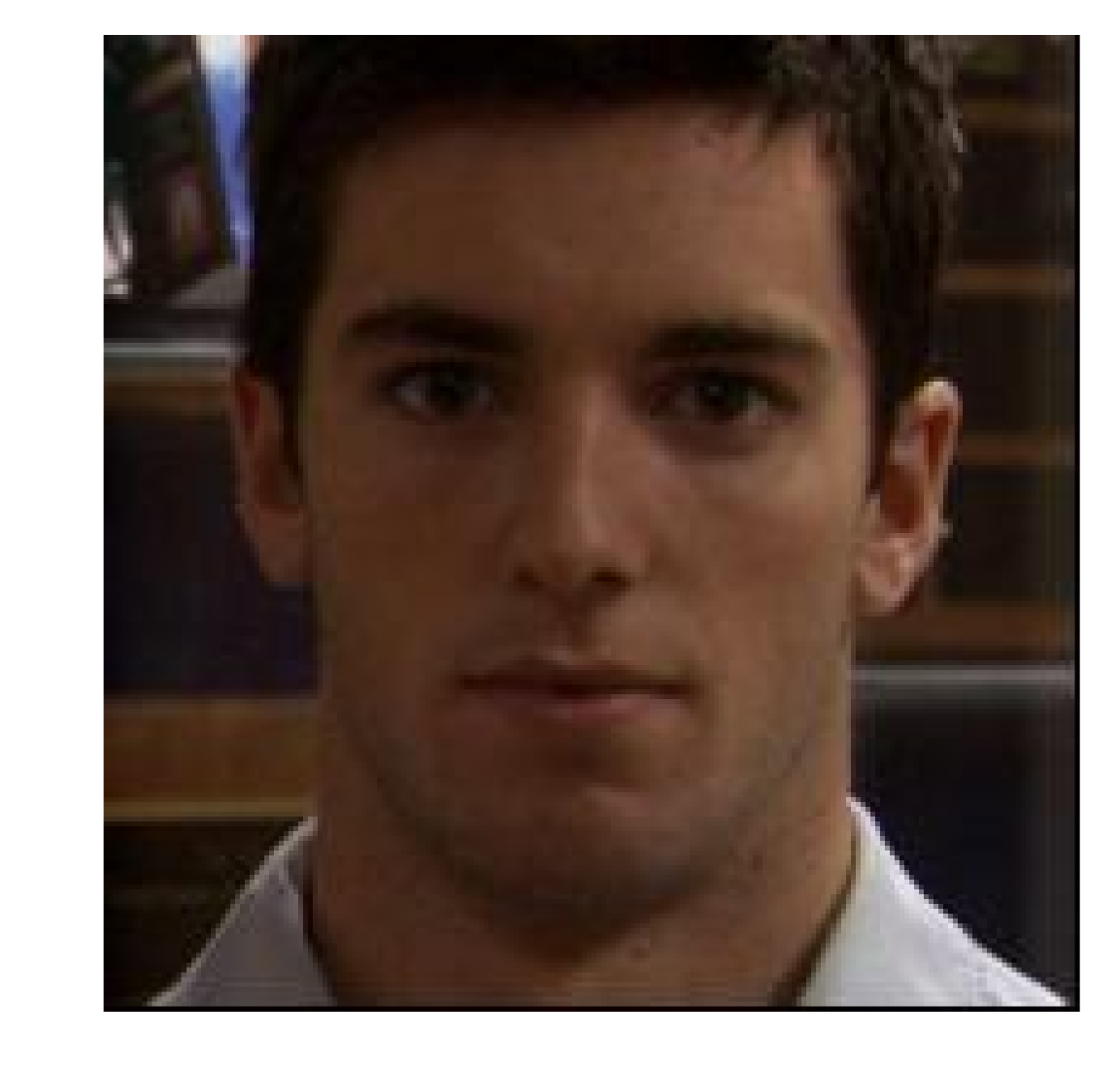}&
 \includegraphics[height=0.18\textwidth,trim = 130 210 100 210,clip=true]{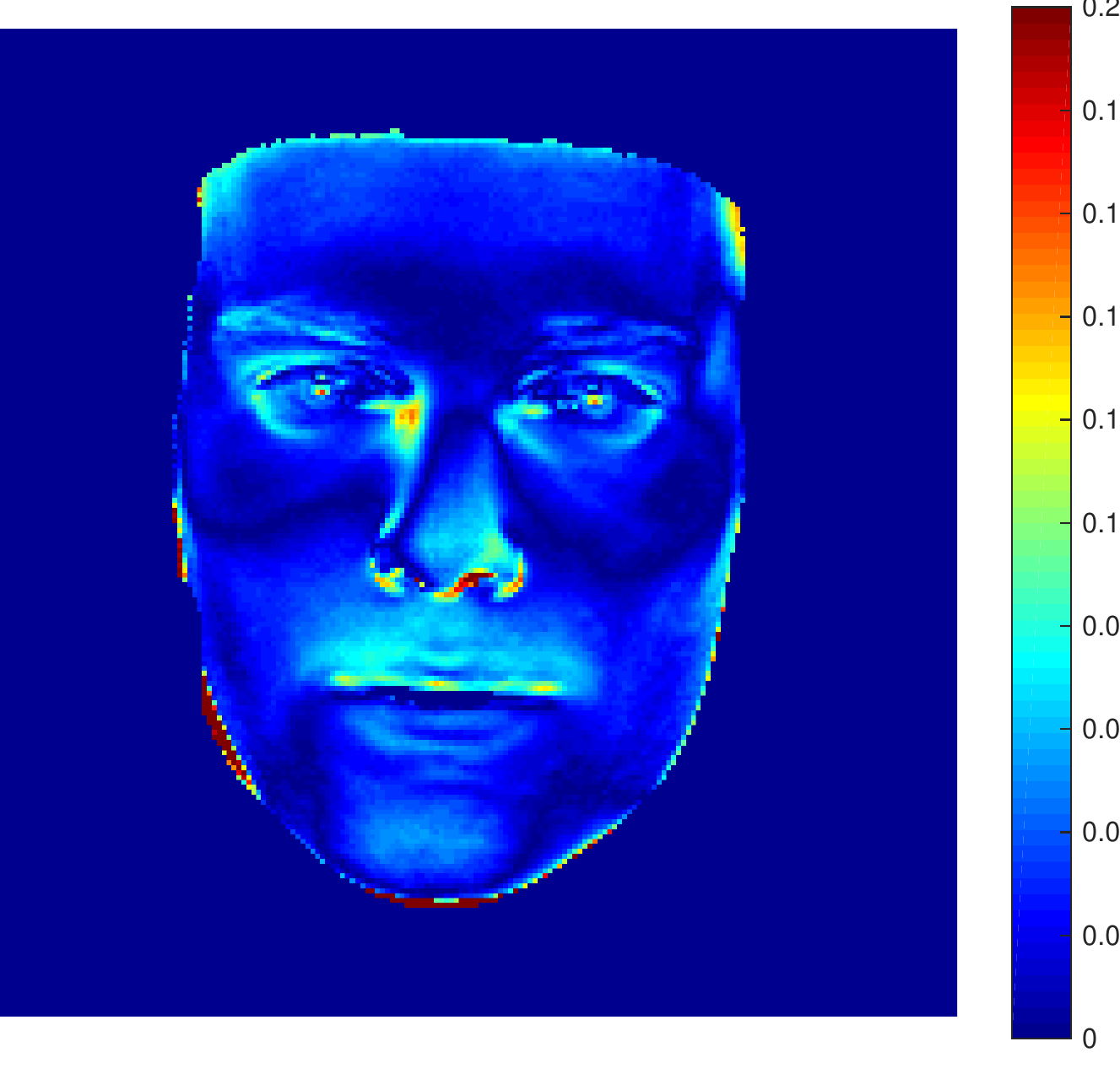}&
 \includegraphics[height=0.18\textwidth,trim = 130 210 100 210,clip=true]{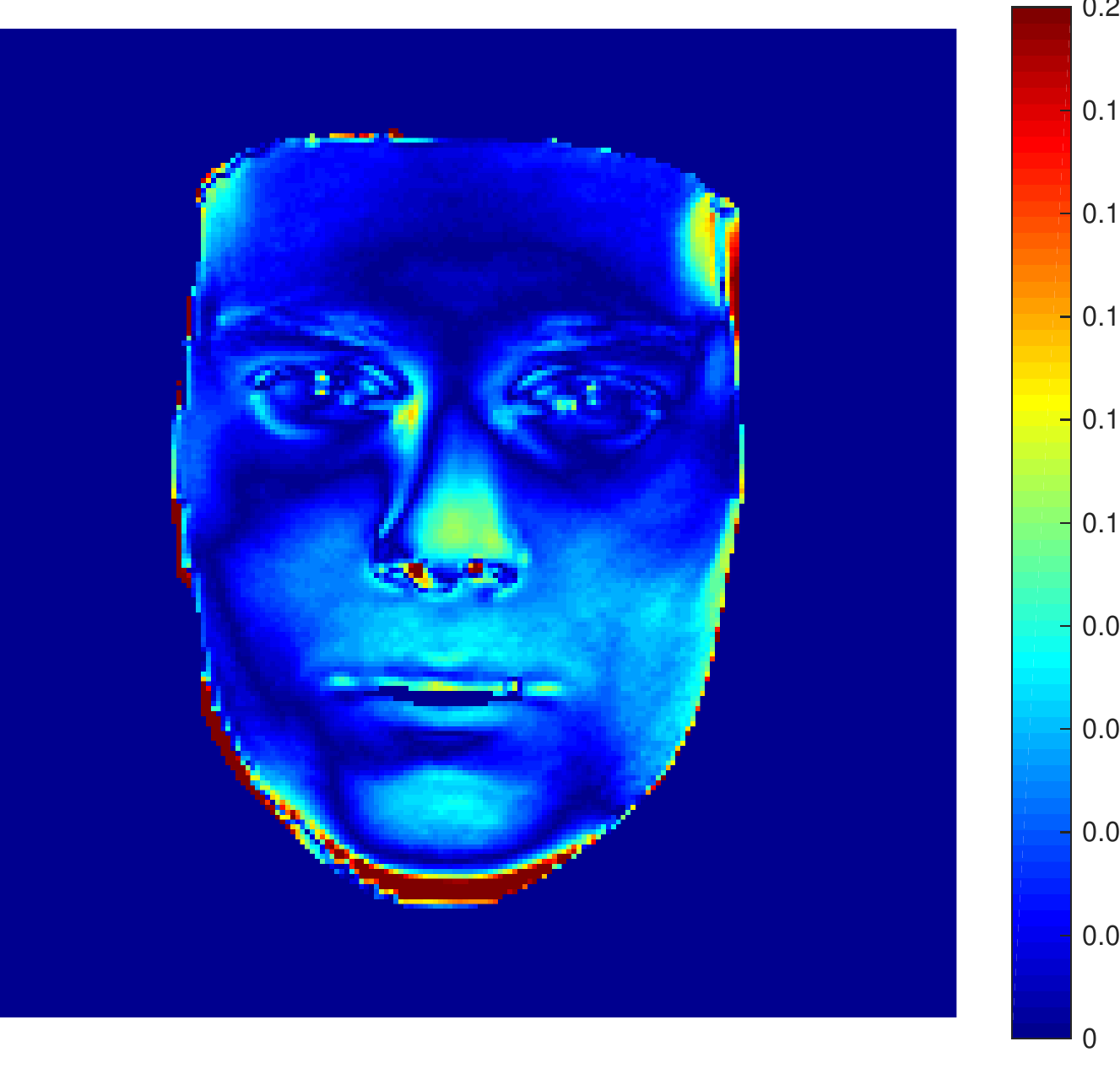}&
 \includegraphics[height=0.18\textwidth,trim = 130 210 100 210,clip=true]{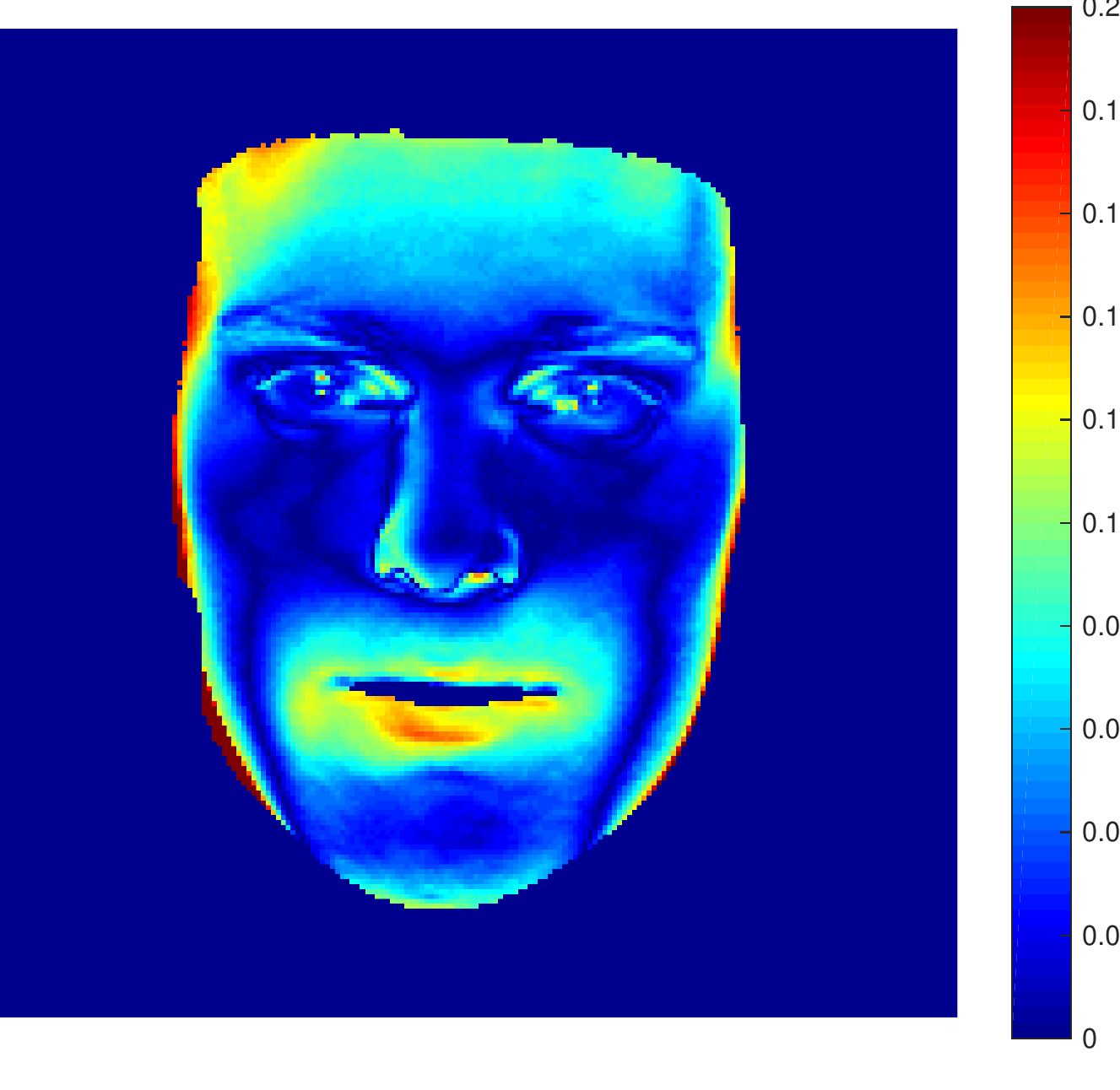}&
 \includegraphics[height=0.18\textwidth,trim = 130 210 100 210,clip=true]{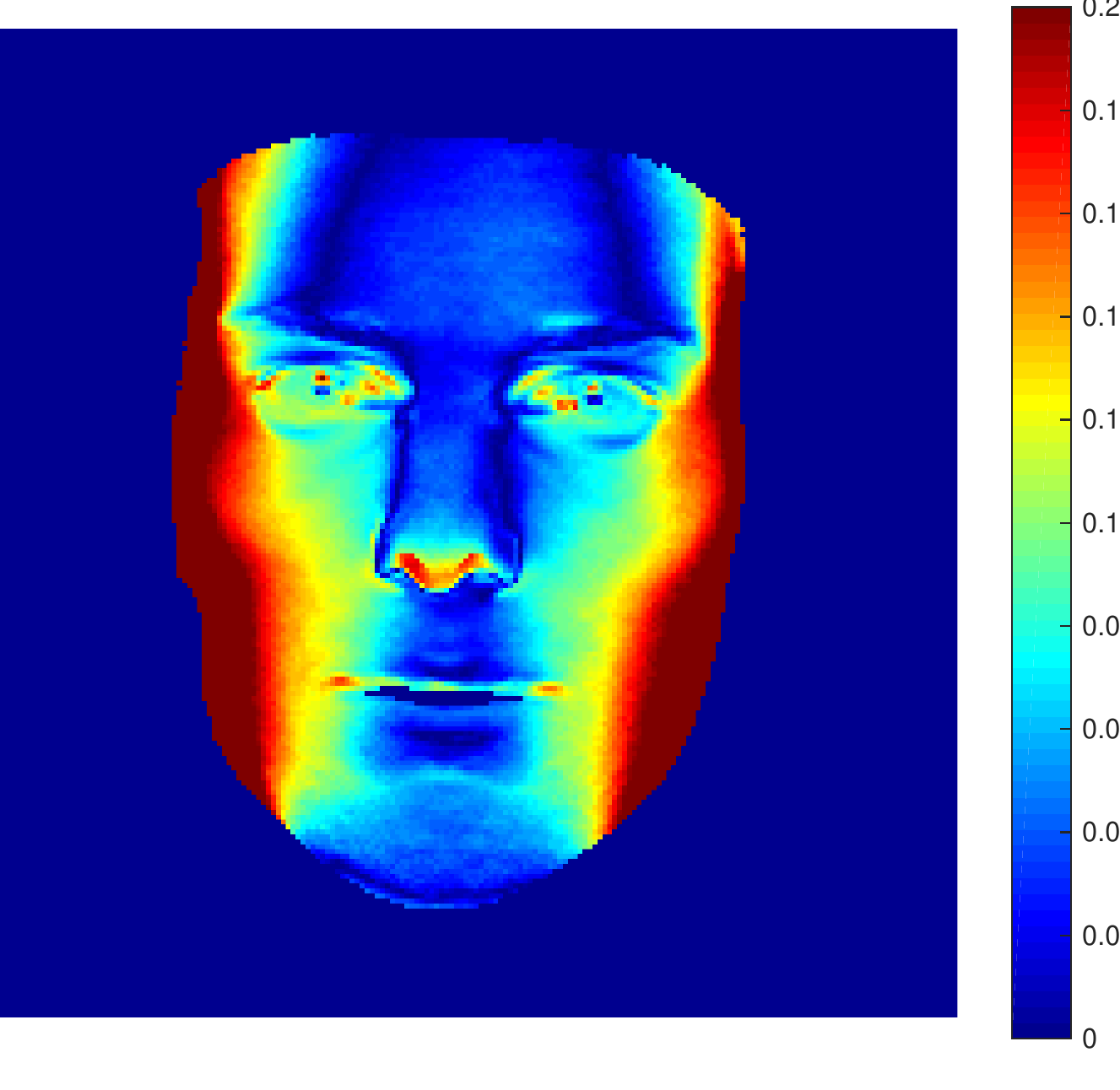}\tabularnewline
  \includegraphics[height=0.18\textwidth,trim = 130 265 150 210,clip=true]{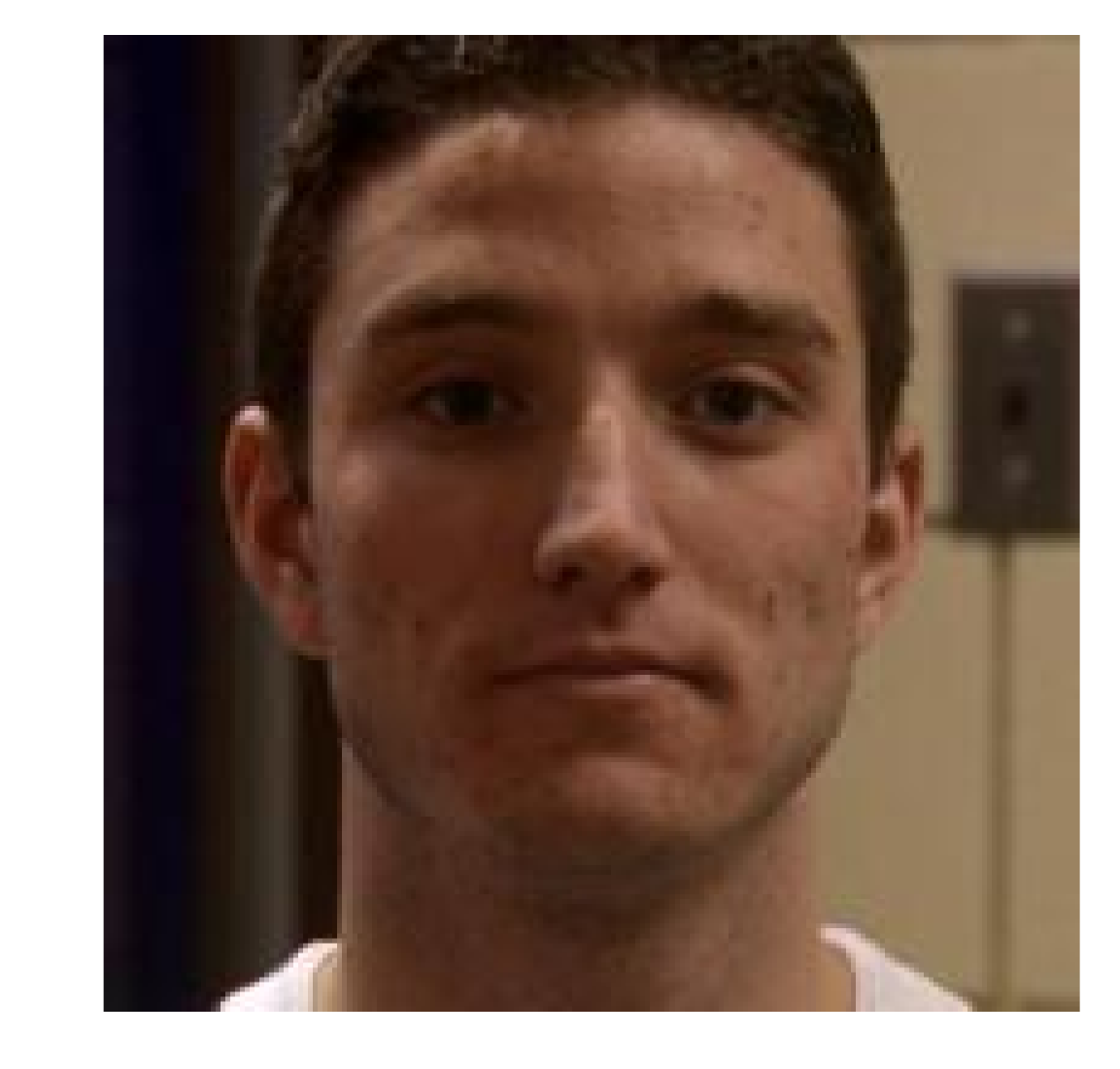}&
 \includegraphics[height=0.18\textwidth,trim = 130 210 100 210,clip=true]{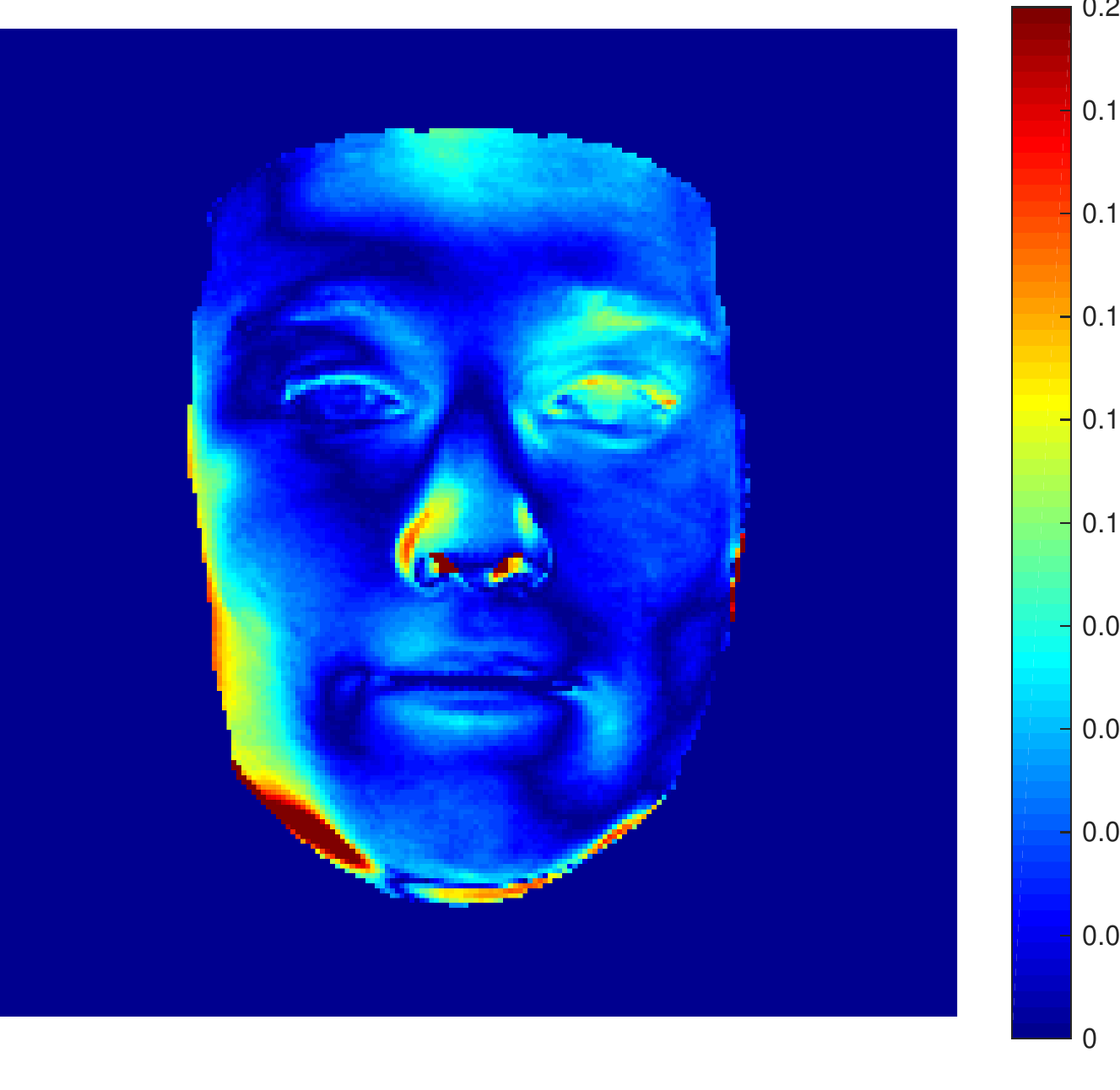}&
 \includegraphics[height=0.18\textwidth,trim = 130 210 100 210,clip=true]{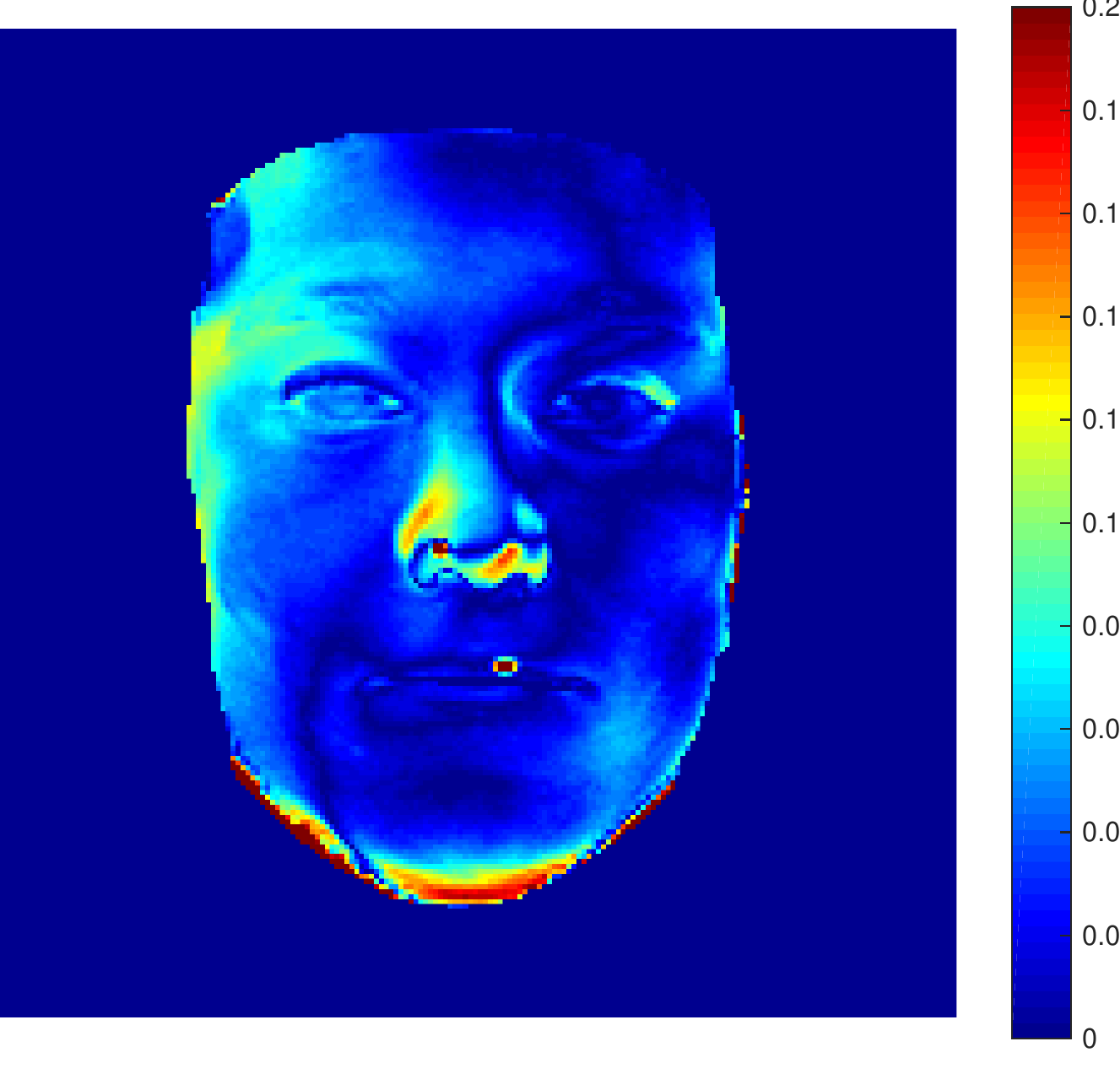}&
 \includegraphics[height=0.18\textwidth,trim = 130 210 100 210,clip=true]{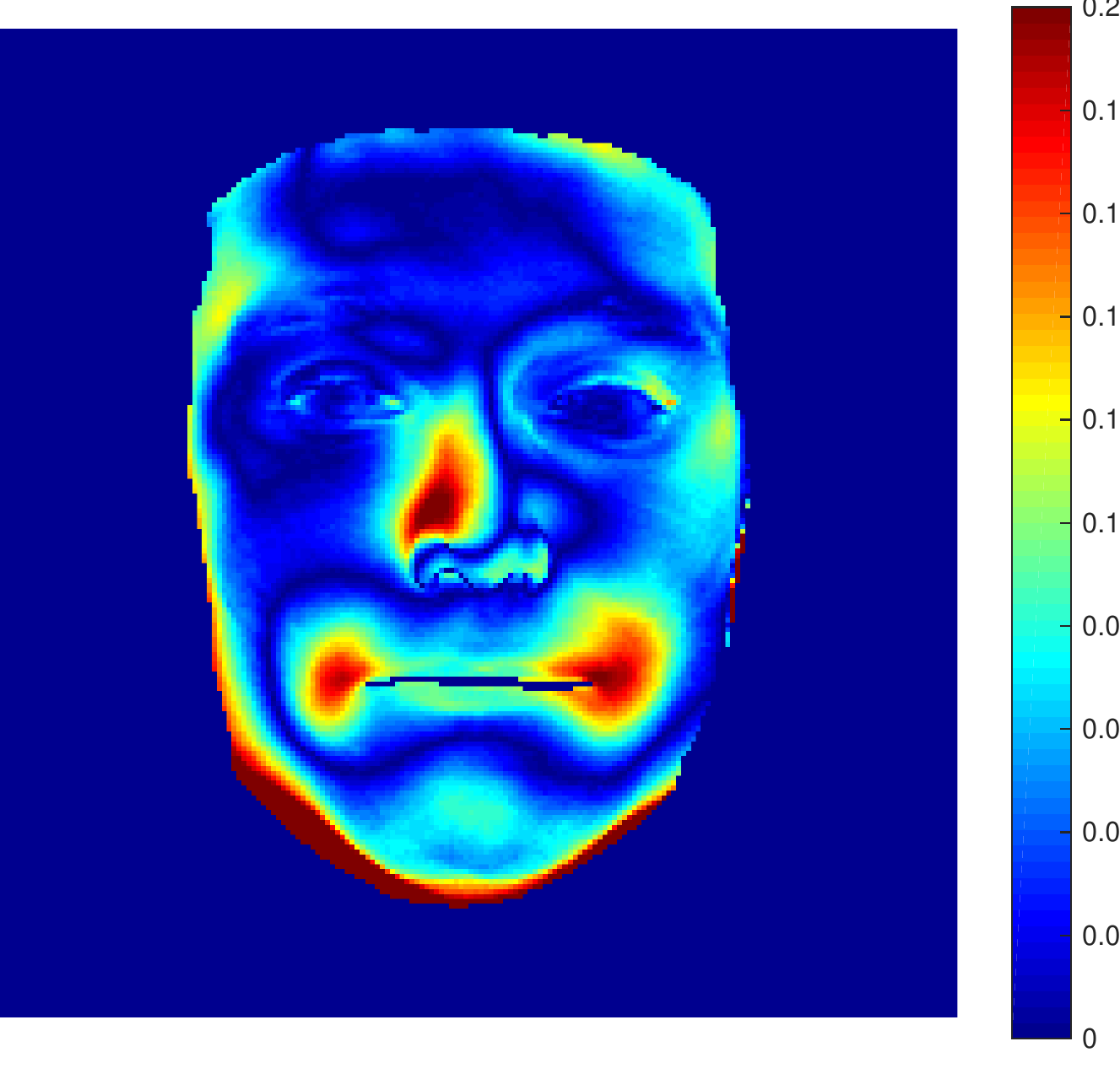}&
 \includegraphics[height=0.18\textwidth,trim = 130 210 100 210,clip=true]{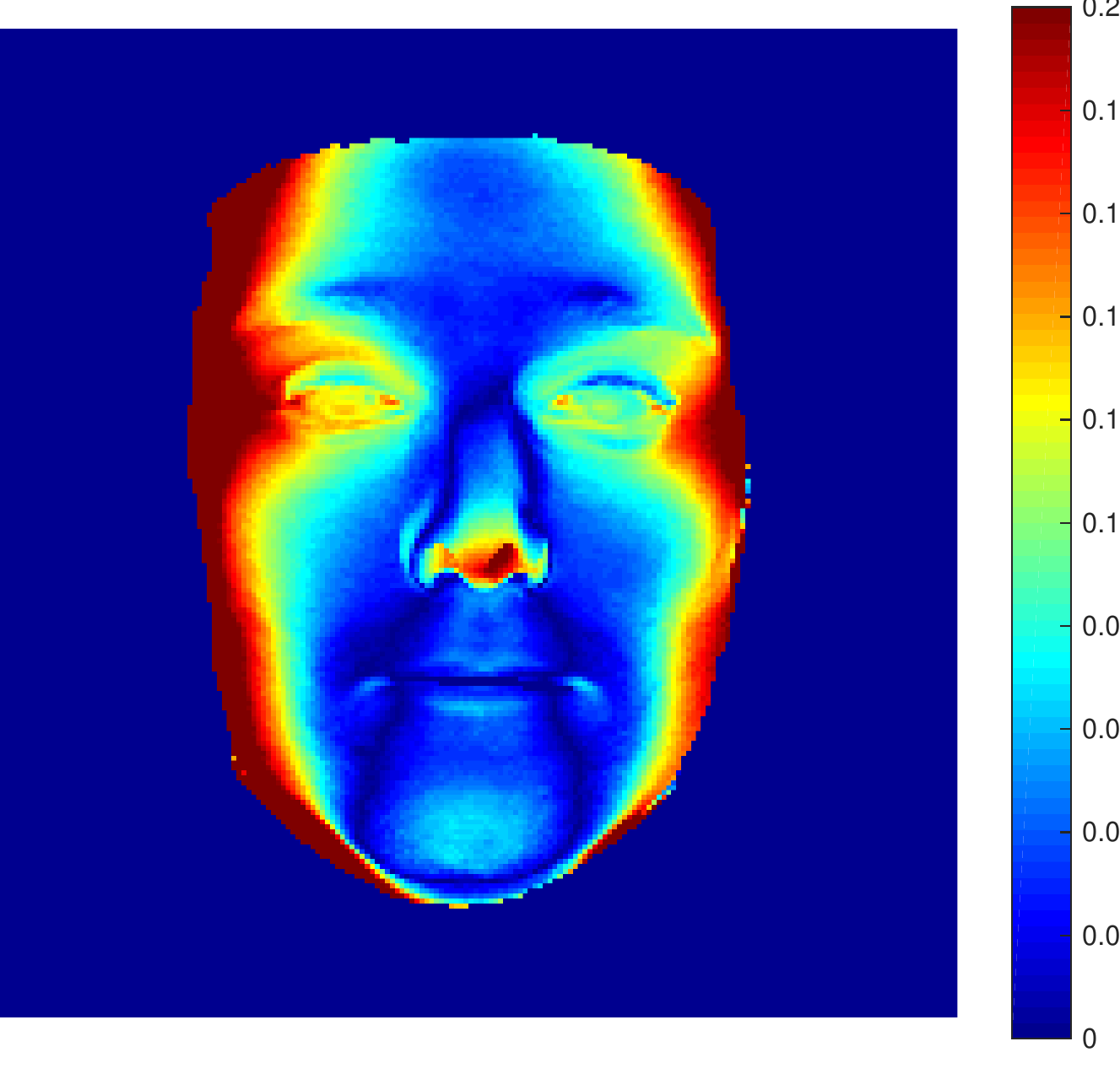}\tabularnewline
 \includegraphics[height=0.18\textwidth,trim = 130 265 150 210,clip=true]{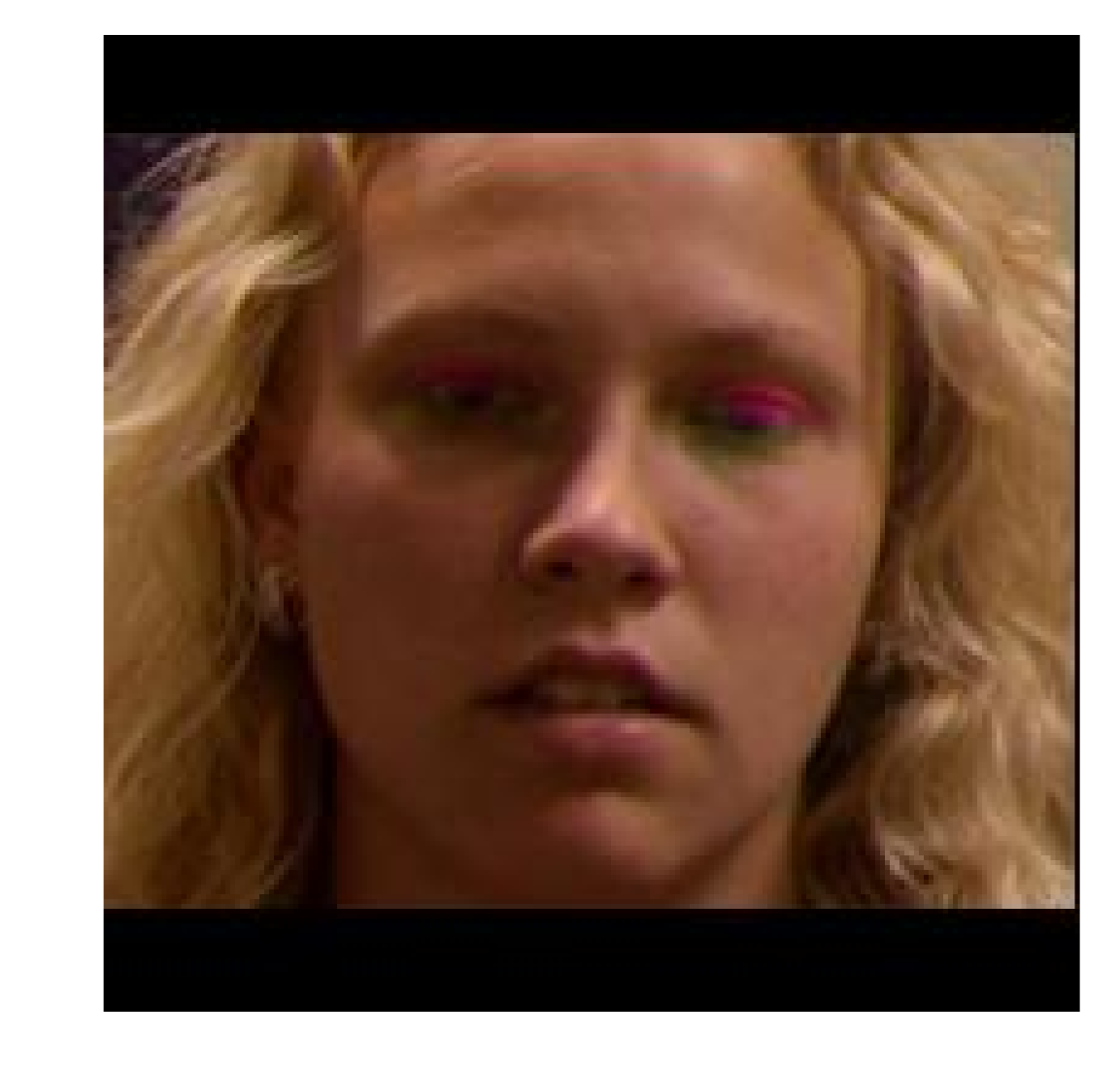}&
 \includegraphics[height=0.18\textwidth,trim = 130 210 100 210,clip=true]{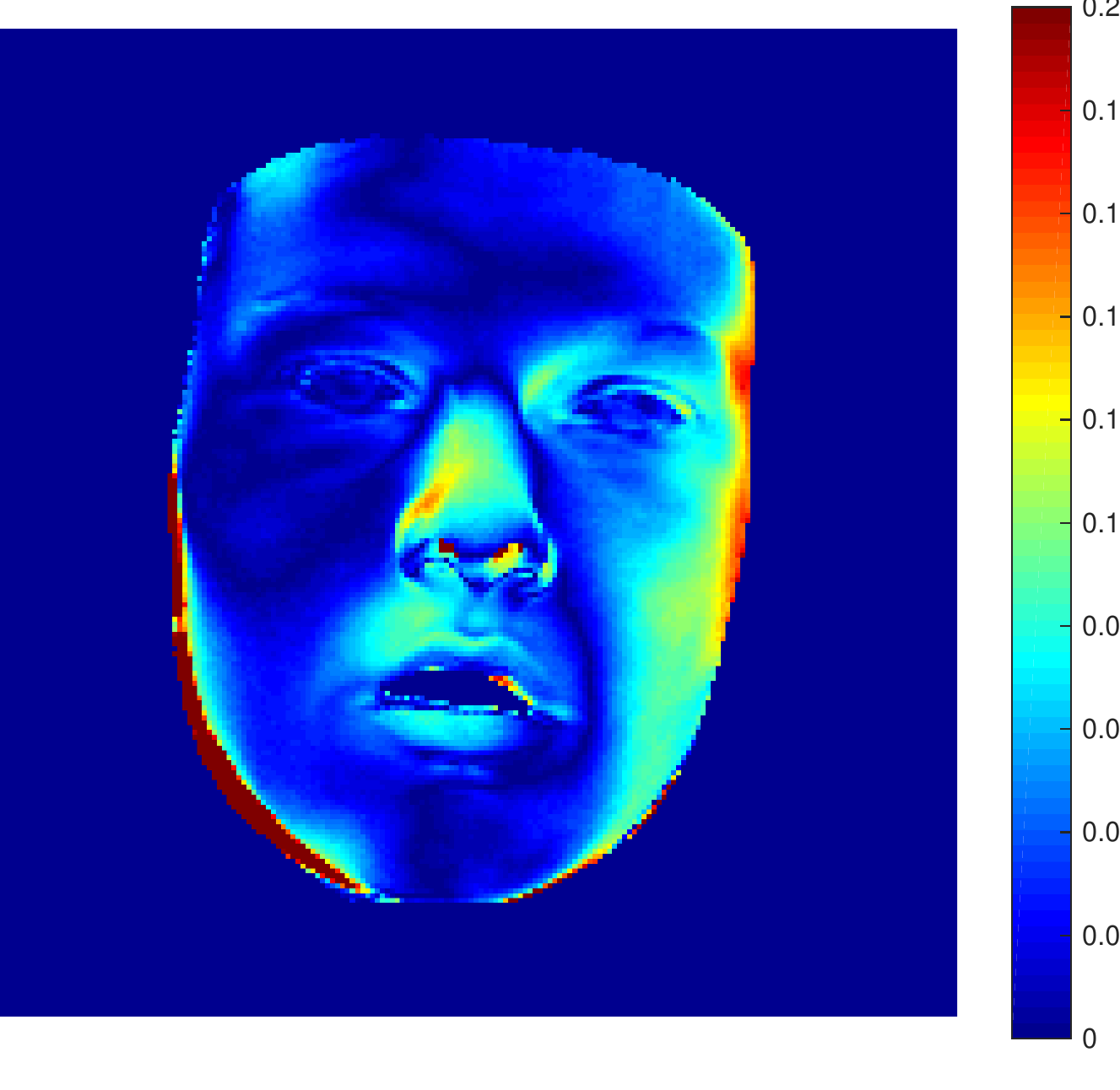}&
 \includegraphics[height=0.18\textwidth,trim = 130 210 100 210,clip=true]{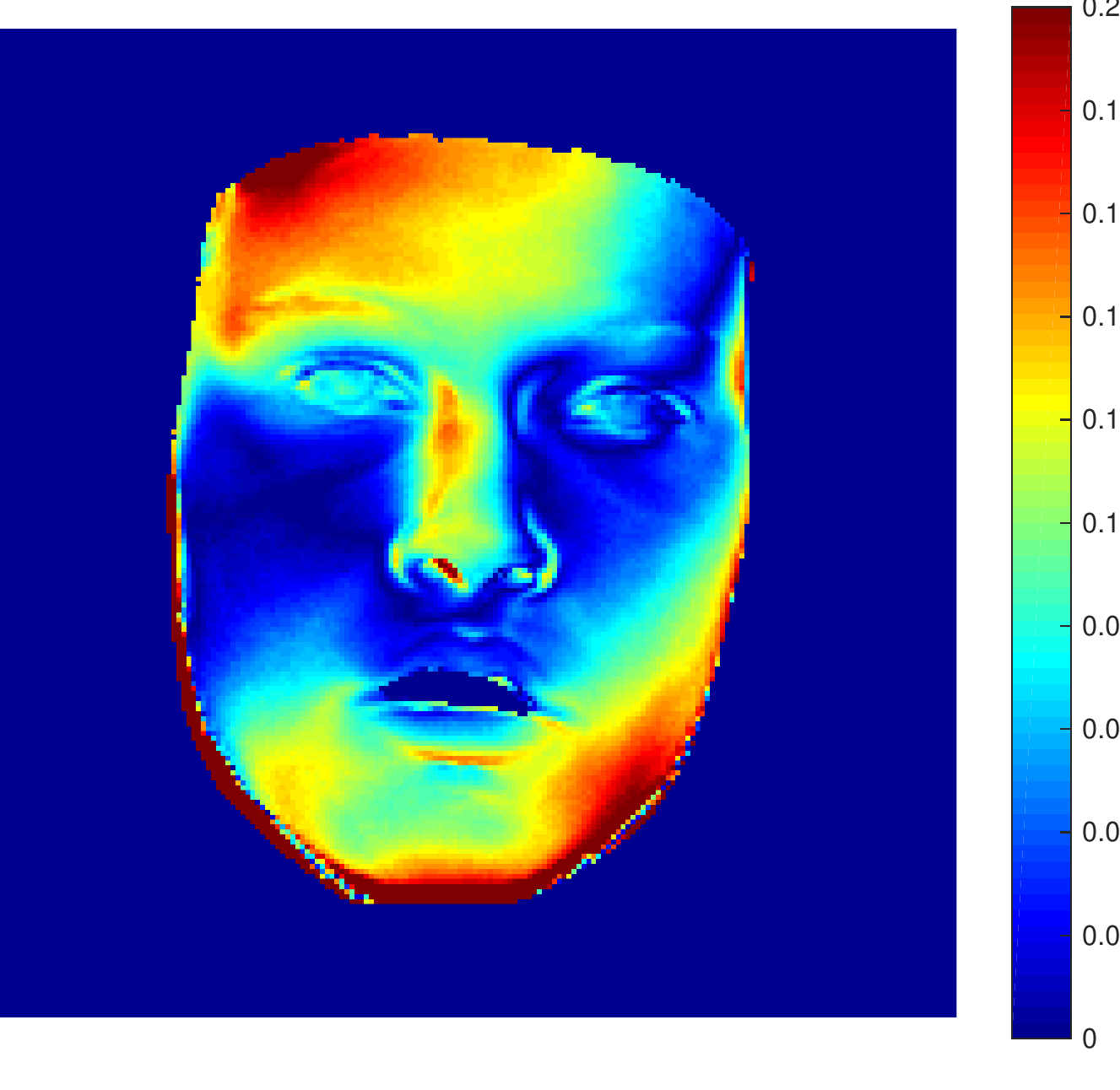}&
 \includegraphics[height=0.18\textwidth,trim = 130 210 100 210,clip=true]{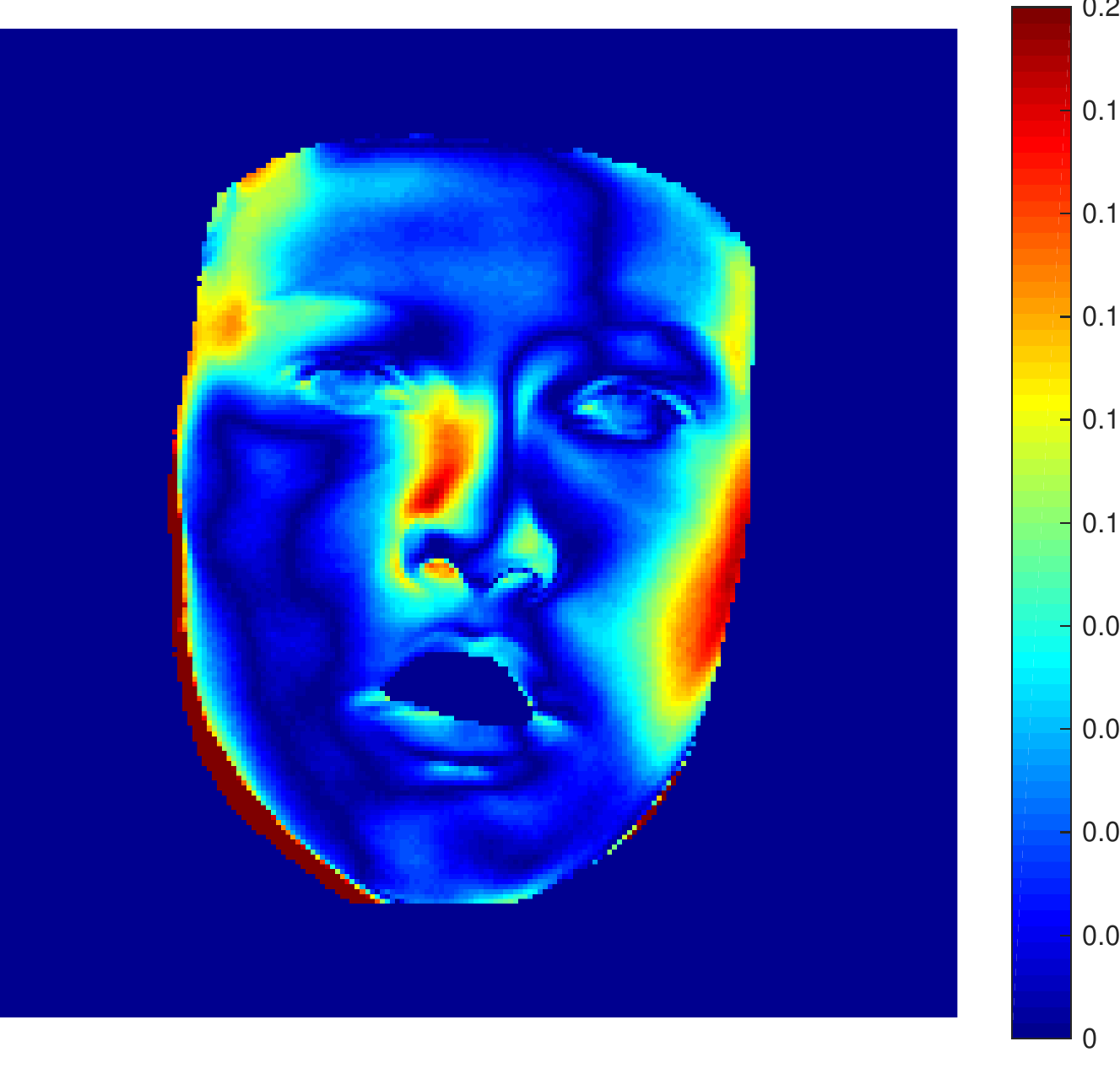}&
 \includegraphics[height=0.18\textwidth,trim = 130 210 100 210,clip=true]{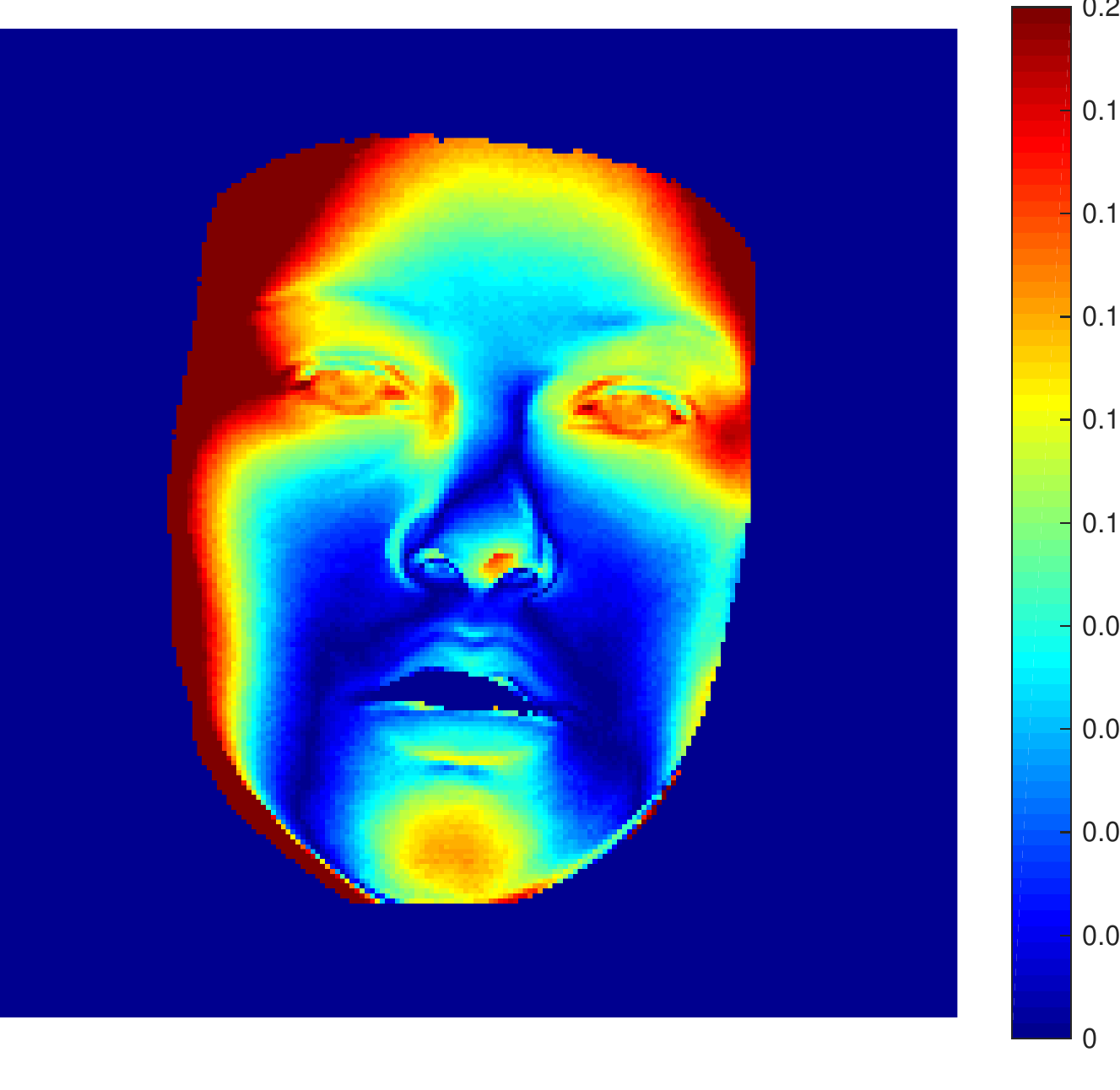}\tabularnewline
  \includegraphics[height=0.18\textwidth,trim = 130 265 150 210,clip=true]{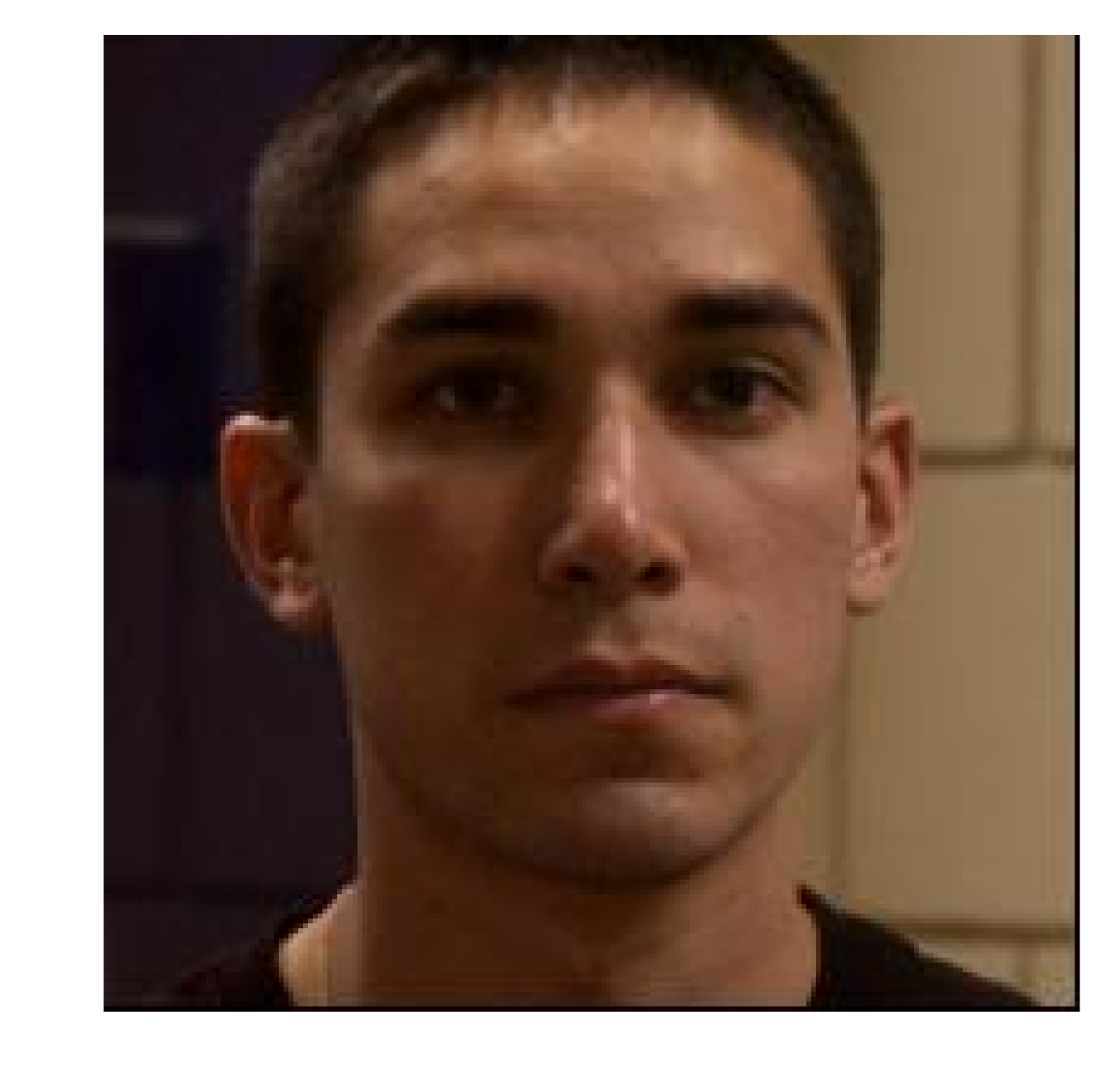}&
 \includegraphics[height=0.18\textwidth,trim = 130 210 100 210,clip=true]{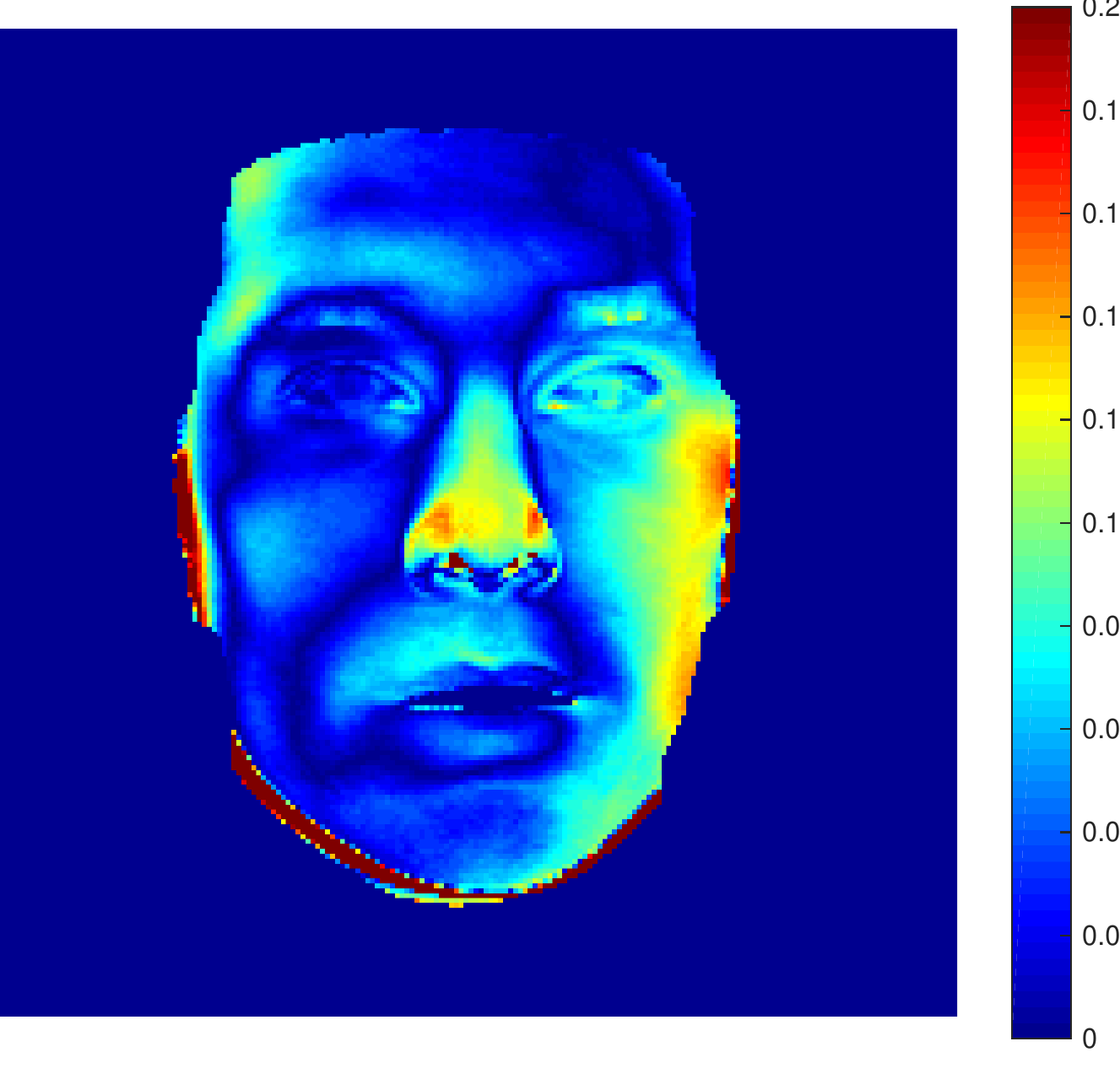}&
 \includegraphics[height=0.18\textwidth,trim = 130 210 100 210,clip=true]{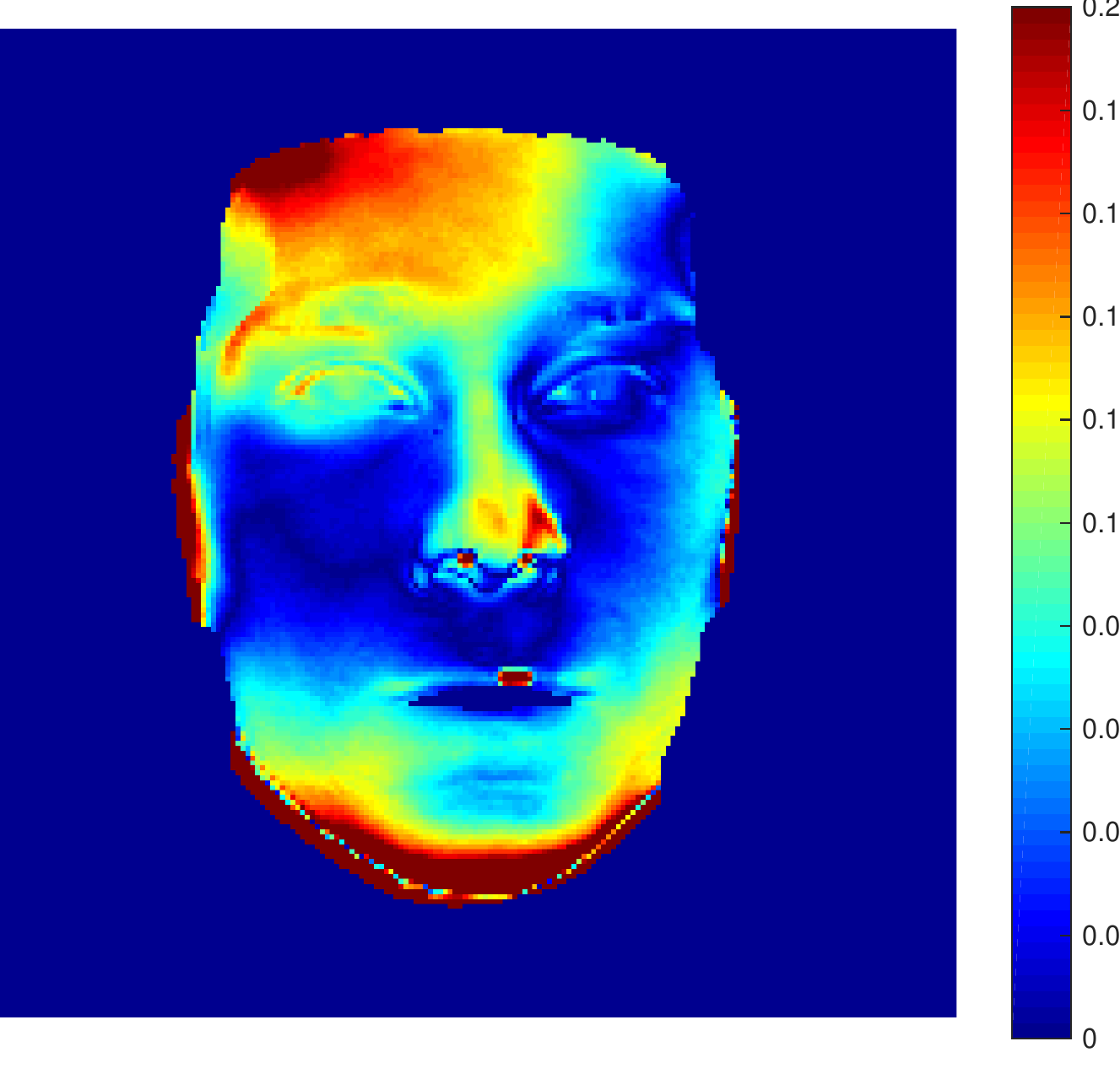}&
 \includegraphics[height=0.18\textwidth,trim = 130 210 100 210,clip=true]{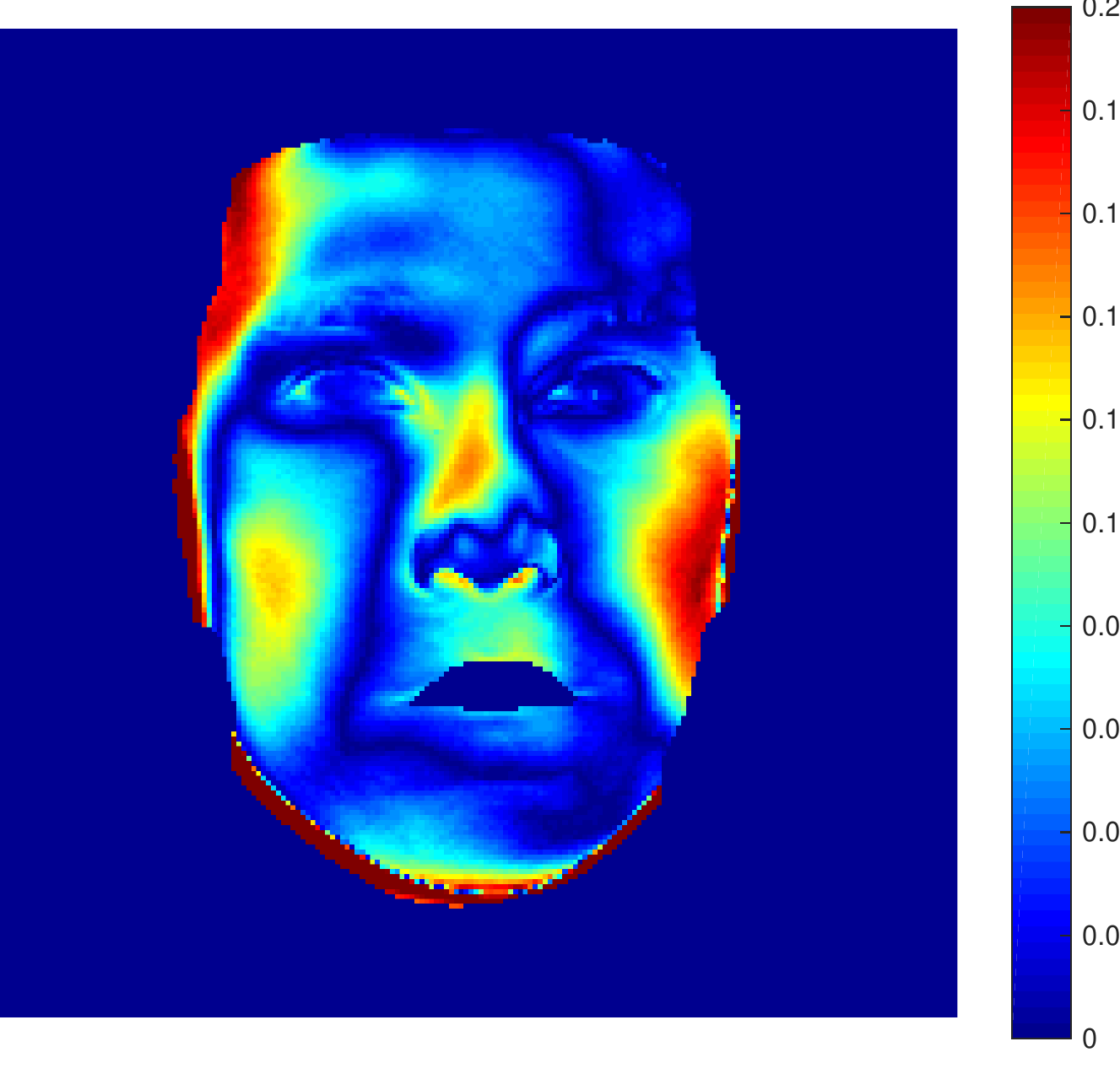}&
 \includegraphics[height=0.18\textwidth,trim = 130 210 100 210,clip=true]{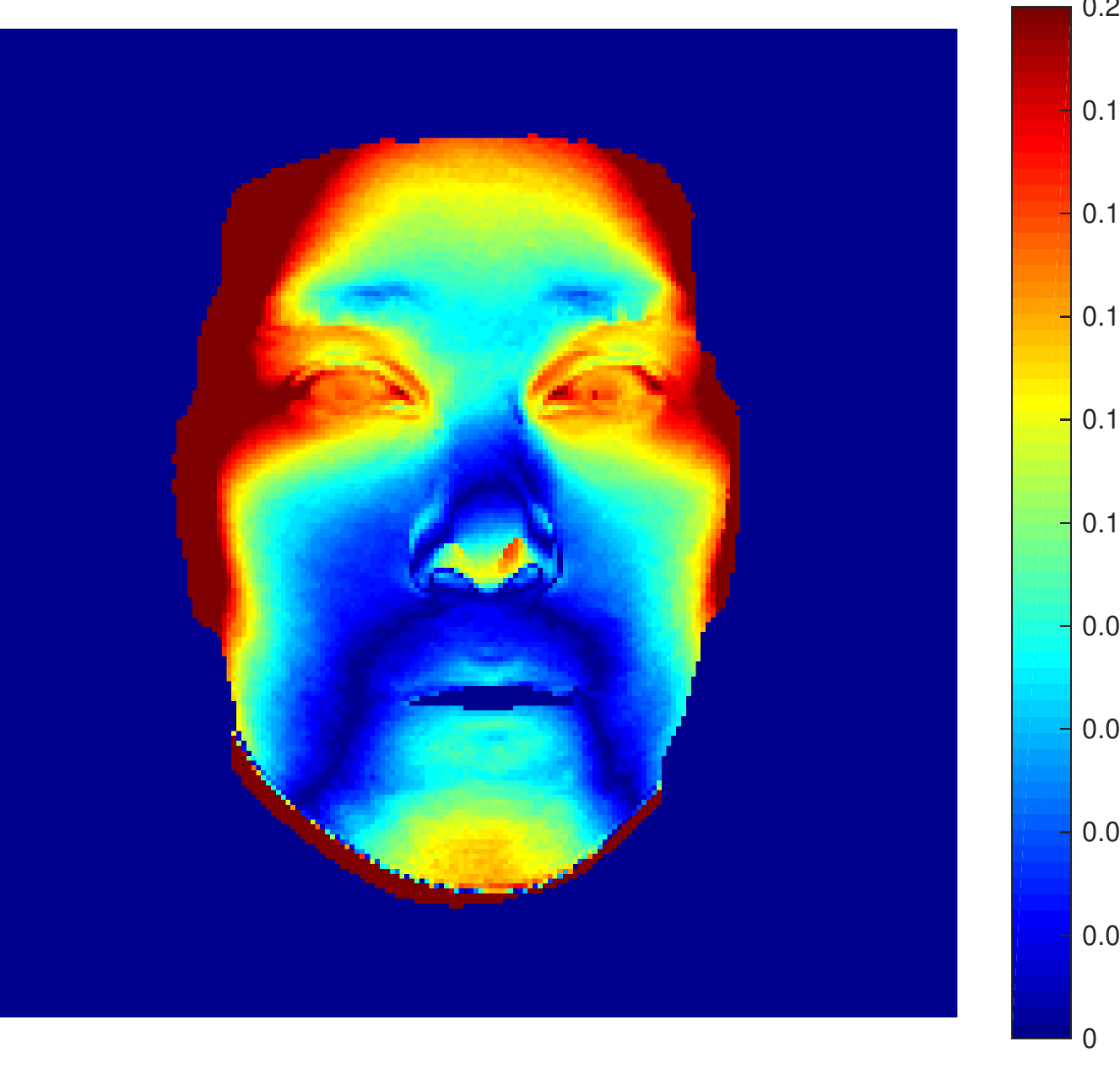}\tabularnewline
    \includegraphics[height=0.18\textwidth,trim = 130 265 150 210,clip=true]{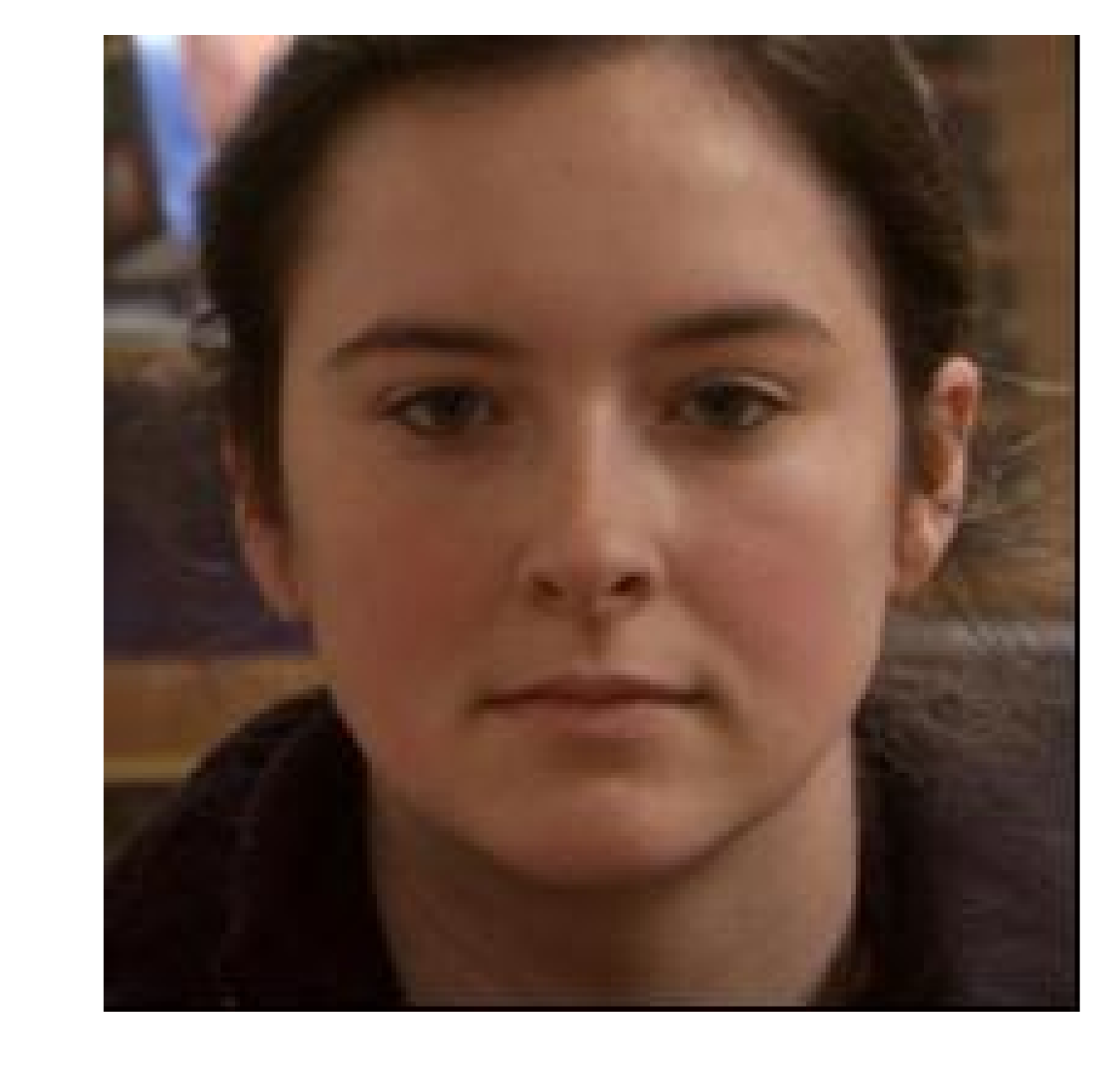}&
 \includegraphics[height=0.18\textwidth,trim = 130 210 100 210,clip=true]{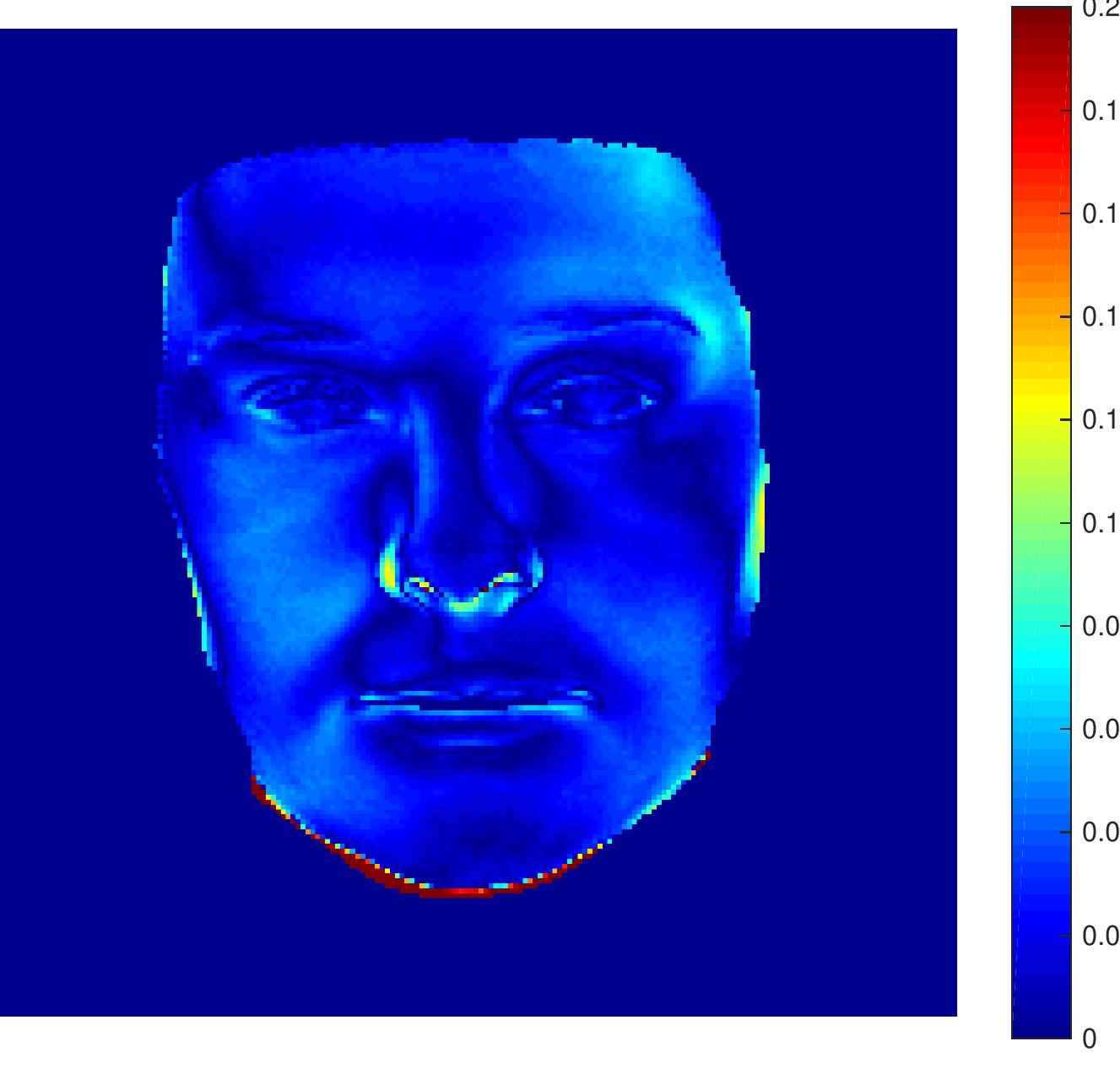}&
 \includegraphics[height=0.18\textwidth,trim = 130 210 100 210,clip=true]{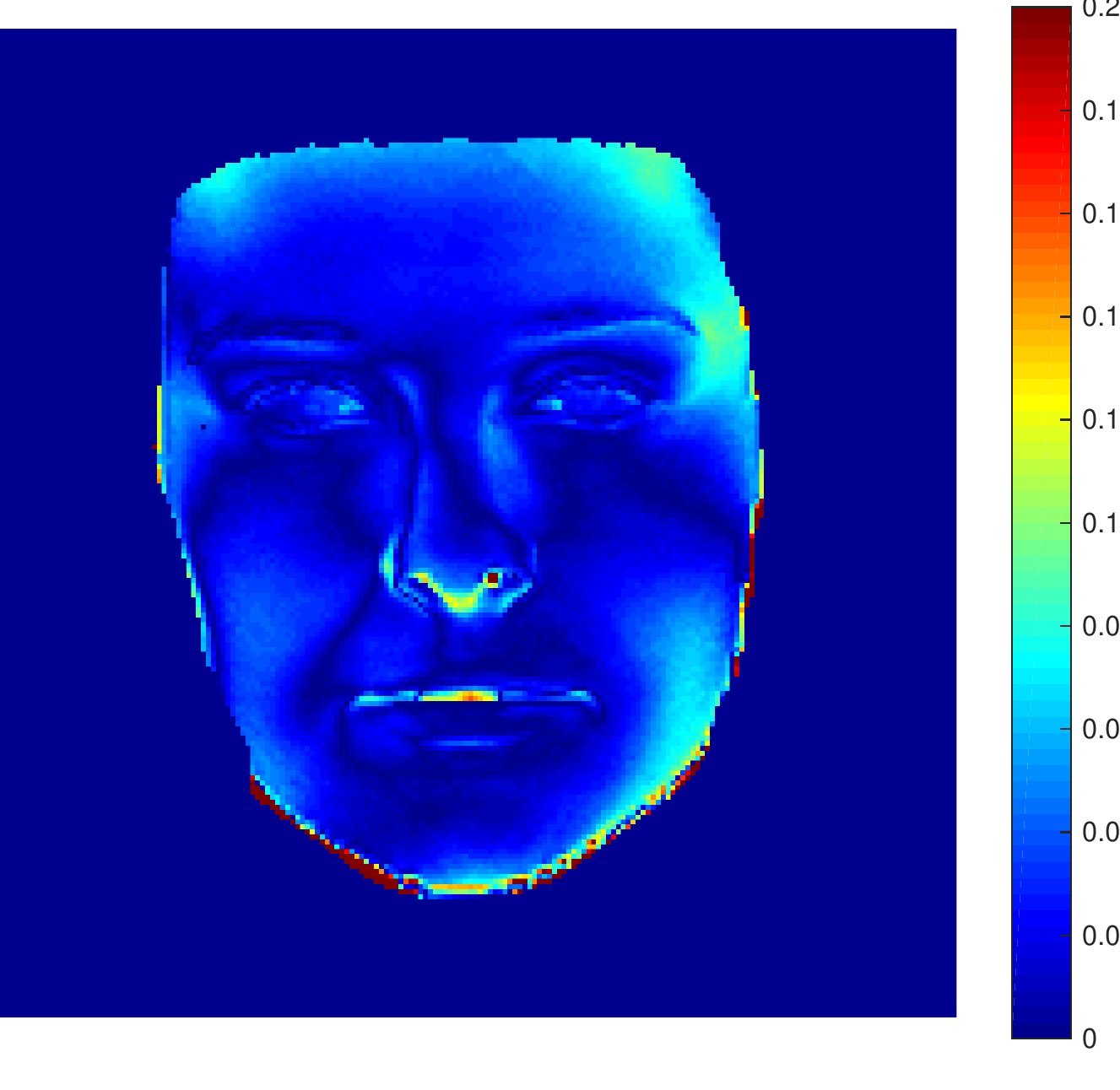}&
 \includegraphics[height=0.18\textwidth,trim = 130 210 100 210,clip=true]{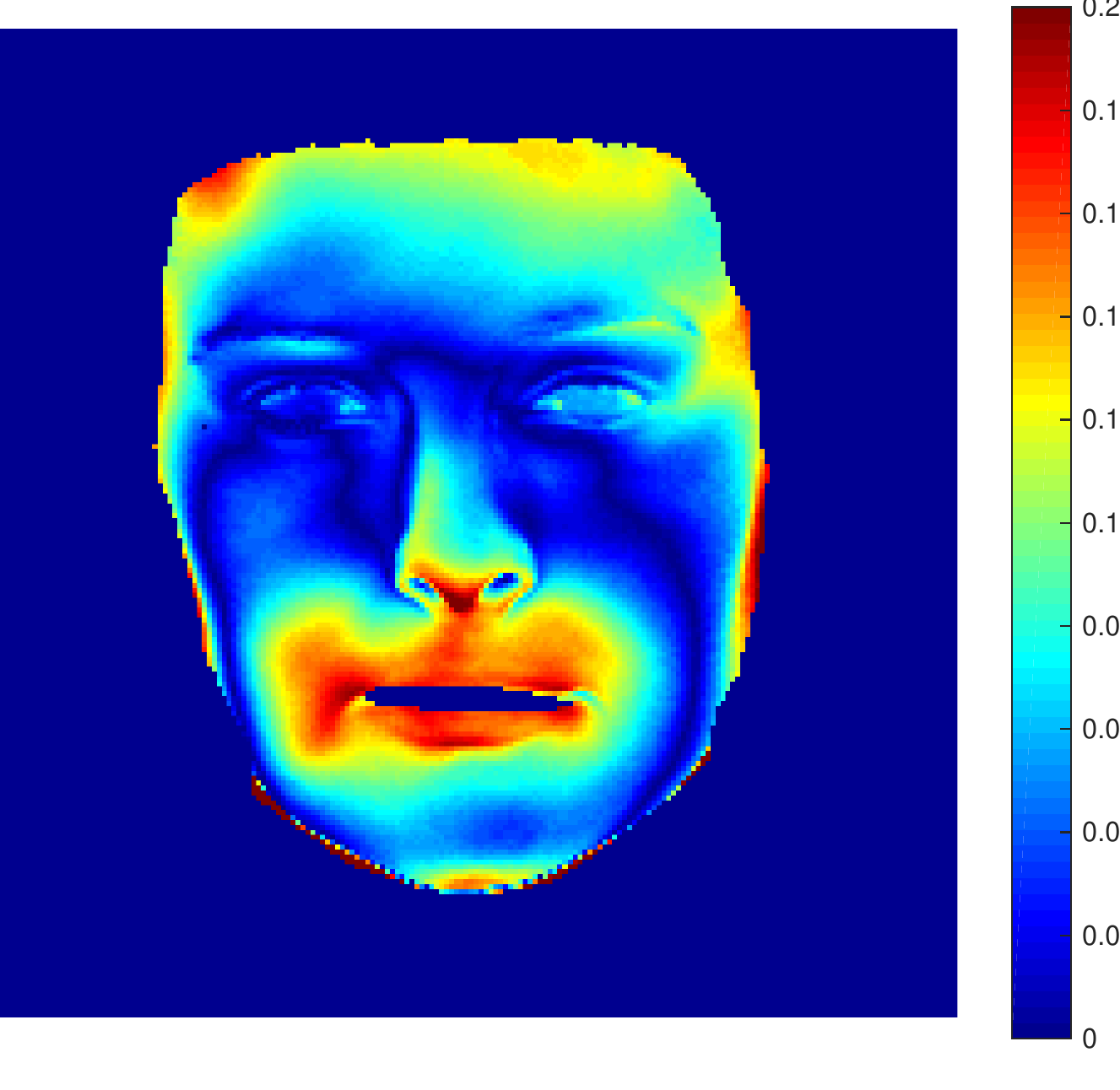}&
 \includegraphics[height=0.18\textwidth,trim = 130 210 100 210,clip=true]{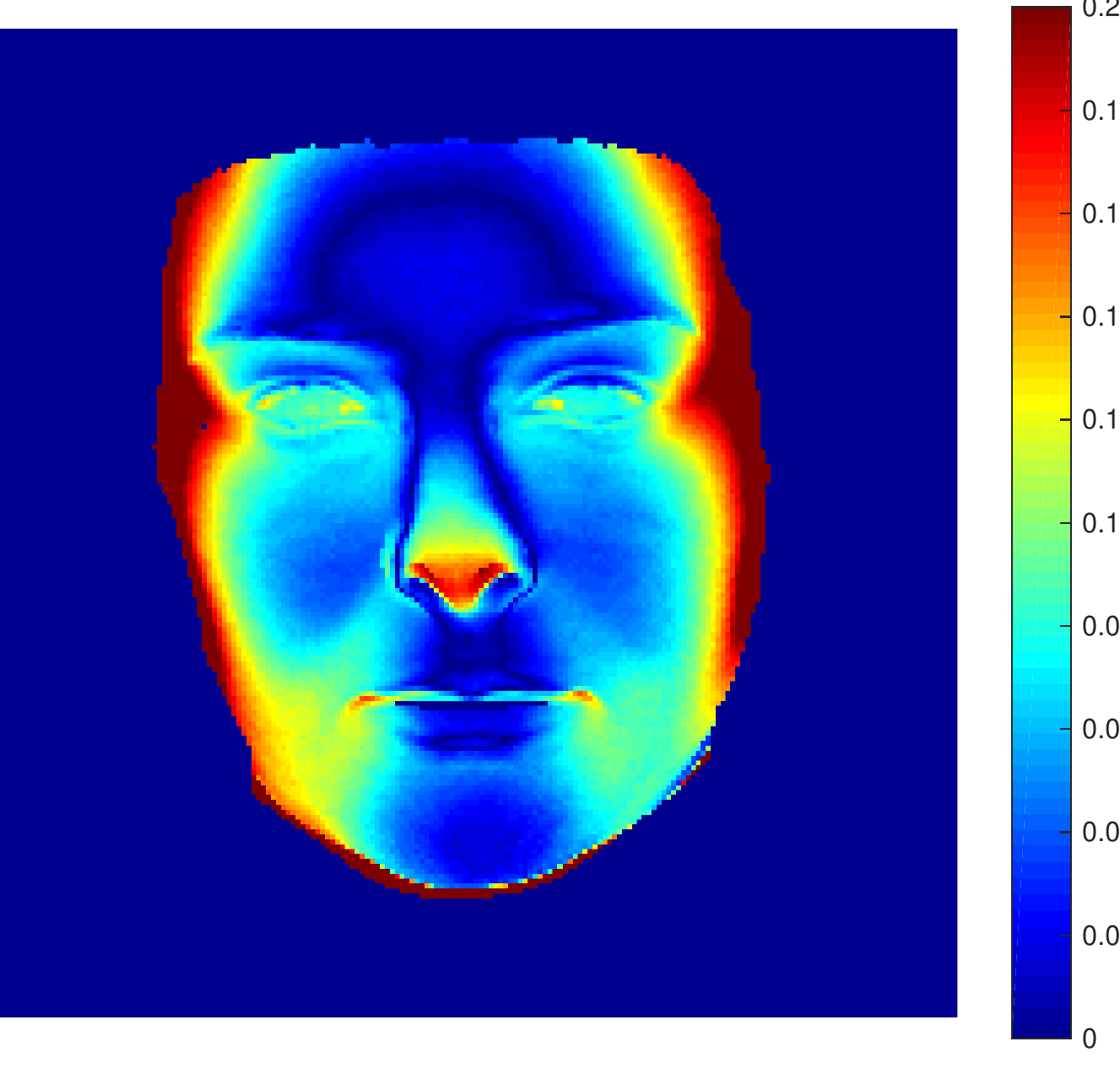}\tabularnewline
 Input & Ours & \cite{kemelmacher20113d} & \cite{richardson20163d} & \cite{zhu2015high}
\end{tabular}
    \caption{Absolute depth error heat maps of different methods. }
    \label{fig:vis_hme_example}
\end{figure}

Here, we demonstrate a quantitative analysis of the performance of the proposed method. The absolute error heat maps in Figure \ref{fig:vis_hme_example} present the typical error distribution of the proposed method versus those of other techniques \cite{kemelmacher20113d,richardson20163d,zhu2015high}.

\end{document}